\pdfoutput = 1
\documentclass[11pt]{article}
\usepackage{fullpage} 
\usepackage[english]{babel}
\usepackage[utf8x]{inputenc}
\usepackage[T1]{fontenc}
\usepackage{bbm,bm}
\usepackage{bbold}
\usepackage{parskip}
\usepackage{amsmath}
\usepackage{amssymb}
\usepackage{amsthm}
\usepackage{thm-restate}
\usepackage{booktabs}
\usepackage{array}
\usepackage{algorithm}
\usepackage{algpseudocode}

\usepackage{hyperref}
\hypersetup{
	colorlinks=true,
	linkcolor=blue!70!black,
	citecolor=blue!70!black,
	urlcolor=blue!70!black
}
\usepackage{graphicx}
\usepackage[english]{babel}
\usepackage[utf8x]{inputenc}
\usepackage[T1]{fontenc}
\usepackage{mathtools}
\usepackage{cleveref}
\usepackage{parskip}
\usepackage{xcolor}
\newtheorem{theorem}{Theorem}
\newtheorem{lemma}{Lemma}
\newtheorem{fact}{Fact}
\newtheorem{definition}{Definition}
\newtheorem{corollary}{Corollary}
\newtheorem{proposition}{Proposition}
\newtheorem{model}{Model}
\newtheorem{assumption}{Assumption}

\newcommand{\defeq}{:=}

\newcommand{\norm}[1]{\left\lVert#1\right\rVert}
\newcommand{\norms}[1]{\lVert#1\rVert}
\newcommand{\normop}[1]{\left\lVert#1\right\rVert_{\textup{op}}}
\newcommand{\normf}[1]{\left\lVert#1\right\rVert_{\textup{F}}}

\newcommand{\normsop}[1]{\lVert#1\rVert_{\textup{op}}}

\newcommand{\eps}{\epsilon}
\newcommand{\lam}{\lambda}
\newcommand{\R}{\mathbb{R}}

\newcommand{\N}{\mathbb{N}}
\newcommand{\diag}[1]{\textbf{\textup{diag}}\left(#1\right)}
\newcommand{\half}{\frac{1}{2}}

\newcommand{\E}{\mathbb{E}}

\newcommand{\Nor}{\mathcal{N}}

\newcommand{\id}{\mathbf{I}}

\newcommand{\dd}{\textup{d}}

\newcommand{\polylog}{\textup{polylog}}
\newcommand{\Par}[1]{\left(#1\right)}
\newcommand{\Brack}[1]{\left[#1\right]}
\newcommand{\Brace}[1]{\left\{#1\right\}}
\newcommand{\Abs}[1]{\left|#1\right|}

\newcommand{\alg}{\mathcal{A}}
\newcommand{\ind}{\mathbb{I}}
\newcommand{\mzero}{\mathbf{0}}
\newcommand{\Sym}{\mathbb{S}}
\newcommand{\PSD}{\Sym_{\succeq \mzero}}
\newcommand{\nnz}{\mathrm{nnz}}
\newcommand{\poly}{\mathrm{poly}}

\newcommand{\bk}{\color{black}}

\newcommand{\RIP}{\mathrm{RIP}}
\newcommand{\MI}{\mathrm{MI}}
\newcommand{\clip}{\mathrm{clip}}
\newcommand{\supp}{\mathrm{supp}}
\newcommand{\range}{\mathrm{range}}

\newcommand{\vth}{\boldsymbol{\theta}}

\newcommand{\matp}{\mathbf{P}}
\newcommand{\vxi}{\boldsymbol{\xi}}
\newcommand{\sig}{\sigma}

\newcommand{\twonorm}[1]{\norm{#1}_2}
\newcommand{\onenorm}[1]{\norm{#1}_1}
\newcommand{\infnorm}[1]{\norm{#1}_\infty}
\newcommand{\msig}{\mathbf{\Sigma}}
\newcommand{\vmu}{\boldsymbol{\mu}}
\newcommand{\vlam}{\boldsymbol{\lambda}}
\newcommand{\vzero}{\mathbf{0}}
\newcommand{\vone}{\mathbf{1}}
\newcommand{\vths}{\vth^\star}
\newcommand{\vhth}{\widehat{\vth}}
\newcommand{\vthp}{\vth^\prime}
\newcommand{\pisupp}{\pi_{\supp}}
\newcommand{\tpisupp}{\tilde{\pi}_{\supp}}

\newcommand{\event}{\mathcal{E}}

\newcommand{\tv}[1]{D_{\textup{TV}}(#1)}

\newcommand{\Bern}{\mathrm{Bern}}
\newcommand{\Lap}{\mathrm{Lap}}

\newcommand{\vvm}{\vv^-}
\newcommand{\vvp}{\vv^+}
\newcommand{\sigm}{\sigma_-}
\newcommand{\sigp}{\sigma_+}
\newcommand{\vum}{\vu^-}
\newcommand{\vup}{\vu^+}

\newcommand{\Prob}{\mathbb{P}}
\renewcommand{\Pr}{\Prob}

\newcommand{\hth}{\widehat{\theta}}

\newcommand{\vhi}{\varphi}
\newcommand{\met}{\mathrm{m}}

\newcommand{\ma}{{\mathbf{A}}}

\newcommand{\mm}{{\mathbf{M}}}

\newcommand{\muu}{{\mathbf{U}}}
\newcommand{\mv}{{\mathbf{V}}}

\newcommand{\mx}{{\mathbf{X}}}

\newcommand{\mproj}{\boldsymbol{\Pi}}
\newcommand{\mxtx}{\mx^\top \mx}

\newcommand{\va}{{\mathbf{a}}}
\newcommand{\vb}{{\mathbf{b}}}
\newcommand{\vc}{{\mathbf{c}}}

\newcommand{\ve}{{\mathbf{e}}}

\newcommand{\vp}{{\mathbf{p}}}
\newcommand{\vq}{{\mathbf{q}}}
\newcommand{\vr}{{\mathbf{r}}}
\newcommand{\vs}{{\mathbf{s}}}

\newcommand{\vu}{{\mathbf{u}}}
\newcommand{\vv}{{\mathbf{v}}}
\newcommand{\vw}{{\mathbf{w}}}
\newcommand{\vx}{{\mathbf{x}}}
\newcommand{\vy}{{\mathbf{y}}}
\newcommand{\vz}{{\mathbf{z}}}
\newcommand{\vsig}{\boldsymbol{\sigma}}

\newcommand{\vDelta}{\boldsymbol{\Delta}}
\newcommand{\hvw}{\hat{\vw}}
\newcommand{\hvz}{\hat{\vz}}

\newcommand{\calA}{{\mathcal{A}}}

\newcommand{\calF}{{\mathcal{F}}}
\newcommand{\calG}{{\mathcal{G}}}

\newcommand{\calK}{{\mathcal{K}}}

\newcommand{\calN}{{\mathcal{N}}}

\newcommand{\bbC}{{\mathbb{C}}}

\newcommand{\bbP}{{\mathbb{P}}}

\newcommand{\bbR}{{\mathbb{R}}}

\newcommand{\bas}[1]{\begin{align*}#1\end{align*}}
\newcommand{\ba}[1]{\begin{align}#1\end{align}}

\newcommand{\paren}[1]{\left(#1\right)}
\newcommand{\bra}[1]{\left[#1\right]}
\newcommand{\cbra}[1]{\left\{#1\right\}}

\newcommand{\bb}[1]{\left(#1\right)}

\newcommand{\rejectionsample}{\mathsf{RejectionSample}}
\newcommand{\productsg}{\mathsf{ProductSampleGaussian}}
\newcommand{\postsg}{\mathsf{PosteriorSampleGaussian}}
\newcommand{\productsl}{\mathsf{ProductSampleLaplace}}
\newcommand{\postsl}{\mathsf{PosteriorSampleLaplace}}

\newcommand{\cps}{\mathsf{ConditionalPoisson}}

\newcommand{\tpi}{\tilde{\pi}}
\newcommand{\tP}{\widetilde{P}}
\newcommand{\tZ}{\widetilde{Z}}
\crefname{assumption}{assumption}{assumptions}

\newcommand{\diff}{\mathrm{d}}

\DeclareMathOperator*{\argmax}{arg\,max}
\DeclareMathOperator*{\argmin}{arg\,min}

\newcommand{\sign}{\mathrm{sign}}

\newcommand{\algest}{\alg_{\textup{est}}}

\definecolor{burntorange}{rgb}{0.8, 0.33, 0.0}

\title{Spike-and-Slab Posterior Sampling in High Dimensions}
\author{Syamantak Kumar\thanks{University of Texas at Austin, \texttt{syamantak@utexas.edu}} \and Purnamrita Sarkar\thanks{University of Texas at Austin, \texttt{purna.sarkar@utexas.edu}} \and Kevin Tian\thanks{University of Texas at Austin, \texttt{kjtian@cs.utexas.edu}} \and Yusong Zhu\thanks{University of Texas at Austin, \texttt{zhuys@utexas.edu}}}

\date{}
\begin{document}
\maketitle
\begin{abstract}
Posterior sampling with the spike-and-slab prior \cite{Mon_MitchellBeauchamp1988}, a popular multimodal distribution used to model uncertainty in variable selection, is considered the theoretical gold standard method for Bayesian sparse linear regression \cite{CarvalhoPS09, Rockova18}. However, designing provable algorithms for performing this sampling task is notoriously challenging. Existing posterior samplers for Bayesian sparse variable selection tasks either require strong assumptions about the signal-to-noise ratio (SNR) \cite{MC_YWJ16}, only work when the measurement count grows at least linearly in the dimension \cite{montanari2024provably}, or rely on heuristic approximations to the posterior.

We give the first provable algorithms for spike-and-slab posterior sampling that apply for any SNR, and use a measurement count sublinear in the problem dimension. Concretely, assume we are given a measurement matrix $\mx \in \R^{n \times d}$ and noisy observations $\vy = \mx \vths + \vxi$ of a signal $\vths$ drawn from a spike-and-slab prior $\pi$ with a Gaussian diffuse density and expected sparsity $k$, where $\vxi \sim \Nor(\vzero_n, \sigma^2 \id_n)$. We give a polynomial-time high-accuracy sampler for the posterior $\pi(\cdot \mid \mx, \vy)$, for any SNR $\sig^{-1} > 0$, as long as $n \ge k^3 \cdot \polylog(d)$ and $\mx$ is drawn from a matrix ensemble satisfying the restricted isometry property. We further give a sampler that runs in near-linear time $\approx nd$ in the same setting, as long as $n \ge k^5 \cdot \polylog(d)$. To demonstrate the flexibility of our framework, we extend our result to spike-and-slab posterior sampling with Laplace diffuse densities, achieving similar guarantees when $\sig = O(\frac 1 k)$ is bounded.
\end{abstract}
\thispagestyle{empty}
\newpage
\tableofcontents
\thispagestyle{empty}

\newpage
\section{Introduction}\label{sec:intro}

Sparse linear regression is a fundamental, well-studied problem in high-dimensional statistics. In this problem, there is an unknown sparse signal $\vths \in \R^d$ that we are trying to estimate, and we are given a measurement matrix $\mx \in \R^{n \times d}$, as well as noisy observations
\begin{equation}\label{eq:slr_model}\vy = \mx \vths + \vxi,\text{ where } \vxi \sim \Nor(\vzero_n, \sig^2 \id_n). \end{equation}
This problem has been studied for decades by statisticians from the perspective of \emph{variable selection}, see e.g., \cite{Hocking76} and references therein, where $\supp(\vths)$ models the ``important features'' for a regression problem, and potentially $|\supp(\vths)| \ll d$. More recently, due to the increasing dimensionality of data in modern statistics and learning tasks, there has been significant interest in understanding the \emph{computational complexity} of sparse linear regression.

One major algorithmic success story in sparse linear regression is a line of work building off breakthrough results of \cite{CandesR05, CandesT05, Donoho06, CandesT06, CandesRT06} for compressed sensing, which we refer to collectively as ``sparse recovery'' methods (some of which are reviewed in Section~\ref{ssec:sparse_recovery}). These algorithms take a frequentist viewpoint of \eqref{eq:slr_model}, and aim to return an estimate $\vhth$ such that $\norms{\vhth - \vths}$ for an appropriate norm $\norm{\cdot}$, e.g., $\ell_2$ or $\ell_\infty$, often via penalized maximum likelihood methods such as the Lasso \cite{tibshirani1996regression}. Impressively, despite the nonconvexity of the problem (i.e., modeling the constraint that $\vths$ is sparse), the aforementioned line of work showed that as long as $\mx$ satisfies the \emph{restricted isometry property} (RIP, cf.\ Definition~\ref{def:RIP}), a sparse vector $\vths$ can be recovered up to a function of the noise level $\sig$ in polynomial time. In fact, sparse recovery is tractable even in the \emph{sublinear measurement} regime $n \ll d$, as long as $n$ is sufficiently larger than the sparsity $|\supp(\vths)|$. This regime is natural for high-dimensional sparse linear regression, because its underconstrained nature necessitates making a structural assumption on the signal $\vths$, e.g., sparsity.

A potential shortcoming of sparse recovery methods is their inability to model uncertainty in variable selection. In particular, these frequentist algorithms commit to a single estimate $\vhth$, and do not return multiple plausible candidate supports. This issue is particularly manifest in the setting of moderate signal-to-noise ratios, where some coordinates of $\vths$ have magnitude at or around the noise level $\sig$. In such settings, it is more appropriate to consider a distribution over estimates to reflect our uncertainty of $\vths$. This motivates the problem of \emph{Bayesian sparse linear regression}, the main problem studied in this paper. In this problem, for some prior distribution $\pi$ that is mostly supported on sparse vectors, the task is to return a sample from the posterior distribution,
\begin{equation}\label{eq:bslr_intro}\vth \sim \pi\Par{\cdot \mid \mx, \vy} \propto \exp\Par{-\frac 1 {2\sig^2}\norm{\vy - \mx \vth}_2^2} \pi(\vth).\end{equation}
Perhaps the most well-studied choice of prior in Bayesian sparse linear regression is the \emph{spike-and-slab} prior, introduced by \cite{Mon_MitchellBeauchamp1988} and subsequently studied in, e.g., \cite{Mon_GeorgeMcCulloch1993, Chipman96, Geweke96}. This prior is parameterized by $\vq \in [0, 1]^d$ and a \emph{diffuse density} $\mu$ governing the non-negligible entries (the ``slab,'' i.e., the selected variables). The prior then takes the product density form\footnote{We refer the reader to Section~\ref{ssec:notation} for notation used throughout the paper, and Section~\ref{ssec:sas_post} for the formal definition of our spike-and-slab posterior sampling problem.}
\begin{equation}\label{eq:sas_intro}
\pi \defeq \bigotimes_{i \in [d]}\Par{(1 - \vq_i)\delta_0 + \vq_i \mu},
\end{equation}
so that the $i^{\text{th}}$ coordinate of $\vths \sim \pi$ is independently set to $0$ except with probability $\vq_i$. Throughout the paper, we define $k \defeq \norm{\vq}_1$ to be the expected sparsity level of $\vths \sim \pi$; we are primarily interested in the regime $k \ll d$. Standard Chernoff bounds show that $|\supp(\vths)| \lesssim k$ with high probability, so that at least the frequentist problem of sparse recovery is tractable with $n \ll d$ measurements.

Sampling from \eqref{eq:bslr_intro} when the prior $\pi$ has the form \eqref{eq:sas_intro} (henceforth, \emph{spike-and-slab posterior sampling}) is often referred to as the ``theoretical ideal'' or ``gold standard'' for modeling uncertainty in Bayesian variable selection \cite{JohnstoneS04, CarvalhoPS09, IshwaranS11, castillo2012needles, Rockova18, PolsonS19}. While the favorable statistical properties of spike-and-slab posterior sampling are well-documented (see e.g.,~\cite{castillo2015bayesian}), actually performing said sampling is notoriously challenging. This is in large part due to the combinatorial and nonconvex nature of \eqref{eq:sas_intro}, which causes the posterior \eqref{eq:bslr_intro} to be a mixture of potentially exponentially-many Gaussians. This difficulty has caused many researchers to consider approximations to \eqref{eq:bslr_intro}, e.g., by relaxing the hard sparsity constraints, considering mean-field approximations, or using other heuristics (see Section~\ref{ssec:related} for further discussion, or the review by~\cite{BaiRG21}). 

We are aware of relatively few spike-and-slab posterior samplers that provably work without restrictive modeling assumptions. The two most relevant to our development are \cite{MC_YWJ16, montanari2024provably}, who both gave samplers in different regimes; we give an extended comparison to both in Section~\ref{ssec:related}. 

In the well-studied setting where $\mu = \Nor(0, 1)$ in \eqref{eq:sas_intro} is a Gaussian, \cite{MC_YWJ16} (which formally studies a different family of sparse priors than \eqref{eq:sas_intro}) gives a sampler that only works under very high or very low signal-to-noise ratios (SNR) $\sig^{-1}$. In such regimes, the posterior distribution either collapses to a single candidate support (as all signal coordinates are above the noise level), or the posterior effectively reduces to the prior (as the noise overwhelms the signal). Both settings appear to be at odds with a major motivation behind Bayesian sparse linear regression, i.e., modeling partial uncertainty in variable selection. Conversely, while \cite{montanari2024provably} works for fairly general diffuse densities and SNRs $(\mu, \sig^{-1})$, it requires that the number of observations $n$
grows at least linearly in the dimension $d$. This regime is again at odds with the high-dimensional, underparameterized setting $n \ll d$ where sparse modeling tasks are typically studied. To the best of our knowledge, the following motivating question for our work has remained largely unaddressed by the existing literature.

\begin{equation}\label{eq:main_q}
\begin{gathered}
\textit{Is there a polynomial-time spike-and-slab posterior sampler} \\
\textit{that uses $n = o(d)$ measurements and works for arbitrary SNRs $\sig^{-1} > 0$?}
\end{gathered}
\end{equation}

\subsection{Our results}\label{ssec:results}

\textbf{Gaussian prior.} Our primary contribution is to resolve \eqref{eq:main_q} when $\mu = \Nor(0, 1)$ is a Gaussian, and the expected sparsity $k$ is sufficiently small. 
For simplicity in this introduction, we only state our results in the case where the measurement matrix $\mx$ has i.i.d.\ Gaussian entries. Our formal theorem statements extend the statements here broadly to any $\mx$ drawn from any commonly-used RIP matrix ensemble (cf.\ Section~\ref{ssec:sparse_recovery} for examples). This can yield improved runtimes when the ensemble supports fast matrix-vector multiplications, e.g., when $\mx$ is a subsampled Discrete Fourier Transform (DFT) matrix.

\begin{theorem}[informal, see Theorem~\ref{thm:sas_basic}, Corollary~\ref{cor:gaussian_X}]\label{thm:sas_basic_intro}
Let $\mu = \calN(0, 1)$ in \eqref{eq:sas_intro}, and suppose $\mx \in \R^{n \times d}$ has i.i.d.\ entries $\sim \Nor(0, \frac 1 n)$. For any $\sigma > 0$, $\delta \in (0, 1)$, and $\vq \in [0, 1]^d$, letting $k \defeq \onenorm{\vq}$, the following hold for the problem of sampling from the posterior $\pi(\cdot \mid \mx, \vy)$ in \eqref{eq:bslr_intro}.
    \begin{enumerate}
        \item If $n = \Omega(k^3 \, \polylog(\frac d \delta))$, there is an algorithm that returns $\vth \sim \pi'$ such that $\tv{\pi^\prime, \pi(\cdot \mid \vy, \mx)} \le \delta$ with probability $\ge 1 - \delta$ over the randomness of $\mx$, $\vths$, and $\vxi$, which runs in time
        \[O\Par{n^2 d^{1.5}\,\textup{polylog}\Par{\frac{d}{\delta\min(1, \sig)}}}.\]
        \item  If $n = \Omega(k^5 \, \polylog(\frac d \delta))$, there is an algorithm that returns $\vth \sim \pi'$ such that $\tv{\pi^\prime, \pi(\cdot \mid \vy, \mx)} \le \delta$ with probability $\ge 1 - \delta$ over the randomness of $\mx$, $\vths$, and $\vxi$, which runs in time
        \[O\Par{nd\log\Par{\frac 1 {\min(1, \sig)}}}.\]
    \end{enumerate}
\end{theorem}

We briefly remark on the form of Theorem~\ref{thm:sas_basic_intro}. There are two sources of failure: the random model instance $(\mx, \vths, \vxi)$, and the inaccuracy of our sampler itself. With positive probability, the model will produce an instance with an intractable posterior sampling task (e.g., if $\mx$ is not RIP or $\vths$ is not sparse). Assuming the success of the model, our sampler gives high-accuracy convergence guarantees to the true posterior (achieving total variation $\delta$ with a $\polylog(\frac 1 \delta)$ dependence).

Our required sample complexity, or the number of random measurements $n$ in Theorem~\ref{thm:sas_basic_intro}, scales sublinearly in $d$ for sufficiently small $k$. Qualitatively, this draws a parallel between Theorem~\ref{thm:sas_basic_intro} and results in the sparse recovery literature, which also estimate $\vths$ (via optimization instead of sampling) from $n = \poly(k, \log(\frac d \delta))$ measurements. It is plausible that \eqref{eq:main_q} could even be answered affirmatively given the natural limit of $n \approx k$ observations. However, this would likely require fundamentally new techniques, which we leave as an exciting open question. 

\textbf{Laplace prior.} Several of the qualitative statements we are able to prove about our Gaussian posterior sampler in Theorem~\ref{thm:sas_basic_intro} remain true for general diffuse densities $\mu$ in \eqref{eq:sas_intro} (which we expand on in Section~\ref{ssec:approach}). For this reason, at least for a range of SNRs $\sigma^{-1} > 0$, we believe our techniques extend to posterior sample from \eqref{eq:bslr_intro} for a fairly broad set of diffuse densities $\mu$ in \eqref{eq:sas_intro}. This is a desirable trait, shared e.g., by the sampler of \cite{montanari2024provably}, which applies to the regime $n = \Omega(d)$. As a demonstration of our framework, we show how to sample from \eqref{eq:bslr_intro} when $\mu = \Lap(0, 1)$ in \eqref{eq:sas_intro}.

\begin{theorem}[informal, see Theorem~\ref{thm:sas_laplace}, Corollary~\ref{cor:gaussian_X_laplace}]\label{thm:laplace_informal}
Let $\mu = \Lap(0, 1)$ in \eqref{eq:sas_intro}, and suppose $\mx \in \R^{n \times d}$ has i.i.d.\ entries $\sim \Nor(0, \frac 1 n)$. For any  $\delta \in (0, 1)$, $\sigma \in (0, O((k + \log(\frac 1 \delta))^{-1})]$, and $\vq \in [0, 1]^d$, letting $k \defeq \onenorm{\vq}$, the following hold for the problem of sampling from the posterior $\pi(\cdot \mid \mx, \vy)$ in \eqref{eq:bslr_intro}.
    \begin{enumerate} 
        \item If $n = \Omega(k^3 \, \polylog(\frac d \delta))$, there is an algorithm that returns 
        $\vth \sim \pi'$ such that $\tv{\pi', \pi(\cdot \mid \vy, \mx)} \le \delta$ with probability $\ge 1 - \delta$ over the randomness of $\mx, \vths$, and $\vxi$, which runs in time
       \[O\Par{n^2 d^{1.5}\log\Par{\frac{d}{\sigma\delta}} + \frac{k^4}{\delta^2}\,\polylog\Par{\frac{d}{\sigma\delta}}}.\]
        \item  If $n = \Omega(k^5 \, \polylog(\frac d \delta))$, there is an algorithm that returns 
        $\vth \sim \pi'$ such that $\tv{\pi', \pi(\cdot \mid \vy, \mx)} \le \delta$ with probability $\ge 1 - \delta$ over the randomness of $\mx, \vths$, and $\vxi$, which runs in time
        \[O\Par{nd\log\Par{\frac{d}{\sigma\delta}} + \frac{k^4}{\delta^2}\,\polylog\Par{\frac{d}{\sigma\delta}}}.\]
    \end{enumerate}
\end{theorem}

Theorems~\ref{thm:sas_basic_intro} and~\ref{thm:laplace_informal} have two main differences. First, the runtime of Theorem~\ref{thm:laplace_informal} scales polynomially with $\frac 1 \delta$ (in a low-order term), as opposed to Theorem~\ref{thm:sas_basic_intro}'s polylogarithmic scaling. This is because in the Gaussian setting, there is an explicit formula for the proportionality constant of each posterior component (corresponding to a possible $\supp(\vths)$). However, in the Laplace setting, we must rely on Monte Carlo estimation of normalizing constants, leading to a total variation error that scales with our estimate's approximation factor. Second, due to technical obstacles in the analysis of a rejection sampling step of our algorithm, Theorem~\ref{thm:laplace_informal} is only able to tolerate a bounded noise level $\sigma \lesssim \frac 1 k$. We think it is an interesting open direction to characterize the types of $\mu$ for which \eqref{eq:main_q} can be answered affirmatively (potentially including $\mu = \Lap(0, 1)$).

\subsection{Overview of approach}\label{ssec:approach}

In this section, we overview the pieces of our algorithm and its analysis. A straightforward calculation based on Bayes' theorem (see Lemmas~\ref{lem:unnorm_post} and~\ref{lem:theta_sample}) shows that when $\mu = \Nor(0, 1)$ in \eqref{eq:sas_intro}, the posterior distribution \eqref{eq:bslr_intro} is a mixture of Gaussians:
\begin{equation}\label{eq:mixture_intro}
\begin{gathered}
\pi\Par{\cdot \mid \mx, \vy} = \sum_{S \subseteq [d]} w(S) \, \Nor\Par{\ma_S^{-1} \vb_S, \ma_S^{-1}}, \\
\text{where } w(S) \propto \Par{\prod_{i \in S} \frac{\vq_i}{1 - \vq_i}}\exp\Par{\half\norm{\vb_S}_{\ma_S^{-1}}^2} \frac 1 {\sqrt{\det \ma_S}}, \\
\ma_S \in \R^{S \times S} \defeq \frac 1 {\sig^2}\Brack{\mxtx}_{S \times S} + \id_S,\; \vb_S \in \R^S \defeq \frac 1 {\sig^2}\mx_{S:}^\top \vy,
\end{gathered}
\end{equation}
and $w$ forms a distribution over $\{S \mid S \subseteq [d]\}$. The key challenge is that, even if one can establish that $\pi(\cdot \mid \mx, \vy)$ is mostly supported on $\approx k$-sparse subsets $S$ (where RIP holds and estimation is easy), there are still $\approx d^k$ candidate small subsets to enumerate over. Thus, a natural starting point is to correct an easier-to-sample proposal distribution that approximates \eqref{eq:mixture_intro}.

\textbf{Approximating with a product mixture.} Our approach is inspired by first-order frequentist methods for solving sparse linear regression \eqref{eq:slr_model}, when $\mx$ is RIP. This condition means that for bounded-size subsets $S \subseteq [d]$, we have $[\mx^\top \mx]_{S \times S} \approx \id_S$ is an approximate isometry, so $\ma_S$ in \eqref{eq:mixture_intro} is close to a multiple of the identity (see Definition~\ref{def:RIP} for a formal statement). Many sparse recovery methods simulate well-conditioned gradient descent, a method for optimizing convex functions, on the naturally nonconvex sparse linear regression objective by enforcing sparsity of iterates \cite{BlumensathD09, NeedellT09} or by projecting onto convex proxy sets for sparsity \cite{AgarwalNW10}. Thus, a natural approach to sampling from \eqref{eq:bslr_intro}, at least restricted to components $S \subseteq [d]$ with small $|S|$, is to approximate each corresponding Gaussian in \eqref{eq:mixture_intro} with an isotropic Gaussian (i.e., a product density).

Fortunately, there is precedent: for the closely-related problem of sampling from the \emph{sparse mean model}, where we receive observations of the form \eqref{eq:slr_model} but $\mx = \id_d$,  \cite{castillo2012needles} provided a simple dynamic programming-based algorithm for drawing a candidate subset $S \sim w$. The key property enabling the \cite{castillo2012needles} algorithm is that both the prior and mixture components are product distributions. Thus, our starting point is to approximate \eqref{eq:slr_model} with a sparse mean model.

\textbf{Denoising the observations.} One candidate for simulating the sparse mean model is 
\[\mx^\top \vy = \mx^\top \mx \vths + \mx^\top \vxi \approx \vths + \mx^\top \vxi,\]
where the above approximation holds due to RIP, assuming the signal $\vths$ is sparse. At least for the natural setting of small $\sig$, this calculation suggests $\mx^\top \vy$ (which we can recover from our observations) is a good estimate of $\vths$. Unfortunately, the coordinates of $\mx^\top \vy$ are highly non-uniform in magnitude, due to randomness inherent in the diffuse density $\mu$. This causes difficulties when setting up a rejection sampling scheme for sampling from $w$ in \eqref{eq:mixture_intro}, because the estimate
\[\exp\Par{\half \norm{\vb_S}_{\ma_S^{-1}}^2} \approx \exp\Par{\frac {\sig^2}{2(1 + \sig^2 )} \norm{\vb_S}_2^2}, \text{ where } \vb \defeq \frac 1 {\sig^2}\mx^\top \vy \]
does not hold, even though $\ma_S \approx (\frac {1 + \sig^2} {\sig^2}) \id_S$ is true for small $S$. This is simply because the scale of $\vb_S$ is too large, due to the aforementioned nonuniformity in $\vths \sim \pi$.

We make the crucial observation that we can ``denoise'' the proportionality constants in \eqref{eq:mixture_intro}, using a vector $\vhth \in \R^d$ that meets a certain criterion, as follows. Let $\vhth \in \R^d$ have $T \defeq \supp(\vhth)$, and let $T \subseteq S, S' \subseteq [d]$. Then, a straightforward calculation (Lemma~\ref{lem:recenter}) shows that
\begin{equation}\label{eq:recenter_intro}\frac{\exp\Par{\half \norm{\vb_S}_{\ma_S^{-1}}^2}}{\exp\Par{\half \norm{\vb_{S^\prime}}_{\ma_{S^\prime}^{-1}}^2}} = \frac{\exp\Par{\half \norm{\vz_S}_{\ma_S^{-1}}^2}}{\exp\Par{\half \norm{\vz_{S^\prime}}_{\ma_{S^\prime}^{-1}}^2}}, \text{ where } \vz \defeq \frac 1 {\sig^2}\Par{\mx^\top \vy - \mx^\top\mx \vhth} - \vhth.\end{equation}
Thus, as long as almost all high-weight $S$ in \eqref{eq:mixture_intro} are supersets of some $T \subseteq [d]$, we have freedom in choosing $\vhth \in \R^d$ supported in $T$, to shrink the scale of our proportionality constants using \eqref{eq:recenter_intro}. 

\textbf{Constructing a posterior estimator.} The next step in our plan is to construct an estimate $\vhth(\mx, \vy)$, for which we can show $\supp(\vth) \supseteq \supp(\vhth)$ with high probability over $\vth \sim \pi(\cdot \mid \mx, \vy)$.

We propose to use $\vhth$ obtained using sparse recovery with $\ell_\infty$ guarantees. To draw a connection between $\vhth \approx \vths$ and $\vhth \approx \vth \sim \pi(\cdot \mid \mx, \vy)$, we extend a powerful result of \cite{JalalADPDT21} (Proposition~\ref{prop:est_to_sample}) that characterizes posterior sampling as an estimation tool. 
Specifically, it shows that posterior sampling achieves estimation guarantees within a constant factor of \emph{any other estimator} in any metric, that is learned from noisy measurements of a signal. Due to the existence of sparse recovery guarantees in $\ell_\infty$, we show that thresholding the coordinates of a frequentist estimate $\vhth$ yields a sufficient $T$ for our framework. This can be intuitively viewed as learning the ``obvious'' coordinates of $\vths$ above the noise level $\sig$, which almost all posterior samples should also contain.

After applying this denoising, we are able to show that our new proportionality constants in \eqref{eq:recenter_intro} are much more closely approximated by a sparse mean model, for small $S \subseteq [d]$. However, we are left to prove that $\pi(\cdot \mid \mx, \vy)$ is mainly supported on small $S \subseteq [d]$. While this is an intuitive property (as the prior is sparse), we were previously only aware of proofs in specific cases. For example, \cite{castillo2015bayesian} showed sparsity of the spike-and-slab posterior when $\mu = \Lap(0, 1)$, but their proof technique required the negative log-density to be a norm, and hence did not apply to $\mu = \Nor(0, 1)$. By a very simple application of Proposition~\ref{prop:est_to_sample} (cf.\ Corollary~\ref{cor:post_sparse}), we prove such a sparsity concentration result for all densities $\mu$, which is potentially of independent interest.

Our final sampler first learns a frequentist estimate $\vhth$ with large coordinates $T$, and draws $S \subseteq [d]$ from a product distribution restricted to $S \supseteq T$ and $|S| \lesssim k$. It then applies a denoised rejection sampling step to correct this proposal. Our proposal distribution is sampled via \emph{conditional Poisson sampling} (Lemma~\ref{lemma:conditional_sampling}), based on a dynamic programming strategy. Interestingly, the product distribution inducing our proposal is actually not concentrated on small subsets, so it only approximates our posterior \eqref{eq:mixture_intro} over a high-probability region of $\supp(\vth)$ where $\vth \sim \pi(\cdot \mid \mx, \vy)$. 

Intuitively, our algorithm reflects uncertainty in the denoised sparse linear regression problem: we do not know the location of the remaining small signal coordinates, but we know they are few.

\textbf{Extending to a Laplace prior.} We now illustrate how to extend our findings to the Laplace prior setting (Theorem~\ref{thm:laplace_informal}). Unlike the Gaussian case, the $\ell_{1}$ norm arising from the Laplace prior precludes a closed-form expression for the density over the posterior support. Consequently, we rely on the triangle inequality to perform approximations, and subsequently establish in Lemma~\ref{lem:rejection_rate_bound_laplace} that for noise level $\sigma = O\bigl(\frac{1}{k}\bigr)$, it is still possible to construct an approximate density that is both efficient to sample from and closely approximates the true posterior. A further challenge in designing an efficient algorithm is computing the rejection ratio, again due to the lack of  a closed-form expression for the posterior density. To overcome this, we use an annealing-based algorithm (Proposition~\ref{prop:est_Z}) combined with an approximate sampler for composite densities developed in \cite{LeeST21} to estimate normalizing constants in polynomial time, enabling us to bypass the need for an exact closed-form density while still ensuring efficient and accurate sampling from the posterior.

\subsection{Related work}\label{ssec:related}

Over the years, there has been a huge amount of work on high dimensional sparse regression in both the Bayesian and frequentist communities~\cite{Mon_MitchellBeauchamp1988,Mon_GeorgeMcCulloch1993,Mon_EfronHastieJohnstoneTibshirani2004,Mon_Wainwright2009,Mon_BertsimasVanParys2020,montanari2024provably}.
Under the Bayesian paradigm, some works focus on finding the mode of the posterior distribution in~\eqref{eq:bslr_intro}~\cite{rockova2014emvs,rockova2018spike}, but such an estimate ignores the uncertainty of the posterior. Directly sampling from the posterior, on the other hand, could address such drawbacks. While there is an active line of research in this direction~\cite{Bai_bhattacharya2016fast,Bai_narisetty2019skinny,Bai_johndrow2020scalable}, computationally efficient algorithms with rigorous theoretical guarantees have been few, due to the key challenge that the posterior resulting from the spike-and-slab prior is high-dimensional and may be very far from unimodal. 

\textbf{MCMC-based methods.} One line of work on posterior sampling exploits MCMC-based methods~\cite{MC_BC09,MC_RBR10,MC_SFLCM15,MC_YWJ16}. 
Motivated by the Bernstein-von Mises theorem, ~\cite{MC_BC09} assume that the posterior is close to a normal density and design a sampler based on this assumption. Closer in spirit to our work,~\cite{MC_YWJ16} assumes their measurement matrix $\mx$ satisfies a condition similar to RIP, and shows that the posterior on the set of ``influential coordinates'' (i.e., those larger than the noise level) is unimodal, if the SNR is very high or very low. However, their work does not capture a broad range of intermediate SNRs (see discussion after Eq.\ (10)), for which the posterior can display multimodal behavior. Moreover, Theorem 6 of~\cite{castillo2015bayesian} shows that under some mild conditions, the posterior converges to a \textit{mixture of Gaussians}, and such a mixture will collapse into a single Gaussian only if the signal dominates the noise. Such a result uncovers a limitation of current MCMC-based methods: their conclusions are based on the premise that posterior is approximately a unimodal distribution, which reduces the difficulty of mixing time analyses. 

Finally, recent work \cite{BrunaH24} studies the use of diffusion methods for posterior sampling in linear inverse problems. We think this is an interesting approach that merits further investigation, e.g., for our specific case of spike-and-slab posterior sampling. However, the aforementioned result is not analyzed in discrete time, and its techniques do not readily apply to our setting; in particular, their parameter $\chi_t(\pi)$ in Eq.\ (14) is unbounded for the spike-and-slab prior $\pi$.

\textbf{Approximating the posterior.} Another line of work on posterior sampling uses variational Bayesian methods, which obtain the closest approximation of a complex posterior distribution from a more tractable family~\cite{VB_JGJS99,VB_WJ08,VB_BKM17}. The most popular approximating family consists of product distributions, and the corresponding approximation is known as a na\"ive mean-field approximation. For example,~\cite{VB_RS22} and~\cite{VB_MS22} show that the mean-field approximation has strong theoretical properties. However, the mean-field optimization problem has been shown to be highly nonconvex with spurious local optima in some simple learning settings~\cite{VB_MS22}. Moreover,~\cite{VB_Ghorbani2018AnII} show the failure of na\"ive mean-field for a simple high-dimensional problem. In some sense, our paper also obtains a variational approximation by sampling from an appropriate product distribution. However, instead of choosing the ``best'' candidate (requiring nonconvex optimization), we choose one guided by fine approximations of the true posterior and the RIP condition.

\textbf{Moderate dimension.} Recently,~\cite{montanari2024provably} gave a computationally-efficient posterior sampler in a setting where the posterior can be highly multimodal. Their algorithm is motivated by decompositions in statistical physics, and introduces a latent variable that is easily sampleable and induces the correct marginal distribution on the parameters $\vth$. However, \cite{montanari2024provably} assume that their design matrix $\mx$ has a number of linear measurements which grows at least linearly in the feature dimension $d$. This can be a costly assumption in practical high-dimensional settings.

In our paper, we remove the aforementioned constraints, i.e., we operate in the \textit{high-dimensional regime} of $n = o(d)$ and under a range of SNRs $\sig^{-1}$ where the posterior is potentially \textit{multimodal}.

\section{Preliminaries}\label{sec:prelims}

\subsection{Notation}\label{ssec:notation}

\textbf{General notation.} We denote matrices in capital boldface and vectors in lowercase boldface. We define $[n] \defeq \{i \in \N \mid i \le n\}$. When $S$ is a subset of a larger set clear from context (e.g., $[d]$), we let $S^c$ denote its complement. We let $\vzero_d$ and $\vone_d$ denote the all-zeroes and all-ones vectors in $\R^d$.

\textbf{Matrices.} We let $\id_d$ denote the $d \times d$ identity matrix, and $\id_S$ is the identity on $\R^S$ for an index set $S$. We use $\preceq$ to denote the Loewner partial order on the cone of positive semidefinite (PSD) matrices (denoted $\PSD^{d \times d}$ when in $d$ dimensions), a subset of the symmetric matrices (analogously, denoted $\Sym^{d \times d}$). When $\mm \in \PSD^{d \times d}$ is non-singular, we define its induced norm by $\norm{\vv}_{\mm}^2 \defeq \vv^\top \mm \vv$.

For $p \ge 1$ (including $p = \infty$), applied to a vector argument, $\norm{\cdot}_p$ denotes the $\ell_p$ norm. For $p, q \ge 1$ and a matrix $\mm \in \R^{n \times d}$, we also use the notation
$\norm{\mm}_{p \to q} \defeq \max_{\substack{\vv \in \R^d \mid \norm{\vv}_p \le 1}} \norm{\mm \vv}_q$.

We use $\normf{\cdot}$ and $\normop{\cdot}$ to denote the Frobenius and ($2 \to 2$) operator norms of a matrix argument, and for $\mm \in \Sym^{d \times d}$, $\vlam(\mm) \in \R^d$ denotes its vector of eigenvalues sorted in nondecreasing order, i.e., $\vlam_1(\mm) \ge \vlam_2(\mm) \ge \ldots \ge \vlam_d(\mm)$. We also use $\vlam_{\max}(\mm)$ and $\vlam_{\min} (\mm)$ to denote the maximal and minimal eigenvalues of $\mm \in \Sym^{d \times d}$. For an asymmetric rank-$r$ $\mm$, we let $\vsig(\mm) \in \R^r_{> 0}$ denote its singular values, and define $\vsig_{\min}$, $\vsig_{\max}$ similarly. For $\mm \in \Sym^{d \times d}$, we let $\mm^\dagger$ denote its pseudoinverse, i.e., the matrix in $\Sym^{d \times d}$ satisfying $\mm \mm^\dagger = \mm^\dagger \mm$ is the projection matrix onto the span of $\mm$.

We let $\omega < 2.373$ \cite{AlmanDWXXZ25} denote the matrix multiplication exponent, and assume multiplying, inverting \cite{Strassen69}, and computing eigendecompositions \cite{PanC99} of $d \times d$ matrices takes $O(d^\omega)$ time.

\textbf{Probability.} For $p \in (0, 1)$, we let $\Bern(p)$ denote the Bernoulli distribution (over $\{0, 1\}$) with mean $p$. We use $\Nor(\vmu, \msig)$ to denote the multivariate normal distribution with specified mean and covariance. For $t \in \R$ we let $\delta_t$ denote the Dirac (generalized) density at $t$, i.e., $x \sim \delta_t \implies x = t$ with probability 1. We let $\Lap(\mu, b)$ be the Laplace distribution with mean $\mu$ and scale $b$, so $\Lap(0, 1)$ is the density on $\R$ that is $\propto \exp(-|\cdot|)$. For an event $\event$, $\ind_\event$ denotes its $0$-$1$ indicator variable. We denote the total variation distance between two distributions $\mu, \nu$ by $\tv{\mu, \nu}$.

For a density $\mu$ over $\R$, we denote $\mu^{\otimes S}$ to denote the product density over $\R^S$ whose coordinates are i.i.d.\ $\sim \mu$. 
We use $\mu^{\otimes d}$ to denote the case when $S = [d]$. More generally, we use $\bigotimes_{i \in [d]} \mu_i$ to mean the product density whose $i^{\text{th}}$ coordinate is distributed $\sim \mu_i$.

For simplicity throughout the paper, we assume that all one-dimensional integration and sampling takes $O(1)$ time. All applications of this assumption involve reasonably well-behaved functions, and we expect this to be a fairly realistic assumption in practice.

\textbf{Indexing.} For a vector $\vv \in \R^d$ and $S \subseteq [d]$, we use $\vv_S \in \R^S$ to denote its restriction to its coordinates in $S$. We let $\nnz(\cdot)$ denote the number of nonzero entries in a matrix or vector argument. We let $\supp(\vv)$ denote the subset of nonzero entries in $\vv$ (i.e., its support). For $\mm \in \R^{n \times d}$, we use $\mm_{i:}$ to denote its $i^{\text{th}}$ row for $i \in [n]$, and $\mm_{:j}$ to denote its $j^{\text{th}}$ column for $j \in [d]$. When $S \subseteq [n]$ and $T \subseteq [d]$ are row and column indices, we let $\mm_{S \times T}$ be the submatrix indexed by $S$ and $T$; if $T = [d]$ we simply denote this submatrix as $\mm_{S:}$ and similarly, we define $\mm_{:T}$. We fix the convention that transposition is done prior to indexing, i.e., $\mm_{S\times T}^\top \defeq [\mm^\top]_{S\times T}$.

\subsection{Spike-and-slab posterior sampling}\label{ssec:sas_post}

In this section, we formally define our problem. We begin by defining the \emph{spike-and-slab} density, originally introduced by \cite{Mon_MitchellBeauchamp1988}. The spike-and-slab density is a simple product density that is parameterized by $\vq \in [0, 1]^d$ and a \emph{diffuse density} $\mu$ over $\R$, given as follows:
\begin{equation}\label{eq:sas_prior}
\pi \defeq \bigotimes_{i \in [d]} \Par{(1 - \vq_i) \delta_0 + \vq_i \mu}.
\end{equation}

In other words, the $i^{\text{th}}$ coordinate of $\vth \sim \pi$ is nonzero with probability $\vq_i$, for all $i \in [d]$.  Finally, if $\mu$ is unspecified we assume $\mu = \Nor(0, 1)$ is a standard Gaussian.

There is an alternative characterization of the distribution in \eqref{eq:sas_prior} which follows by first describing the distribution $\pisupp$ of the support of $\vth \sim \pi$, and then sampling $\vth$ from the conditional distribution (i.e., via Bayes' theorem). We describe this alternative characterization as follows.

\begin{enumerate}
    \item First, a support $S \subseteq [d]$ is sampled from $\pi_{\supp}$, where 
    \begin{equation}\label{eq:pisupp}\pisupp(S) = \Par{\prod_{i \in S} \vq_i} \Par{\prod_{i \in S^c} (1 - \vq_i)}\text{ for all } S \subseteq [d].\end{equation}
    \item Second, $\vths_{S^c}$ is set to $\vzero_{S^c}$, and $\vths_S$ is sampled from
    \begin{equation}\label{eq:spike_def}\vths_S \sim \mu^{\otimes S}.\end{equation}
\end{enumerate}

We now define our posterior sampling problem.

\begin{model}[Spike-and-slab posterior sampling]\label{model:sas_basic}
Let $\vq \in [0, 1]^d$, $n, d \in \N$, $k = \onenorm{\vq}$, and $\sigma > 0$ be known, and let $\mu = \Nor(0, 1)$. Let $\mx \in \R^{n \times d}$ be a known measurement matrix, and suppose that we observe $(\mx, \vy)$ where, following the notation \eqref{eq:sas_prior}, $\vy$ is generated via
\begin{equation}\label{eq:slr}\vths \sim \pi,\quad \vxi \sim \Nor(\vzero_d, \sigma^2 \id_d),\quad\vy = \mx \vths + \vxi,\end{equation}
and $\vth^\star$ and $\vxi$ are independent. Our goal is to sample from the posterior $\pi(\cdot \mid \mx, y)$.
\end{model}

That is, $\vy$ in \eqref{eq:slr} follows a standard noisy linear model, so our posterior task is a Bayesian linear regression problem with spike-and-slab prior \eqref{eq:sas_prior}. In light of our earlier viewpoint \eqref{eq:pisupp}, \eqref{eq:spike_def} of sampling from $\pi$, there is similarly a two-stage process that characterizes sampling $\vth \sim \pi(\cdot \mid \mx, \vy)$. We now provide an explicit description of this process in the case when $\mu = \Nor(0, 1)$.

\begin{enumerate}
    \item First, a support $S \subseteq [d]$ is sampled from $\pisupp(\cdot \mid \mx, \vy)$,
    where
    \begin{equation}\label{eq:pisupp_post}\pisupp(S \mid \mx, \vy) \propto \frac{(\prod_{i \in S} \vq_i)(\prod_{i \in S^c} (1 - \vq_i))}{ (2\pi)^{\frac{|S|}{2}} }\int_{\R^S} \exp\Par{-\frac{1}{2\sigma^2}\norm{\vy - \mx \vth_S}_2^2-\half\norm{\vth_S}_2^2} \dd \vth_S,
    \end{equation}
    and the integral extends each $\vth_S \in \R^S$ to $\vth \in \R^d$ via padding by zeroes.
    \item Second, $\vth \in \R^d$ is sampled from $\pi(\cdot \mid \mx, \vy, S)$, where
    \begin{equation}\label{eq:thsupp_post}
    \pi(\vth \mid \mx, \vy, S) \propto \exp\Par{-\frac{1}{2\sigma^2}\norm{\vy - \mx \vth}_2^2 - \half \norm{\vth}_2^2} \cdot \ind_{\supp(\vth) \subseteq S} .
    \end{equation}
\end{enumerate}
In \eqref{eq:pisupp_post}, we used the following standard calculation to compute the normalizing constants of the relevant multivariate Gaussian densities, which will be used several times throughout.

\begin{fact}[Gaussian normalization]
\label{fact:gauss_norm}
Let $\msig \in \PSD^{d \times d}$ be non-singular and let $\vmu \in \R^d$. Then,
\begin{equation}\label{eq:gaussian_integral}\int \exp\Par{-\half \vth^\top \msig^{-1} \vth + \vth^\top \vmu} \dd \vth = (2\pi)^{\frac d 2}\exp\Par{\half\norm{\vmu}_{\msig}^2}\det(\msig)^{\half}.\end{equation}
\end{fact}

\subsection{Sparse recovery}\label{ssec:sparse_recovery}

In this section, we state several preliminaries on noisy sparse recovery in the following model. In particular, Model~\ref{model:sparse_recovery} differs from Model~\ref{model:sas_basic} primarily in that $\vhth$ is not assumed to be drawn from a fixed prior distribution, but is instead a ``worst-case'' sparse signal to be estimated. We defer all proofs from this section to Appendix~\ref{app:sparse}, as most are well-known in the literature.

\begin{model}[Sparse recovery]
\label{model:sparse_recovery}
Let $n, d \in \N$, $k \in [1, d]$, and $\sigma > 0$ be known. Let $\mx \in \R^{n \times d}$ be a known measurement matrix, and let $\vths \in \R^d$, $\vxi \in \R^n$ satisfy $\nnz(\vths) \le k$. Suppose we observe $(\mx, \vy)$ where
\begin{equation*}
    \vy = \mx \vths + \vxi, \quad \vxi \sim \calN(\vzero_n, \sigma^2\id_n).
\end{equation*}
Our goal is to output an estimate of $\vths$.
\end{model}

For a survey of techniques and known results in sparse recovery, we refer the reader to Chapter 7 of \cite{wainwright2019high} and references therein. We provide two recovery results under Model~\ref{model:sparse_recovery} under assumptions on the pair $(\mx, \vxi)$. To state our assumptions, we require the following standard definition.

\begin{definition}[Restricted isometry property]
    \label{def:RIP}
    We say $\mx \in \bbR^{n\times d}$ satisfies the $(\epsilon, s)$-restricted isometry property, or $\mx$ is $(\epsilon, s)$-$\RIP$, if for all $\vth \in \bbR^d$ with $\nnz(\vth) \leq s$, 
    \begin{equation*}
        (1 - \epsilon) \twonorm{\vth}^2 \leq \twonorm{\mx\vth}^2 \leq (1 + \epsilon) \twonorm{\vth}^2.
    \end{equation*}
    An equivalent condition is that $\vlam([\mx^\top \mx]_{S \times S}) \in [1 - \eps, 1 + \eps]^s$ for all $S \subseteq [d]$ with $|S| \le s$.
\end{definition}
Intuitively, RIP implies $\mx$ acts as an
approximate isometry when restricted to sparse vectors, so $[\mx^\top \mx]_{S\times S}$ is well-conditioned for any sparse support $S$. 
Various random matrix ensembles have been shown to satisfy Definition~\ref{def:RIP}; we recall two such standard constructions below.

\begin{proposition}[Theorem 9.2, \cite{foucart13}]
    \label{prop:RIP_sub_gaussian}
    Let $\mm$ be an $n\times d$ random matrix, where entries of $\mm$ are independent mean-zero sub-Gaussian random variables with variance $1$.
    There exists a constant $C > 0$ such that for any $s \in [d]$, $\mx = \frac{1}{\sqrt{n}}\mm$ is $(\epsilon, s)$-RIP with probability $\ge 1 - \delta$ if
    \begin{equation*}
        n \geq C\cdot \frac{s\ln\frac{d}{s} + \log\frac{1}{\delta}}{\epsilon^2}.
    \end{equation*}
\end{proposition}

\begin{proposition}[Theorem 4.5, \cite{haviv2017restricted}]
    \label{prop:RIP_SDFT}
    Let $\mm \in \bbC^{d\times d}$ be a discrete Fourier matrix whose elements are given by 
    $\mm_{jk} = d^{-1/2} \cdot \exp(2\pi \iota \cdot \frac{jk}{d})$ where $\iota \defeq \sqrt{-1}$. Let $\mx \in \bbC^{n\times d}$ be a matrix whose $d$ rows are chosen uniformly and independently from the rows of $\mm$ multiplied by $\sqrt{d/n}$. 
    There exists a constant $C > 0$ such that for any $s \in [d]$, $\mx$ is $(\epsilon, s)$-RIP with probability $\ge 1 - 2^{-\Omega(\log d \log (s/\epsilon))}$ if 
    \begin{equation*}
        n \geq C \cdot \frac{s\cdot \log d \cdot \log^2(\frac{s}{\epsilon})\cdot \log^2(\frac{1}{\epsilon})}{\epsilon^2}.
    \end{equation*}
\end{proposition}

For more constructions of RIP matrix ensembles, including sampling bounded orthonormal systems and real trignometric polynomials, we refer the reader to Chapter 12 of \cite{foucart13}.

We next state the main sparse recovery results used in our samplers. These sparse recovery algorithms require some mild additional technical conditions for runtime bounds, which are satisfied by all of the constructions in Propositions~\ref{prop:RIP_sub_gaussian} and~\ref{prop:RIP_SDFT} with high probability, after minor modifications. 

\begin{assumption}[Assumptions for $\ell_\infty$ sparse recovery]\label{assume:linf}
In the setting of Model~\ref{model:sparse_recovery}, suppose that $\mx \in \R^{n \times d}$ is $(\frac 1 {4k}, k + 1)$-RIP, that the columns of $\mx$ are in general position (i.e., for any $r \in [n]$ and $S \subseteq [d]$ with $|S| = r$, $\mx_{:S}$ is full-rank), and $\frac{\vsig_1(\mx)}{\vsig_n(\mx)} \le \kappa$.
\end{assumption}

In our applications of our framework (cf.\ Section~\ref{ssec:gaussian_proof}), we provide explicit estimates of $\kappa$ which hold with high probability. The assumption that $\mx$ has columns in general position holds with probability $1$ if $\mx$'s entries are drawn from continuous probability densities (measurable w.r.t.\ Lebesgue), as was observed in e.g., \cite{tibshirani2013lasso}. In other settings, we can enforce this general position assumption by adding infinitesimal noise to $\mx$'s entries, and absorb the error under Model~\ref{model:sparse_recovery} into $\vxi$. 

\begin{assumption}[Assumptions for $\ell_2$ sparse recovery]\label{assume:l2}
In the setting of Model~\ref{model:sparse_recovery}, suppose that for some $m \in [n]$, $\sqrt{m/n} \cdot \mx_{[m]:}$ is $(\frac 1 {10}, C_2 k)$-RIP for a universal constant $C_2$. 
\end{assumption}

We include Assumption~\ref{assume:l2} for our sampling applications, which use $\mx$ satisfying $(\eps, O(k))$-RIP for some $\eps \ll 1$. Since our $\ell_2$ sparse recovery result only requires a constant $\eps$ parameter, but incurs error proportional to the noise level (scaling with the number of rows used), Assumption~\ref{assume:l2} (satisfied by all our RIP matrix ensembles) sharpens our measurement complexity's dependence on $k$.

We defer a proof of Proposition~\ref{prop:sparse_recovery} to Appendices~\ref{app:L_infty_estimator} and~\ref{app:L_2_estimator}.

\begin{proposition}\label{prop:sparse_recovery}
In the setting of Model~\ref{model:sparse_recovery}, the following hold.
\begin{enumerate}
    \item Under Assumption~\ref{assume:linf}, there is an algorithm $\alg_\infty$ that returns $\vhth$ satisfying
    \[\norm{\vhth - \vths}_\infty \le C_\infty r_\infty, \text{ for any } r_\infty \ge \norm{\mx^\top \vxi}_\infty\]
    and for a universal constant $C_\infty$. Moreover, $\alg_\infty$ runs in time
    \[O\Par{d^{1.5}n^2\log\Par{d\kappa\cdot\Par{1 + \frac{R_\infty}{r_\infty}}}}, \text{ for any } R_\infty \ge \norm{\mx^\top \vy}_\infty.\]
    \item Under Assumption~\ref{assume:l2}, there is an algorithm $\alg_2$ that returns $\vhth$ satisfying
    \[\norm{\vhth - \vths}_2 \le C_2 r_2,\text{ for any } r_2 \ge \norm{\vxi_{[m]}}_2.\]
    Moreover, $\alg_2$ runs in time
    \[O\Par{nd\log\Par{\frac{R_2}{r_2}}}, \text{ for any } R_2 \ge \norm{\vths}_2.\]
\end{enumerate}
\end{proposition}

Finally, we give a helper tool bounding quantities appearing in Proposition~\ref{prop:sparse_recovery} for our model.

\begin{restatable}{lemma}{restatemaxrowip}\label{lem:maxrow_ip}
Let $\mx \in \R^{n \times d}$ satisfy $(\epsilon, 1)$-RIP 
and $\vxi \sim \Nor(\vzero_n, \sigma^2\id_n)$. Then 
\[\norm{\mx^\top \vxi}_\infty \le \sigma(1 + \epsilon) \sqrt{2\log\Par{\frac d \delta}}, \text{ with probability } \ge 1 - \delta, \text{ for all } \delta \in (0, 1).\]
\end{restatable}

\subsection{Performance of posterior sampling via estimation}
\label{ssec:posterior_estimate}

We next introduce a key technical tool from that shows posterior sampling performs nearly as well at estimation as \emph{any other procedure}, in a precise sense. To our knowledge, this fact was first stated in \cite{JalalADPDT21}. We use this in several places to prove properties of our posterior distributions. Here, we state a slight generalization of the result from \cite{JalalADPDT21} and defer a proof to Appendix~\ref{app:posterior_estimate}.

\begin{restatable}[Generalization of Theorem 3.4, \cite{JalalADPDT21}]{proposition}{restateesttosamp}\label{prop:est_to_sample}
Let $\met(\cdot, \cdot)$ be an arbitrary metric over $\calK \times \calK$ and suppose $\mu$ is a distribution over $\calK$. Let $\theta^\star \sim \mu$, let $\calF: \calK \to \Omega$ be an arbitrary (possibly randomized) forward operator, and let $\varphi = \calF(\theta^\star)$. Suppose there is any (possibly randomized) algorithm $\calA: \Omega \to \calK$ such that for $\eps > 0$, $\delta \in (0, 1)$,
\begin{equation}\label{eq:assume_estimator}\Pr\Brack{\met(\hth, \theta^\star) > \eps} \le \delta, \text{ for } \hth \sim \alg(\varphi), \end{equation}
where probabilities are taken over the joint distribution of $(\theta^\star, \varphi, \hth)$.
Then, letting $\theta \sim \mu(\cdot \mid \varphi)$, i.e., the posterior distribution of $\theta^\star$ given observations $\varphi$, we have
\[\Pr\Brack{\met(\theta, \theta^\star) > 2\eps} \le 2\delta.\]
\end{restatable}

To demonstrate the power of Proposition~\ref{prop:est_to_sample}, we prove a support size concentration result for spike-and-slab posteriors. A similar result was shown in Theorem 1, \cite{castillo2015bayesian}, which did not give a quantitative convergence rate. Moreover, the strategy in \cite{castillo2015bayesian} is somewhat more ad hoc, as it was specialized to Laplace priors (for example, it relied on the negative log-density obeying the triangle inequality, which does not hold for e.g., the Gaussian prior). In contrast, we prove a non-asymptotic result for arbitrary product priors, using an arguably simpler proof.

\begin{corollary}[Sparsity of posterior]\label{cor:post_sparse}
In the setting of Model~\ref{model:sas_basic}, let $\vth \sim \pi(\cdot \mid \mx, \vy)$. Then
\begin{equation}\label{eq:post_sparse}
    \bbP\Brack{\Abs{\supp(\vth)} \leq 6\Par{k + \log\Par{\frac 3 \delta}}} \geq 1 - \delta, \text{ for all } \delta \in (0, 1),
\end{equation}
Moreover, \eqref{eq:post_sparse} holds for any choice of $\mu$ in Model~\ref{model:sas_basic}.
\end{corollary}
\begin{proof}
We apply Proposition~\ref{prop:est_to_sample} to the distribution $\pisupp(\cdot \mid \mx, \vy)$ defined in \eqref{eq:pisupp}, \eqref{eq:pisupp_post}. Let $\calK$ be the set of subsets of $[d]$, and let $\vhi = (\mx, \vy)$ (so $\Omega = \R^{n \times d} \times \R^n$). Moreover, let $\met(S_1, S_2)$ be the Hamming distance between $S_1, S_2 \subseteq [d]$, i.e.,
\[
        \met(S_1, S_2) = \Abs{\Brace{i \in [d] \mid  \Par{i \in S_1 \wedge i \not\in S_2} \vee \Par{i \in S_2 \wedge i \not\in S_1}}}.\]
    It is well-known that $\met(\cdot, \cdot)$ is a metric on $\calK$. We let $\alg(\vhi)$ always return the empty set $\emptyset$. By a Chernoff bound, letting $S^\star \sim \pisupp$, since $|S^\star|$ is the sum of variables $\sim \textup{Bern}(\vq_i)$ for all $i \in [d]$,
    \[
    \Pr\Brack{|S^\star| \ge 2\Par{k + \log\Par{\frac 3 \delta}}} =
    \Pr\Brack{\met(S^\star, \alg(\vhi)) \ge 2\Par{k + \log\Par{\frac 3 \delta}}} \le \frac \delta 3.
    \]
    Thus, \eqref{eq:assume_estimator} holds with $\eps = 2(k + \log(\frac 3 \delta))$ so that, for $S \sim \pisupp(\cdot \mid \mx, \vy)$,
    \[\Pr\Brack{\met\Par{S, S^\star} \ge 4\Par{k + \log\Par{\frac 3 \delta}}} \le \frac{2\delta}{3}.\]

    Finally, by the triangle inequality on $\met(\cdot, \cdot)$, 
    \begin{equation*}
        |S| = \met(S, \calA(\vhi)) \leq \met(S, S^\star) + \met(S^\star, \calA(\vhi)) = \met(S, S^\star) + |S^\star|. 
    \end{equation*}
    Hence, we have
    \begin{align*}
        \bbP\bra{|S| \leq 6 \paren{k + \log\paren{\frac 3 \delta}}} &\geq
        \bbP\bra{|S^\star| + m(S, S^\star) \leq 6 \paren{k + \log\paren{\frac 3 \delta}}} \\
        &\geq \bbP\bra{|S^\star|\leq 2\paren{k + \log\paren{\frac 3 \delta}} \land 
        m(S, S^\star )\leq 4\paren{k + \log\paren{\frac 3 \delta}}}.
    \end{align*}
    The conclusion follows from a union bound over the above two events.
\end{proof}

Proposition~\ref{prop:est_to_sample}'s conclusion holds over the randomness of both $(\theta^\star, \vhi)$ (collectively, defining the ``model'' we are given in the posterior sampling problem) and the sample $\theta \sim \mu(\cdot \mid \vhi)$. This guarantee is not quite compatible with our applications, as we would only like to reason about the failure probability of our sampler for a given posterior distribution, holding the model fixed. 

We give a simple reduction showing that an event which holds with very high probability over the randomness of $(\theta^\star, \vhi, \theta)$ also holds with high probability over $\theta \sim \mu(\cdot \mid \vhi)$, for ``most'' models $(\theta^\star, \vhi)$. 

\begin{restatable}{lemma}{restategoodmodels}\label{lem:most_models_good}
Let $(\alpha, \beta)$ be random variables, and let $\event_{\alpha, \beta}$ be an event that depends on their realizations. Suppose for some $\delta_1, \delta_2 \in (0, 1)$ that $\Pr_{(\alpha, \beta)}[\event_{\alpha, \beta}] \ge 1 - \delta_1 \delta_2$. Then, letting $g(\alpha) \defeq \Pr_{\beta}\Brack{\event_{\alpha, \beta} \mid \alpha}$,
\[\Pr_\alpha[g(\alpha) \le 1 - \delta_1] \le \delta_2.\]
\end{restatable}

Our final theorem statements use Lemma~\ref{lem:most_models_good} by applying it with $\alpha \gets (\vths, \vxi)$ (i.e., the randomness used to generate observations $\vy$ in Model~\ref{model:sas_basic}), and $\beta \gets \vth \sim \mu(\cdot \mid \mx, \vy)$. We refer to this usage by qualifying that statements hold ``with probability $\ge 1 - \delta_1$ over the randomness of Model~\ref{model:sas_basic}.''

\subsection{Rejection sampling}
\label{ssec:rejection_sampling}
Our posterior sampler is based on \emph{approximate rejection sampling}, i.e., sampling a (high-probability) restriction of the target posterior distribution which has bounded relative density compared to an easier-to-sample conditional product distribution. We provide the following standard result on approximate rejection sampling, adapted from \cite{LeeST21} and proven in Appendix~\ref{app:reject}.

\begin{restatable}{lemma}{rejectionsampling}\label{lem:reject}
Let $\pi$, $\mu$ be distributions over the same domain $\Omega$, and suppose that $\pi \propto P$ and $\mu \propto Q$ for unnormalized densities $P, Q$. Moreover, suppose that for all $\omega \in \Omega$,
\begin{align*}
\frac{P(\omega)}{Q(\omega)} \le C, \; \frac{Q(\omega)}{P(\omega) } \le C.
\end{align*}
There is an algorithm $\rejectionsample(\mu, P, Q, \Omega, C, \delta)$ that, for any $\delta \in (0, 1)$, outputs a sample within total variation distance $\delta$ from $\pi$. The algorithm uses $O(C^2\log(\frac 1 \delta))$ samples from $\mu$, and evaluates $\frac{P(\omega)}{Q(\omega)}$ $O(C^2\log(\frac 1 \delta))$ times.
\end{restatable}

\subsection{Conditional Poisson sampling}
\label{ssec:conditional_poisson_sampling}

In this section, we state the guarantees on a conditional Poisson sampling algorithm (see \cite{grafstrom2005comparisons}), which will be useful subsequently. In short, this sampler gives a subset $S \subseteq [d]$ where indices are included as a product of Bernoulli indicators, conditioned on the event $|S| \le k$ for some specified $k \in [d]$. While we believe the dynamic programming-based solution we present is classical, we include a full proof in Appendix~\ref{app:cps_helper} along with pseudocode in Algorithm~\ref{alg:cond_sampling}.

\begin{restatable}{lemma}{restatecps}\label{lemma:conditional_sampling}
Let $\vp \in [0, 1]^d$, and let $\pi \defeq \bigotimes_{i \in [d]} \Bern(\vp_i)$ be a product distribution over $\{0, 1\}^d$, identified with sets $S \subseteq [d]$ in the canonical way. Define $\Omega_k \defeq \{S \subseteq [d] \mid |S| \le k \}$.  $\cps$ (Algorithm~\ref{alg:cond_sampling}) runs in time $O(dk)$ and returns $S \in \Omega_k$ such that $\Pr[S = T] = \pi(T \mid T \in \Omega_k )$.
\end{restatable}

Our final sampler uses Lemma~\ref{lemma:conditional_sampling} to sample from a base distribution $\mu$ in Lemma~\ref{lem:reject}.

\section{Spike-and-slab posterior sampling with Gaussian prior}\label{sec:gaussian}

In this section, we give our posterior sampler under Model~\ref{model:sas_basic} when $\mu = \Nor(0, 1)$, i.e., a Gaussian prior is used. To motivate our development, we make the following observation.

\begin{lemma}\label{lem:recenter}
In the setting of Model~\ref{model:sas_basic}, let
\[\ma \defeq \frac 1 {\sig^2} \mx^\top \mx + \id_d,\;\vb \defeq \frac 1 {\sig^2} \mx^\top \vy, \]
and denote $\ma_S \defeq \ma_{S \times S}$ for brevity. For some $\vhth \in \R^d$ with $T \defeq \supp(\vhth)$, let $\vz \defeq \vb - \ma \vhth$. Then,
\[\frac{\exp\Par{\half\norm{\vb_S}_{\ma_S^{-1}}^2}}{\exp\Par{\half\norm{\vb_{S'}}_{\ma_{S'}^{-1}}^2}} = \frac{\exp\Par{\half\norm{\vz_S}_{\ma_S^{-1}}^2}}{\exp\Par{\half\norm{\vz_{S'}}_{\ma_{S'}^{-1}}^2}}, \text{ for any } T \subseteq S, S' \subseteq [d].\]
\end{lemma}
\begin{proof}
First, observe that since $T \subseteq S, S'$,
\[\Brack{\ma \vhth}_S = \Brack{\Par{\frac 1 {\sig^2} \mx^\top \mx + \id_d} \vhth}_S = \ma_S \vhth_S,\]
and similarly $[\ma \vhth]_{S'} = \ma_{S'} \vhth_{S'}$. The result follows by directly expanding:
\begin{align*}
\half\norm{\vz_S}_{\ma_S^{-1}}^2 - \half\norm{\vz_{S'}}_{\ma_{S'}^{-1}}^2 &= \half\norm{\vb_S}_{\ma_S^{-1}}^2 - \vhth_S^\top \ma_S \ma_S^{-1} \vb_S + \half\norm{\ma_S \vhth_S}_{\ma_S^{-1}}^2 \\
&-\half\norm{\vb_{S'}}_{\ma_{S'}^{-1}}^2 + \vhth_{S'}^\top \ma_{S'} \ma_{S'}^{-1} \vb_{S'} - \half\norm{\ma_{S'} \vhth_{S'}}_{\ma_{S'}^{-1}}^2 \\
&= \half\norm{\vb_S}_{\ma_S^{-1}}^2 - \half\norm{\vb_{S'}}_{\ma_{S'}^{-1}}^2 - \vhth_S^\top \vb_S + \vhth_{S'}^\top \vb_{S'} \\
&+ \half \vhth_S^\top \ma_S \vhth_S - \half \vhth_{S'}^\top \ma_{S'} \vhth_{S'} = \half\norm{\vb_S}_{\ma_S^{-1}}^2 - \half\norm{\vb_{S'}}_{\ma_{S'}^{-1}}^2,
\end{align*}
where in the last equality, we used our support set assumption.
\end{proof}

To explain how Lemma~\ref{lem:recenter} is used, a direct calculation (combining \eqref{eq:pisupp_post} with Fact~\ref{fact:gauss_norm}, see Lemma~\ref{lem:unnorm_post}) shows that up to a term depending only on $\vq_S$, we have 
\[\pisupp(S \mid \mx, \vy) \propto \exp\Par{\frac{1}{2}\norm{\vb_S}_{\ma_S^{-1}}^2}.\]
We would like to sample from this distribution using rejection sampling (Lemma~\ref{lem:reject}) on top of a product distribution. However, some coordinates of $\mx^\top \vy = \mx^\top \mx \vths + \mx^\top \vxi \approx \vths + \mx^\top \vxi$ are potentially much larger than others, depending on the randomness of the model. This non-uniformity breaks straightforward applications of rejection sampling.

We instead use Lemma~\ref{lem:recenter} with an estimated ``hint'' vector $\vhth$ that has two properties: $\supp(\vhth) \subseteq S \sim \pi(\cdot \mid \mx, \vy)$ with high probability, and $\vz \defeq \vb - \ma \vhth$ is small in an appropriate norm. Lemma~\ref{lem:recenter} then lets us use a product distribution that depends on the size of $\vz$, rather than the less uniform $\vb$. This hint vector estimation effectively denoises our observations by first crudely estimating of the large coordinates of $\vths$ up to the noise level, via sparse recovery.

In Section~\ref{ssec:estimate_hint}, we give helper results leveraging Proposition~\ref{prop:sparse_recovery} to produce an estimate $\vhth$ to guide our sampler. In Section~\ref{ssec:recenter}, we give a rejection sampling procedure based on a conditional Poisson distribution, building upon Lemma~\ref{lem:recenter} to efficiently sample from a high-probability region over $\pisupp(\cdot \mid \mx, \vy)$. Finally, we combine these results to prove Theorem~\ref{thm:sas_basic} in Section~\ref{ssec:gaussian_proof}.

\subsection{Posterior estimation}\label{ssec:estimate_hint}
We first give an estimator $\vhth$ whose residual $\vy - \mx \vhth$ is bounded in an appropriate norm, with high probability over our  model. 
Our estimator builds upon an arbitrary $\ell_\infty$ sparse recovery algorithm, e.g., those in Proposition~\ref{prop:sparse_recovery}. We state the guarantees we require of such an algorithm as follows.

\begin{assumption}\label{assume:sr_linf}
Let $\alg(\mx, \vy)$ be an algorithm with the following guarantee, parameterized by a universal constant $C_{\alg}$ and $R_{\alg}: \N \times (0, 1) \times (0, 1) \to \R_{\ge 0}$. In the setting of Model~\ref{model:sparse_recovery}, suppose $\mx$ is $(\eps, C_{\alg}k)$-RIP for $\eps \in (0, 1)$. Then $\alg(\mx, \vy)$ returns $\vth'$ satisfying
\[\norm{\vth' - \vth^\star}_\infty \le \sigma \cdot R_{\alg}(k, \eps, \delta), \text{ with probability } \ge 1 - \delta,\]
for any $\delta \in (0, 1)$, over the randomness of $\vxi$.\footnote{We state our result for deterministic algorithms $\alg$ for simplicity, as Proposition~\ref{prop:sparse_recovery} is deterministic.}
\end{assumption}

For example, $\alg \gets \alg_\infty$ in Proposition~\ref{prop:sparse_recovery} satisfies Assumption~\ref{assume:sr_linf} with $C_{\alg} = 2$ and $R_{\alg}(k, \eps, \delta) \approx \sqrt{\log(d/\delta)}$ whenever $\eps \le \frac 1 {4k}$, by using Lemma~\ref{lem:maxrow_ip}. Similarly, $\alg \gets \alg_2$ in Proposition~\ref{prop:sparse_recovery} satisfies Assumption~\ref{assume:sr_linf} with $C_{\alg} \gets C_2$ and $R_{\alg}(k, \eps, \delta) \approx \sqrt{k}$ (up to a $\textup{polylog}(\frac 1 \delta)$ factor) whenever $\eps \le \frac 1 {10}$.

We now show how to use Assumption~\ref{assume:sr_linf} to produce an estimator of the large coordinates of a sample from the posterior distribution $\pi(\cdot \mid \mx, \vy)$ with high probability, and which has a bounded residual.

\begin{lemma}
    \label{lem:post_concentration}
    Let Assumption~\ref{assume:sr_linf} hold. In the setting of Model~\ref{model:sas_basic}, let $\mu$ be arbitrary, let $\eps, \delta \in (0, 1)$, and suppose $\mx$ is $(C_{\alg}k^\star, \eps)$-RIP for some $k^\star \in \N$ with $k^\star \ge 2(k + \log(\frac 5 \delta))$.
    Then there is an estimation procedure $\algest$ that takes as input $(\mx, \vy)$, and produces $\vhth$ satisfying
    \begin{equation}\label{eq:post_concentration}
    \Pr_{\vth \sim \pi(\cdot \mid \mx, \vy)}\Brack{\Par{\infnorm{\vhth - \vth} \le 6\sigma \cdot R_{\alg}\Par{k^\star, \eps, \frac {\delta^2} 5}} \wedge \Par{\supp(\vhth) \subseteq \supp(\vth)}} \ge 1 - \delta,    
    \end{equation}
    and
    \begin{equation}\label{eq:residual_bound}
    \begin{gathered}
        \norm{\mx_{S:}^\top\Par{\vy - \mx \vhth}}_2^2 \le 16\sigma^2 k^\star \cdot \Par{R_{\alg}\Par{k^\star, \eps, \frac{\delta^2}{5}}^2 + \log\Par{\frac{5d}{\delta}}}, \\
    \text{for all } S \subseteq [d] \text{ with } |S| + 4\Par{k + \log\Par{\frac 5 \delta}} \le k^\star,
    \end{gathered}
    \end{equation}
    with probability $\ge 1 - \delta$ over the randomness of Model~\ref{model:sas_basic}.
\end{lemma}
\begin{proof}
We begin by proving \eqref{eq:post_concentration}. By a Chernoff bound, with probability $\ge 1 - \frac{\delta^2}{5}$ over $\vth^\star \sim \pi$ in Model~\ref{model:sas_basic}, we have $|\supp(\vths)| \leq 4(k + \log(\frac 5 \delta))$. Under this condition, Assumption~\ref{assume:sr_linf} states that $\alg$ outputs an estimate $\vth'$ that satisfies
\begin{equation}\label{eq:condition_infbound_star}\infnorm{\vthp - \vths} \le \sigma \cdot R_{\alg}\Par{k^\star, \eps, \frac {\delta^2} 5}\end{equation}
with probability $\ge 1 - \frac {\delta^2} 5$ over the randomness of $\vxi$. Thus, after taking a union bound over these two failure probabilities, applying Proposition~\ref{prop:est_to_sample} shows that with probability $\ge 1 - \frac{4\delta^2} 5$ over the randomness of $\vth \sim \pi(\cdot \mid \mx, \vy)$ and Model~\ref{model:sas_basic}, our estimate $\vth'$ satisfies
\begin{equation}\label{eq:condition_infbound}\infnorm{\vthp - \vth} \le \infnorm{\vthp - \vths} + \infnorm{\vth - \vths} \le 3\sigma \cdot R_{\alg}\Par{k^\star, \eps, \frac{\delta^2}{5}}.\end{equation}
Therefore, applying Lemma~\ref{lem:most_models_good} with $\delta_1 \gets \delta$ and $\delta_2 \gets \frac{4\delta}{5}$, we have with probability $\ge 1 - \frac{4\delta}{5}$ over the randomness of Model~\ref{model:sas_basic} that \eqref{eq:condition_infbound} holds with probability $\ge 1 - \delta$ for $\vth \sim \pi(\cdot \mid \mx, \vy)$.

Define the coordinatewise operation $\clip(\cdot, \cdot): \R^d \times \R_{\ge 0} \to \R^d$ by
\[\Brack{\clip(\vx, \alpha)}_i = \begin{cases}
\vx_i & |\vx_i| > \alpha \\ 
0 & \text{else}
\end{cases} \text{ for all } i \in [d].\]
Our estimation procedure $\algest$ returns
\[\vhth \gets \clip\Par{\vthp, 3\sigma \cdot R_{\alg}\Par{k^\star, \eps, \frac {\delta^2} 5}}.\]
It is clear that $\norms{\vhth - \vthp}_\infty \le 3\sigma \cdot R_{\alg}(k^\star, \eps, \frac {\delta^2} 5)$, so applying the triangle inequality with \eqref{eq:condition_infbound} gives the first bound in \eqref{eq:post_concentration}. Moreover, to see that $\supp(\vhth) \subseteq \supp(\vth)$, suppose that some $i \in [d]$ has $i \notin \supp(\vth)$ but $i \in \supp(\vhth)$. The latter condition and the definition of $\clip$ implies
\[\Abs{\vthp_i - \vth_i} = |\vthp_i| > 3\sigma \cdot R_{\alg}\Par{k^\star, \eps, \frac {\delta^2} 5},\]
which contradicts \eqref{eq:condition_infbound}. We remark that the same logic with \eqref{eq:condition_infbound_star} shows $\supp(\vhth) \subseteq \supp(\vths)$.

Finally, we prove that \eqref{eq:residual_bound} holds. By Lemma~\ref{lem:maxrow_ip}, with probability $\ge 1 - \frac \delta 5$,
\[\norm{\mx^\top \vxi}_\infty \le \sigma \sqrt{8\log\Par{\frac {5d} \delta}}.\]
Condition on the above event, $\supp(\vhth) \supseteq \supp(\vths)$, and \eqref{eq:condition_infbound_star}, which all hold with probability $\ge 1 - \delta$ over Model~\ref{model:sas_basic} by a union bound. Then letting $T \defeq S \cup \supp(\vths)$, which satisfies $|T| \le k^\star$ by assumption, we have
\begin{align*}
\norm{\mx_{S:}^\top \Par{\vy - \mx \vhth}}_2^2 &\le 2\norm{\mx_{S:}^\top \mx\Par{\vths - \vhth}}_2^2 + 2\norm{\mx_{S:}^\top \vxi}_2^2 \\
&\le 2\norm{\Brack{\mx^\top \mx}_{T \times T}\Par{\vths - \vhth}}_2^2 + 2|S| \norm{\mx^\top \vxi}_\infty^2 \\
&\le 2 |T| \normop{\Brack{\mx^\top \mx}_{T \times T}}^2 \norm{\vths - \vhth}_\infty^2 + 16\sigma^2 k^\star \log\Par{\frac{5d}{\delta}} \\
&\le 8k^\star \norm{\vths - \vhth}_\infty^2 + 16\sigma^2 k^\star \log\Par{\frac {5d}\delta}
\end{align*}
where the first line used the inequality $\norm{\va + \vb}_2^2 \le 2\norm{\va}_2^2 + 2\norm{\vb}_2^2$, the second used that $\supp(\vhth) \subseteq \supp(\vths) \subseteq T$ and $S \subseteq T$, the third used our bound on $\norms{\mx^\top \vxi}_\infty$, and the last used our assumption that $\mx$ is RIP so $\normsop{[\mx^\top \mx]_{T \times T}} \le 2$. The conclusion follows from the above and \eqref{eq:condition_infbound_star}.
\end{proof}

We also show that Lemma~\ref{lem:post_concentration} gives concentration bounds on our estimator.

\begin{lemma}
    \label{lem:vhth_norm_bound}
    In the setting of Lemma~\ref{lem:post_concentration}, let $k_0 \defeq 4(k + \log(\frac{10}{\delta}))$, and suppose that for some $N: (0, 1) \to \R_{\ge 0}$ the density $\mu$ in Model~\ref{model:sas_basic} satisfies
    \[\Pr_{\vv \sim \mu^{\otimes k_0}}\Brack{\norm{\vv}_2^2 \ge N(\delta)} \le \frac \delta 2, \text{ for all } \delta \in (0, 1).\]
    Then for $\vhth$ the estimator returned by Lemma~\ref{lem:post_concentration}, we have 
    \begin{equation*}
        \twonorm{\vhth}^2 \leq 8k_0 \sigma^2 R_{\alg}\Par{k^\star, \eps, \frac {\delta} {10}}^2 + 2N(\delta),\text{ with probability } \ge 1 - \delta,
    \end{equation*}
    for any $\delta \in (0, 1)$, over the randomness of Model~\ref{model:sas_basic}.
\end{lemma}
\begin{proof}
As argued in Lemma~\ref{lem:post_concentration}, with probability $\ge 1 - \frac \delta 2$ over Model~\ref{model:sas_basic}, we have both
\[\supp(\vths) \le k_0,\; \norm{\vhth - \vths}_\infty \le 2\sigma \cdot R_{\alg}\Par{k^\star, \eps, \frac \delta {10}}.\]
Condition on the above events and that $\norms{\vths}_2^2 \le N(\delta)$, which gives the failure probability of $\delta$ by a union bound. Under these events we have the conclusion:
\begin{align*}
\twonorm{\vhth}^2 &\leq 2\twonorm{\vhth - \vths}^2 + 2\twonorm{\vths}^2 \leq 2k_0\norm{\vhth - \vths}_\infty^2 + 2N(\delta) \\
&\le 8k_0 \sigma^2 R_{\alg}\Par{k^\star, \eps, \frac \delta {10}}^2 + 2N(\delta).
\end{align*}
\end{proof}

We will instantiate Lemma~\ref{lem:vhth_norm_bound} with different choices of $\mu$. For the rest of this section, we focus on the basic setting of $\mu = \Nor(0,1)$, for which we have the following corollary.

\begin{corollary}\label{cor:vhth_gaussian}
Let $\delta \in (0, 1)$. In the setting of Lemma~\ref{lem:post_concentration}, let $k_0 \defeq 4(k + \log(\frac{10}{\delta}))$, and suppose $\mu = \Nor(0, 1)$. Then for $\vhth$ the estimator returned by Lemma~\ref{lem:post_concentration}, all of the following hold with probability $\ge 1 - \delta$ over the randomness of Model~\ref{model:sas_basic}: \eqref{eq:post_concentration} (with $\delta \gets \frac \delta 2$), \eqref{eq:residual_bound} (with $\delta \gets \frac \delta 2$), and
    \begin{equation*}
        \twonorm{\vhth}^2 \leq 8k_0 \Par{\sigma^2 R_{\alg}\Par{k^\star, \eps, \frac \delta {10}}^2 + 1}.
    \end{equation*}
\end{corollary}
\begin{proof}
We apply Lemma~\ref{lem:vhth_norm_bound} with the following bound from Fact~\ref{fact:chisquare} with $t \gets \log(\frac 2 \delta)$:
\[N(\delta) \le 2k_0 + 3\log\Par{\frac 2 \delta} \le 4k_0.\]
\end{proof}

\subsection{Centered rejection sampling}\label{ssec:recenter}
We now develop our posterior sampler, an instance of rejection sampling (Lemma~\ref{lem:reject}). In Algorithm~\ref{alg:sample_tpi}, we first define the proposal distribution our rejection sampler is based on. This proposal is a product distribution that we show approximates the true posterior density $\pi(\cdot \mid \mx, \vy)$ over a high-probability region of candidate subsets according to $\pisupp(\cdot \mid \mx, \vy)$. It takes an additional input $\vhth$ that is used to center its samples, as discussed following Lemma~\ref{lem:recenter}. Finally, it produces its output in Line~\ref{line:cps_gaussian} by calling our conditional Poisson sampler (Algorithm~\ref{alg:cond_sampling}, cf.\ Lemma~\ref{lemma:conditional_sampling}).

\begin{algorithm}
    \caption{$\productsg(\mx, \vy, \vhth, \sigma, \vq, k^\star)$}
    \label{alg:sample_tpi}
    \begin{algorithmic}
    \State \textbf{Input:} {$\mx \in \R^{n \times d}$, $\vy \in \R^n$, $\sigma > 0$, $\vq \in [0, 1]^d$ produced by Model~\ref{model:sas_basic}, $\vhth \in \R^d$, $k^\star \in \N$.}
    \State \textbf{Output:} {Sample $S$ from a conditional Poisson distribution over
    \begin{equation}\label{eq:cond_set_def}S \in \Omega_{\vhth, k^\star} \defeq \Brace{S \subseteq [d] \mid S \supseteq \supp(\vhth), |S| \le k^\star}\end{equation}}
    
        \State $\vz \gets \frac{1}{\sigma^2}\mx^\top(\vy - \mx\vhth) - \vhth$\;
        \State $T \gets \supp(\vhth)$\;
        \For {$i \in T^c$}
        \State $\vr_i \gets \vq_i\sqrt{\frac{\sigma^2}{1 + \sigma^2}}\exp\paren{\frac{\sigma^2}{2(1 +\sigma^2)}\vz_i^2}$\label{line:rdef}\;
        
        \State $\vp_i \gets \frac{\vr_i}{1 - \vq_i + \vr_i}$\label{line:pdef}\;
        \EndFor
        \State\Return $T \cup \cps(\vp, k^\star - |T|)$ \label{line:cps_gaussian}
    \end{algorithmic}
\end{algorithm}
We next derive unnormalized distributions according to $\pisupp(\cdot \mid \mx, \vy)$ (cf.\ \eqref{eq:pisupp_post}) and Algorithm~\ref{alg:sample_tpi}.

\begin{lemma}
\label{lem:unnorm_post}
In the setting of Model~\ref{model:sas_basic}, let $\vhth\in \bbR^d$, and let $\tilde{\pi}_{\supp}(\cdot) := \textup{Law}(S)$ where $S \in \Omega_{\vhth, k^\star}$ (defined in \eqref{eq:cond_set_def}) is a sample from Algorithm~\ref{alg:sample_tpi}.
Then,
\begin{equation}\label{eq:def_Q_S}
\tpisupp(S ) \propto \Par{\prod_{i\in S}\frac{\vq_i}{1 - \vq_i}}\exp\paren{\frac{\sigma^2}{2(1 + \sigma^2)}\twonorm{\vz_S}^2}\paren{\frac{\sigma^2}{1 + \sigma^2}}^{\frac{|S|}{2}} =: Q(S), 
\end{equation}
and further, the conditional distribution of $\pisupp(\cdot \mid \mx, \vy)$ over $\Omega_{\vhth, k^\star}$ satisfies
\begin{equation}
        \label{eq:def_P_S}
        \pisupp\Par{S\mid \mx, \vy, S \in \Omega_{\vhth, k^\star}} \propto \Par{\prod_{i\in S}\frac{\vq_i}{1 - \vq_i}}\exp\paren{\frac{1}{2}\norm{\vz_S}_{\ma_S^{-1}}^2}\frac{1}{\sqrt{\det \ma_S}} =: P(S),
    \end{equation}
    where $\vz$ and $\ma_S$ are defined as 
    \begin{equation}
     \label{eq:def_vz_ma_S}
        \vz \defeq \frac{1}{\sigma^2}\mx^\top(\vy - \mx \vhth) - \vhth,\quad \ma_S\defeq \frac{1}{\sigma^2}[\mx^\top \mx]_{S\times S} + \id_S.
    \end{equation}
\end{lemma}
\begin{proof}
First we show \eqref{eq:def_Q_S} is true. Observe that the output of Algorithm~\ref{alg:sample_tpi} is the conditional distribution over $\Omega_{\vhth, k^\star}$, of a Bernoulli product density over subsets of $[d]$. Without loss of generality, we let $i \in T$ be included in the output with probability $\frac{\vr_i}{1 - \vq_i + \vr_i}$ defined consistently with Lines~\ref{line:rdef} and~\ref{line:pdef}. Hence, for the unconditional product distribution, we have
\begin{align*}
    \tpisupp(S) &= \paren{\prod_{i \in S}\frac{\vr_i}{1 - \vq_i+ \vr_i}} \cdot \paren{\prod_{i\in S^c} \frac{1 - \vq_i}{1 - \vq_i + \vr_i}}\\
    &= \prod_{i\in S}\paren{\frac{\vr_i}{1 - \vq_i}} \cdot \prod_{i \in [d]}(1 - \vq_i) \cdot \prod_{i\in [d]}(1 - \vq_i + \vr_i)\\
    &\propto \prod_{i\in S}\paren{\frac{\vr_i}{1 - \vq_i}},
\end{align*}
where $\vp_i$ is defined in Algorithm~\ref{alg:sample_tpi}. This proportionality remains true after conditioning on $S \in \Omega_{\vhth, k^\star}$. Finally, by the definition of $\vr_i$, we have the claim:
\begin{align*}
    \tpisupp(S) &\propto \prod_{i\in S}\paren{\frac{\vq_i}{1 - \vq_i} \sqrt{\frac{\sigma^2}{1 + \sigma^2}}\exp \paren{\frac{\sigma^2}{2(1 + \sigma^2)}\vz_i^2}}\\
    &= \Par{\prod_{i\in S}\frac{\vq_i}{1 - \vq_i}}\exp\paren{\frac{\sigma^2}{2(1 + \sigma^2)}\twonorm{\vz_S}^2}\paren{\frac{\sigma^2}{1 + \sigma^2}}^{\frac{|S|}{2}}.
\end{align*}
Next we prove~\eqref{eq:def_P_S}. Recall from~\eqref{eq:pisupp_post} that 
\begin{equation}
\label{eq:pisupp_core_term}
\begin{aligned}
\pisupp(S \mid \mx, \vy) &\propto \Par{\prod_{i\in S} \frac{\vq_i}{1 - \vq_i}}\paren{\frac{1}{2\pi}}^{\frac{|S|}{2}}\int_{\R^S} \exp\Par{-\frac{1}{2\sigma^2}\norm{\vy - \mx \vth_S}_2^2-\half\norm{\vth_S}_2^2} \dd \vth_S \\
&\propto \Par{\prod_{i\in S} \frac{\vq_i}{1 - \vq_i}}\paren{\frac{1}{2\pi}}^{\frac{|S|}{2}} \int_{\R^S} \exp\Par{-\half \vth_S^\top \ma_S \vth_S + \Par{\frac 1 {\sig^2} [\mx^\top \vy]_S}^\top \vth_S} \dd \vth_S \\
&\propto \Par{\prod_{i\in S} \frac{\vq_i}{1 - \vq_i}} \exp\Par{\half\norm{\frac 1 {\sig^2}\mx^\top \vy}_{\ma_S^{-1}}^2} \frac{1}{\sqrt{\det \ma_S}},
\end{aligned}
\end{equation}
where in the last line, we used Fact~\ref{fact:gauss_norm}. The conclusion follows from combining \eqref{eq:pisupp_core_term} with Lemma~\ref{lem:recenter} because the resulting distribution is supported on sets that all contain $\supp(\vhth)$.
\end{proof}

We next bound the ratio between the unnormalized distributions $P, Q$ defined in \eqref{eq:def_Q_S}, \eqref{eq:def_P_S}.

\begin{lemma}
\label{lem:rejection_rate_bound}
In the setting of Lemma~\ref{lem:unnorm_post}, suppose that $\mx$ is $(\eps, k^\star)$-RIP for $\eps \in (0, \half)$. Then for any $S \subseteq [d]$ such that $\supp(\vhth) \subseteq S$ and $|S| \le k^\star$, following the notation \eqref{eq:def_Q_S}, \eqref{eq:def_P_S}, \eqref{eq:def_vz_ma_S},
\[\paren{\frac{1}{1 + \epsilon}}^{\frac{k^\star}{2}}\exp\paren{-\frac{\sigma^2\epsilon}{(1 + \sigma^2)^2}\twonorm{\vz_S}^2} \leq \frac{P(S)}{Q(S)} \leq \paren{\frac{1}{1 - \epsilon}}^{\frac{k^\star}{2}}\exp\paren{\frac{\sigma^2\epsilon}{(1 + \sigma^2)^2}\twonorm{\vz_S}^2}.\]
\end{lemma}

\begin{proof}
First, we can directly compute that 
\begin{align}
\nonumber
    \frac{P(S)}{Q(S)} &= \frac{\Par{\prod_{i\in S}\frac{\vq_i}{1 - \vq_i}}\exp\paren{\frac{1}{2}\norm{\vz_S}_{\ma_S^{-1}}^2}\det(\ma_S^{-1})^{\frac{1}{2}}}{\paren{\prod_{i\in S}\frac{\vq_i}{1 - \vq_i}}\paren{\frac{\sigma^2}{1 + \sigma^2}}^{\frac{|S|}{2}}\exp\paren{\frac{\sigma^2}{2(1 + \sigma^2)}\twonorm{\vz_S}^2}}\\
\nonumber
    &= \det(\ma_S^{-1})^{\frac{1}{2}} \cdot \paren{1 + \frac{1}{\sigma^2}}^{\frac{|S|}{2}}
    \exp\paren{\frac{1}{2}\paren{\norm{\vz_S}_{\ma_S^{-1}}^2 - \frac{\sigma^2}{1 + \sigma^2} \twonorm{\vz_S}^2}}\\
\label{eq:dist_ratio_1}
    &= \det(\ma_S^{-1})^{\frac{1}{2}} \cdot \paren{1 + \frac{1}{\sigma^2}}^{\frac{|S|}{2}}
    \exp\paren{\frac{1}{2}\vz_S^\top \paren{\ma_S^{-1} - \frac{\sigma^2}{1 + \sigma^2}\id_S}\vz_S}.
\end{align}
Since $\mx$ is $(\epsilon, k^\star)$ RIP and $|S|\leq k^\star$,
\begin{equation*}
    \paren{1 + \frac{1-\epsilon}{\sigma^2}}\id_S \preceq \ma_S = \frac{1}{\sigma^2}\Brack{\mx^\top\mx}_{S\times S} + \id_S \preceq \paren{1 + \frac{1+\epsilon}{\sigma^2}}\id_S.
\end{equation*}
Thus, we have that
\begin{gather*}
    \paren{\frac{\sigma^2}{1 + \sigma^2 +\epsilon}}^{k^\star} \leq \det\paren{\ma_S^{-1}}\leq \paren{\frac{\sigma^2}{1 + \sigma^2 -\epsilon}}^{k^\star},\\
    \normop{\ma_S^{-1} - \frac{\sigma^2}{1 + \sigma^2}\id_S} \leq \frac{\sigma^2\epsilon}{(1 + \sigma^2)(1 + \sigma^2 - \epsilon)}.
\end{gather*}
Plugging these bounds into~\eqref{eq:dist_ratio_1}, we have the desired claims for $\eps \in (0, \half)$:
\begin{align*}
    \frac{P(S)}{Q(S)} &\leq \paren{\frac{1 + \sigma^2}{1 + \sigma^2 - \epsilon}}^{\frac{k^\star}{2}}\exp\paren{\frac{\sigma^2\epsilon}{2(1 + \sigma^2)(1 + \sigma^2 - \epsilon)}\twonorm{\vz_S}^2} \\
    &\leq \paren{\frac{1}{1 - \epsilon}}^{\frac{k^\star}{2}}\exp\paren{\frac{\sigma^2\epsilon}{(1 + \sigma^2)^2}\twonorm{\vz_S}^2},\\
    \frac{P(S)}{Q(S)} &\geq \paren{\frac{1 + \sigma^2}{1 + \sigma^2 + \epsilon}}^{\frac{k^\star}{2}}\exp\paren{-\frac{\sigma^2\epsilon}{2(1 + \sigma^2)(1 + \sigma^2 + \epsilon)}\twonorm{\vz_S}^2} \\
    &\geq \paren{\frac{1}{1 + \epsilon}}^\frac{k^\star}{2}\exp\paren{-\frac{\sigma^2\epsilon}{(1 + \sigma^2)^2}\twonorm{\vz_S}^2}.
\end{align*}
\end{proof}
\begin{algorithm}
    \caption{$\postsg(\mx, \vy, \vhth, \sigma, \vq, k^\star, \delta)$}
    \label{alg:post_sample}
    \begin{algorithmic}[1]
        \State \textbf{Input:} $\mx \in \R^{n \times d}$, $\vy \in \R^n$, $\sig > 0$, $\vq \in [0, 1]^d$ produced by Model~\ref{model:sas_basic}, $\vhth \in \R^d$, $k^\star \in \N$, $\delta \in (0, 1)$
        \State \textbf{Output:} Sample $S$ approximately distributed as $ \pisupp(\cdot \mid \mx, \vy)$ (cf.\ Proposition~\ref{prop:post_supp_sample})
        \State $\mu \gets \productsg(\mx, \vy, \vhth, \sigma, \vq, \frac 3 4 k^\star)$\;
        \State $P, Q \gets $ unnormalized distributions in \eqref{eq:def_P_S}, \eqref{eq:def_Q_S}\;
        \State $\Omega \gets \Omega_{\vhth, \frac 3 4 k^\star}$ defined in \eqref{eq:cond_set_def}
        \State \Return $\rejectionsample(\mu, P, Q, \Omega, 3, \frac \delta 2)$
    \end{algorithmic}
\end{algorithm}

\begin{proposition}
    \label{prop:post_supp_sample}
In the setting of Model~\ref{model:sas_basic}, let $\vhth$ be produced as in Corollary~\ref{cor:vhth_gaussian} with $\delta \gets \frac \delta 4$, and let 
\begin{equation}\label{eq:keps_bound}k^\star \ge 16\Par{k + \log\Par{\frac{40}{\delta}}},\; \eps \le \frac 1 {32k^\star \cdot \Par{R_{\alg}\Par{k^\star, \eps, \frac{\delta^2}{320}}^2 + \log\Par{\frac{40}{\delta}}}}.\end{equation}
Then if $\mx$ is $(\eps, C_{\alg}k^\star)$-RIP for the universal constant $C_{\alg}$ from Assumption~\ref{assume:sr_linf}, with probability $\ge 1 - \delta$ over the randomness of Model~\ref{model:sas_basic}, Algorithm~\ref{alg:post_sample} returns $S \sim \nu$ satisfying $\tv{\nu, \pisupp(\cdot \mid \mx, \vy)} \le \delta$. If $\vhth$ is given, Algorithm~\ref{alg:post_sample} runs in time
\[O\Par{nd + \Par{dk^\star +n\Par{k^\star}^{\omega - 1}}\log\Par{\frac 1 \delta}}.\]
\end{proposition}
\begin{proof}
First, applying Corollary~\ref{cor:vhth_gaussian} with $\delta \gets \frac \delta 4$ shows that with probability $\ge 1 - \frac \delta 4$ over the randomness of Model~\ref{model:sas_basic}, we have
\begin{equation}\label{eq:goodmodel1}
\begin{gathered}
\Pr_{S \sim \pisupp(\cdot \mid \mx, \vy)}\Brack{\supp(\vhth) \subseteq \supp(\vth)} \ge 1 - \frac \delta 4, \\
\norm{\vhth}_2^2 \le 32\Par{k + \log\Par{\frac{40}{\delta}}}\Par{ \sig^2 R_{\alg}\Par{k^\star, \eps, \frac \delta {40}}^2 + 1}, \\
\text{and } \norm{\mx_{S:}^\top \Par{\vy - \mx \vhth}}_2^2 \le 16\sig^2 k^\star \cdot \Par{R_{\alg}\Par{k^\star, \eps, \frac{\delta^2}{320}}^2 + \log\Par{\frac{40d}{\delta}}}, \\
\text{for all } S \subseteq [d] \text{ with } |S| \le k^\star - 4\Par{k + \log\Par{\frac{40}{\delta}}} = \frac{3k^\star}{4}.
\end{gathered}
\end{equation}
Next, by applying Corollary~\ref{cor:post_sparse} with $\delta \gets \frac {\delta^2}{8}$, we have by Lemma~\ref{lem:most_models_good} that with probability $\ge 1 - \frac \delta 2$ over the randomness of Model~\ref{model:sas_basic}, we have
\begin{equation}\label{eq:goodmodel2}
\Pr_{S \sim \pisupp(\cdot \mid \mx, \vy)}\Brack{|S| \le \frac{3k^\star}{4}} \le \frac \delta 4,
\end{equation}
for our choice of $k^\star$. In the remainder of the proof, we union bound over all conditions in \eqref{eq:goodmodel1} and~\eqref{eq:goodmodel2} holding, which gives the failure probability of $\delta$ over Model~\ref{model:sas_basic}.

Next, following notation of Algorithm~\ref{alg:post_sample}, for $S \in \Omega_{\vhth, \frac 3 4 k^\star}$ and $\vz$ defined in \eqref{eq:def_vz_ma_S}, we have
\begin{align*}
\norm{\vz_S}_2^2 &\le \frac{2}{\sig^4}\norm{\mx_{S:}^\top (\vy - \mx \vhth)}_2^2 + 2\norm{\vhth}_2^2 \\
&\le 32k^\star\Par{R_{\alg}\Par{k^\star, \eps, \frac{\delta^2}{320}}^2\Par{\frac 1 {\sig^2} + \sig^2} + 1 + \frac{\log(\frac{40}{\delta})}{\sig^2}} \\
&\le 32k^\star \cdot \frac{(1 + \sig^2)^2}{\sig^2} \cdot \Par{R_{\alg}\Par{k^\star, \eps, \frac{\delta^2}{320}}^2 + \log\Par{\frac{40}{\delta}}},
\end{align*}
where we used the second and third bounds in \eqref{eq:goodmodel1}. Thus, by Lemma~\ref{lem:rejection_rate_bound}, for $S \in \Omega_{\vhth, \frac 3 4 k^\star}$,
\begin{align*}
\frac{P(S)}{Q(S)} &\le \Par{1 + \frac{1}{32k^\star}}^{\frac{k^\star}{2}}\exp\Par{\frac{\sig^2 \eps}{(1 + \sig^2)^2}\norm{\vz_S}_2^2} \\
&\le 1.1 \exp\Par{32\eps k^\star \cdot \Par{R_{\alg}\Par{k^\star, \eps, \frac{\delta^2}{320}}^2 + \log\Par{\frac{40}{\delta}}}} \le 3,
\end{align*}
and an analogous calculation shows $\frac{Q(S)}{P(S)} \le 3$. Therefore, Lemma~\ref{lem:reject} shows that Algorithm~\ref{alg:post_sample} correctly returns a sample within total variation distance $\frac \delta 2$ from 
\[\pisupp\Par{S \mid \mx, \vy, S \in \Omega_{\vhth, \frac 3 4 k^\star}}.\]
This distribution is in turn within total variation distance $\frac \delta 2$ from the unconditional distribution $\pisupp(\cdot \mid \mx, \vy)$, by the first bound in \eqref{eq:goodmodel1} and \eqref{eq:goodmodel2}. This concludes the proof of correctness.

For the runtime bound, we first precompute $\vz$ required by Algorithm~\ref{alg:sample_tpi} in $O(nd)$ time. Next, by Lemma~\ref{lem:reject}, we need to produce a sample $S \sim \mu$ and calculate the likelihood ratio $\frac{P(S)}{Q(S)}$ at most $O(\log(\frac 1 \delta))$ times. By Lemma~\ref{lemma:conditional_sampling}, each draw from $\mu$ takes time $O(dk^\star)$. Further, by the definition \eqref{eq:def_P_S}, each evaluation of $P(S)$ takes time
\[O\Par{n|S|^{\omega - 1}} \le O\Par{n\Par{k^\star}^{\omega - 1}}.\]
To see this, it is clear that we can compute $\ma_S$ in this time, and the costs of inverting \cite{Strassen69} and computing the determinant \cite{PanC99} do not dominate. Similarly, it is clear that evaluating $Q(S)$ only takes $O(|S|)$ time once $\vz$ is precomputed. Combining these costs yields the claim.
\end{proof}

We conclude with a closed-form expression for $\pi(\vth \mid \mx, \vy, S)$ in \eqref{eq:thsupp_post}.

\begin{lemma}\label{lem:theta_sample}
    In the setting of Model~\ref{model:sas_basic}, if 
    $\mu = \calN(0, 1)$, then $\pi(\vth\mid \mx, \vy, S) = \calN(\vmu_S, \ma_S^{-1})$, where $\ma_S$ is defined in~\eqref{eq:def_vz_ma_S}, and $\vmu \defeq \frac{1}{\sigma^2}\ma_S^{-1}\mx^\top \vy$.
\end{lemma}
\begin{proof}
This follows immediately from rewriting \eqref{eq:thsupp_post}: for $\vth$ with $\supp(\vth) \subseteq S$,
    \begin{align*}
        \pi(\vth \mid \mx, \vy, S) &\propto \exp\paren{-\frac{1}{2\sigma^2}\twonorm{\vy - \mx\vth}^2 - \frac{1}{2}\twonorm{\vth}^2}\\
        &\propto \exp\paren{-\frac{1}{2}\vth^\top \Par{\frac{1}{\sigma^2}\Brack{\mxtx}_{S\times S} + \id_S}\vth + \frac 1 {\sigma^2} (\mx^\top \vy)^\top \vth}\\
        &\propto \exp\Par{-\frac{1}{2}(\vth - \vmu_S)^\top \ma_S (\vth - \vmu_S)}.
    \end{align*}
\end{proof}

\subsection{Proof of Theorem~\ref{thm:sas_basic}}\label{ssec:gaussian_proof}

We now instantiate Proposition~\ref{prop:post_supp_sample} with the solvers from Proposition~\ref{prop:sparse_recovery} to prove our main theorem.

\begin{theorem}\label{thm:sas_basic}
Let $\delta \in (0, 1)$. In the setting of Model~\ref{model:sas_basic}, suppose $\mx$ is $(\eps, k^\star)$-RIP where
\[k^\star = \Omega\Par{k + \log\Par{\frac 1 \delta}},\; \eps =O\Par{ \frac{1}{\Par{k + \log\Par{\frac 1 \delta}}\log\Par{\frac d \delta}}},\]
for appropriate constants.
There is an algorithm (Algorithm~\ref{alg:post_sample} using $\vhth$ produced by $\alg_\infty$ in Proposition~\ref{prop:sparse_recovery}) that returns $\vth \sim \pi'$ for a distribution $\pi'$ satisfying $\tv{\pi', \pi(\cdot \mid \mx, \vy)} \le \delta$, with probability $\ge 1 - \delta$ over the randomness of Model~\ref{model:sas_basic}. The algorithm runs in time
\[O\Par{d^{1.5}n^2\log\Par{d\kappa\Par{1 + \frac{1}{\sigma}}} + d\log^2\Par{\frac 1 \delta} + n\log^\omega\Par{\frac 1 \delta}}, \text{ for any } \kappa \ge \frac{\vsig_1(\mx)}{\vsig_n(\mx)}.\]
Alternatively, suppose $\mx$ is $(\eps, k^\star)$-RIP, and for some $m \in [n]$, $\sqrt{m/n} \cdot \mx_{[m]:}$ is $(\frac 1 {10}, k^\star)$-RIP where
\[k^\star = \Omega\Par{k + \log\Par{\frac 1 \delta}},\; \eps = O\Par{\frac{1}{(k + \log(\frac 1 \delta)) (m + \log\Par{\frac 1 \delta})}},\]
for appropriate constants. There is an algorithm (Algorithm~\ref{alg:post_sample} using $\vhth$ produced by $\alg_2$ in Proposition~\ref{prop:sparse_recovery}) that returns $\vth \sim \pi'$ for a distribution $\pi'$ satisfying $\tv{\pi', \pi(\cdot \mid \mx, \vy)} \le \delta$, with probability $\ge 1 - \delta$ over the randomness of Model~\ref{model:sas_basic}. The algorithm runs in time
\[O\Par{nd\log\Par{1 + \frac 1 \sigma} + \Par{dk^\star + n\Par{k^\star}^{\omega - 1}}\log\Par{\frac 1 \delta}}.\]
\end{theorem}
\begin{proof}
The first claim follows from the $\ell_\infty$ recovery result in Proposition~\ref{prop:sparse_recovery}, combined with Proposition~\ref{prop:post_supp_sample}. Note that Lemma~\ref{lem:maxrow_ip} bounds $R_{\alg}(k^\star, \eps, \frac{\delta^2}{320})^2 = O(\log(\frac d \delta))$ in this setting. Moreover, to obtain our runtime bound, we use the estimates 
\[r_\infty = \Theta\Par{\sigma\sqrt{\log\Par{\frac d \delta}}},\; R_{\infty} = \Theta\Par{r_\infty\Par{1 + \frac d \sigma}}.\]
Correctness of the former bound follows from Lemma~\ref{lem:maxrow_ip}, and the latter follows from
\[\norm{\mx^\top \vy}_\infty \le \norm{\mx^\top \vxi}_\infty + \norm{\mx^\top \mx \vths}_\infty = O\Par{r_\infty + \norm{\vths}_2} = O\Par{r_\infty\Par{1 + \frac d \sigma}},\]
where the first equality holds since $\vths$ is $k^\star$-sparse within the allotted failure probability, and we can bound $\norm{\vths}_2^2 = O(k^\star + \log(\frac 1 \delta))$ using Fact~\ref{fact:chisquare} to obtain the second.

The second claim follows from the $\ell_2$ recovery result in Proposition~\ref{prop:sparse_recovery}, combined with Proposition~\ref{prop:post_supp_sample}. Again by Fact~\ref{fact:chisquare}, we have that $R_{\alg}(k^\star, \eps, \frac{\delta^2}{320})^2 = O(m + \log(\frac 1 \delta))$ in this setting. For the runtime, we set $R_2^2 = O(k^\star + \log(\frac 1 \delta))$ as above, and $r_2^2 = O(\sigma^2 \cdot (m + \log(\frac 1 \delta)))$ as before. The conclusion follows because $m \ge k^\star$: since $\mx_{[m]:}$ is RIP, otherwise column subsets of size $k^\star$ would not be full rank.

Finally, we note that in both cases Proposition~\ref{prop:post_supp_sample} only refers to the distribution of the support $S \subseteq [d]$. However, by applying Lemma~\ref{lem:theta_sample} on the sampled $S$, we can exactly sample from the corresponding distribution over $\vth$, which cannot increase the total variation distance by using the same coupling over $S$. The runtime of this last step does not dominate.
\end{proof}

To give a concrete example of applying Theorem~\ref{thm:sas_basic} to a well-studied random matrix ensemble, consider the case where $\mx \in \R^{n \times d}$ has i.i.d.\ entries $\sim \Nor(0, \frac 1 n)$ (to which Proposition~\ref{prop:RIP_sub_gaussian} applies). We require the following concentration inequalities on $\kappa(\mx)$.

\begin{lemma}[\cite{ChenD05}]\label{lem:subgaussian_condition_number}
Assume that $d$ is sufficiently larger than $n$. Let $\mx \in \R^{n \times d}$ have entries i.i.d.\ $\sim \Nor(0, \frac 1 n)$, and let $\delta \in (0, 1)$. Then with probability $\ge 1 - \delta$,
\begin{equation}
\frac{\vsig_1(\mx)}{\vsig_n(\mx)} = O\Par{d + \Par{\frac 1 \delta}^{O(\frac 1 d)}}.
\end{equation}
\end{lemma}

We remark that similar high-probability condition number bounds exist for other matrix ensembles, see, e.g., Theorem 4.6.1 in \cite{vershynin2018high} for the case of entrywise sub-Gaussian matrices. 

By plugging in Proposition~\ref{prop:RIP_sub_gaussian} and Lemma~\ref{lem:subgaussian_condition_number} into Theorem~\ref{thm:sas_basic}, we have the following specialization of our posterior sampler to Gaussian measurement matrices $\mx$.

\begin{corollary}\label{cor:gaussian_X}
Let $\delta \in (0, 1)$, and suppose $\mx \in \R^{n \times d}$ has i.i.d.\ entries $\sim \Nor(0, \frac 1 n)$. If
\[n = \Omega\Par{\Par{k + \log\Par{\frac 1 \delta}}^3\log^3\Par{\frac d \delta}}\]
for an appropriate constant, in the setting of Model~\ref{model:sas_basic}, there is an algorithm that returns $\vth \sim \pi'$ for a distribution $\pi'$ satisfying $\tv{\pi', \pi(\cdot \mid \mx, \vy)} \le \delta$, with probability $\ge 1 - \delta$ over the randomness of Model~\ref{model:sas_basic} and $\mx$. The algorithm runs in time
\[O\Par{n^2 d^{1.5}\log\Par{d\Par{1 + \frac 1 \sig}} + n^2 \sqrt d \log\Par{\frac d \delta\Par{1 + \frac 1 \sig}}}.\]
Alternatively, if
\[n = \Omega\Par{\Par{k + \log\Par{\frac 1 \delta}}^5 \log^2\Par{\frac d \delta}\log\Par{d}}\]
for an appropriate constant, in the setting of Model~\ref{model:sas_basic}, there is an algorithm that returns $\vth \sim \pi'$ for a distribution $\pi'$ satisfying $\tv{\pi', \pi(\cdot \mid \mx, \vy)} \le \delta$, with probability $\ge 1 - \delta$ over the randomness of Model~\ref{model:sas_basic} and $\mx$. The algorithm runs in time
\[O\Par{nd\log\Par{1 + \frac 1 \sig}}.\]
\end{corollary}
\begin{proof}
The result using the $\ell_\infty$ estimator follows immediately from combining Theorem~\ref{thm:sas_basic} with Proposition~\ref{prop:RIP_sub_gaussian} and Lemma~\ref{lem:subgaussian_condition_number}. To obtain the result using the $\ell_2$ estimator, we first set
\[m = O(k^\star \log(d))\]
which is enough to guarantee the precondition of Theorem~\ref{thm:sas_basic} on $\sqrt{m/n} \cdot \mx_{[m]:}$ holds, by Proposition~\ref{prop:RIP_sub_gaussian}. We substitute this into our choice of $\eps$ and obtain our bound on $n$, again through Proposition~\ref{prop:RIP_sub_gaussian}. In both cases, we simplified by dropping all non-dominant runtime terms.
\end{proof}
\section{Spike-and-slab posterior sampling with Laplace prior}\label{sec:laplace}

In this section, we generalize the results of Section~\ref{sec:gaussian} to the setting where $\mu(x) \propto \exp(-|x|)$ (i.e., a Laplace prior) is used in Model~\ref{model:sas_basic}, for a range of signal-to-noise ratios $\sigma$. We begin by giving an annealing strategy in Section~\ref{ssec:anneal_laplace}, following \cite{GeLL20} (see also \cite{LovaszV06, CousinsV18, BrosseDM18}), for estimating normalizing constants used in our approximate rejection sampler. Next, we provide our proposal distribution and bound normalizing constant ratios in Section~\ref{ssec:reject_laplace}. Finally, we combine these results in Section~\ref{ssec:proof_laplace} to prove Theorem~\ref{thm:sas_laplace}, our main result on posterior sampling under a Laplace prior.

\subsection{Annealing}\label{ssec:anneal_laplace}

In this section, we consider the following computational task, parameterized by an approximation tolerance $\Delta \in (0, 1)$, a failure probability $\delta \in (0, 1)$, a matrix $\ma \in \PSD^{k \times k}$ satisfying $\mu \id_k \preceq \ma \preceq L \id_k$, and a vector $\vb \in \R^k$. Our goal is to produce an estimate $\tZ$, such that with probability $\ge 1 - \delta$,
\begin{equation}\label{eq:approx_z}\begin{aligned}
\int \exp\Par{-f(\vth)} \dd \vth &\in \Brack{(1 - \Delta)\tZ, (1 + \Delta) \tZ}, \\
\text{where }f\bb{\vth} &:= \frac{1}{2}\vth^{\top}\ma\vth - \vb^{\top}\vth + \norm{\vth}_1.
\end{aligned}
\end{equation}
This general question was studied by \cite{GeLL20}, who gave the following framework based on an annealing scheme, patterned in part off of prior work by \cite{LovaszV06, CousinsV18, BrosseDM18}.

\begin{proposition}[Section 3, Appendix B, \cite{GeLL20}]\label{prop:est_Z}
Let $f: \R^k \to \R$ be $\mu$-strongly convex, let $Z \defeq \int \exp(-f(\vth))\dd \vth$, and fix an approximation tolerance $\Delta \in (0, 1)$ and a failure probability $\delta \in (0, 1)$. Let $\sig_1 > 0$ be a parameter satisfying
\ba{
    \Par{1 - \frac \Delta 2} \int_{\vth \in \R^{k}} \exp\bb{-\frac{\norm{\vth}_{2}^{2}}{2\sigma_1^{2}}} \diff\vth
    \leq \int_{\vth \in \R^{k}} \exp\bb{-f_{1}\bb{\vth}}\diff\vth
    \leq \int_{\vth \in \R^{k}} \exp\bb{-\frac{\norm{\vth}_{2}^{2}}{2\sigma_1^{2}}} \diff\vth, \label{eq:sigma_1_condition}
}
where, for a sequence of increasing $\{\sig_i\}_{i \in [M]}$ specified in \cite{GeLL20}, we let
\[f_i(\vth) \defeq f(\vth) + \frac{\norm{\vth}_2^2}{2\sig_i^2}, \text{ for all } i \in [M].\]
Further, let
\[M = \Theta\Par{\sqrt{k}\log\Par{\frac{k}{\mu\sig_1^2}}},\; N = \Theta\Par{\frac{M^2}{\Delta^2}},\]
for appropriate constants. Finally, let $\alg$ be an algorithm that takes as input $i \in [M]$, and produces a sample within total variation distance $\frac 1 {8N}$ from the density on $\R^k$ that is $\propto \exp(-f_i)$. There is an algorithm that queries $\alg$ $O(N\log(\frac 1 \delta))$ times, and produces an estimate $\tZ$ satisfying
\[Z \in \Brack{(1 - \Delta)\tZ, (1 + \Delta)\tZ}, \text{ with probability } \ge 1 - \delta.\]
\end{proposition}
\begin{proof}
This is almost the exact derivation carried out in \cite{GeLL20}, except for two differences. First, they assume that the relevant negative log-density $f$ is $L$-smooth in addition to being strongly convex. However, one can check that the only place this assumption is used in their analysis is in Lemma 3.1 to choose $\sig_1$ satisfying \eqref{eq:sigma_1_condition}, which we replace with by isolating the sufficient condition \eqref{eq:sigma_1_condition}. Second, the \cite{GeLL20} result is stated for an algorithm which produces a correct estimate $Z$ with probability $\ge \frac 3 4$, using exactly $N$ calls to $\alg$. However, by repeating $\log(\frac 1 \delta)$ times and taking a median, Chernoff bounds show that we can boost the failure probability as stated.
\end{proof}

We next provide a sufficient choice of $\sig_1$ that satisfies \eqref{eq:sigma_1_condition} in the setting of \eqref{eq:approx_z}.

\begin{lemma}\label{lemma:choice_of_sigma_1}
    Let $\ma \in \PSD^{k \times k}$, let $\vb \in \R^{k}$ such that $\norm{\vb}_2 \leq R$, let $\Delta \in (0, 1)$, and let 
    \ba{
    \lambda \geq \frac{8R\sqrt k}{\Delta} + \frac{192k}{\Delta^2}\Par{k +\log\Par{\frac 4 \Delta}}.\label{eq:lam_lower_bound}
    }
    Denoting $\mm \defeq \ma + \lam \id_k$, we have
    \ba{
        \Par{1 - \frac \Delta 2}Z_{\mm, \vb} \leq \int\exp\bb{-\frac{1}{2}\vth^{\top}\mm\vth + \vb^{\top}\vth - \norm{\vth}_1}\diff\vth \leq Z_{\mm,\vb}\label{eq:Z1bounds}
    }
    where $Z_{\mm, \vb} := \int\exp\bb{-\frac{1}{2}\vth^{\top}\mm\vth + \vb^{\top}\vth}\diff\vth$.
\end{lemma}
\begin{proof}
The right-hand side of \eqref{eq:Z1bounds} is immediate from $\norm{\vth}_1 \ge 0$ for all $\vth \in \R^k$. In the rest of the proof, we prove the left-hand side of \eqref{eq:Z1bounds}. Define the set $\Omega \defeq \{\vth : \norm{\vth}_2 \leq \frac{\Delta}{4\sqrt{k}}\}$. Then, we have
    \ba{
        \int_{\R^{k}}\exp\bb{-\frac{1}{2}\vth^{\top}\mm\vth + \vb^{\top}\vth - \norm{\vth}_1}\diff\vth &\geq \int_{\Omega}\exp\bb{-\frac{1}{2}\vth^{\top}\mm\vth + \vb^{\top}\vth - \norm{\vth}_1}\diff\vth \notag \\
        &= \exp\bb{-\sup_{\Omega}\norm{\vth}_1}\int_{\Omega}\exp\bb{-\frac{1}{2}\vth^{\top}\mm\vth + \vb^{\top}\vth}\diff\vth\notag \\
        &\geq \exp\bb{-\frac{\Delta}{4}}\int_{ \Omega}\exp\bb{-\frac{1}{2}\vth^{\top}\mm\vth + \vb^{\top}\vth}\diff\vth, \label{eq:gaussian_high_prob_event}
    }
    where the last line used $\norm{\vth}_1 \leq \sqrt{k}\norm{\vth}_{2}$. Moreover, we claim that
    \begin{equation}\label{eq:g_omega}\Pr_{\vth \sim \Nor(\mm^{-1} \vb, \mm^{-1})}\Brack{\vth \in \Omega} \ge 1 - \frac \Delta 4,\end{equation}
    which completes the proof when combined with \eqref{eq:gaussian_high_prob_event}, since
    \begin{gather*}\exp\bb{-\frac{\Delta}{4}}\int_{ \Omega}\exp\bb{-\frac{1}{2}\vth^{\top}\mm\vth + \vb^{\top}\vth}\diff\vth \\
    \ge \Par{1 - \frac \Delta 4}\exp\Par{-\frac \Delta 4} \int_{\R^k}\exp\Par{-\half \vth^\top \mm \vth + \vb^\top \vth}\dd \vth \ge \Par{1 - \frac \Delta 2}Z_{\mm, \vb}, \end{gather*}
    as $\exp(-\half \vth^\top \mm \vth + \vb^\top\vth)$ is the unnormalized density corresponding to $\vth \sim \Nor(\mm^{-1}\vb, \mm^{-1})$. We now prove \eqref{eq:g_omega}. First, observe that for $\vth \sim \Nor(\mm^{-1} \vb, \mm^{-1})$, we can write
    \[\vth = \mm^{-1}\vb + \mm^{-\half}\vz, \text{ where } \vz \sim \Nor(\vzero_k, \id_k). \]
    Since $\mm \succeq \lam \id_k$, we have
    \[\norm{\mm^{-1}\vb}_2 \le \frac{R}{\lam} \le \frac{\Delta}{8\sqrt{k}},\; \norm{\mm^{-\half} \vz}_2 \le \frac{\norm{\vz}_2}{\sqrt{\lam}},\]
    for our range of $\lam$. Returning to \eqref{eq:g_omega}, this implies the desired
    \begin{align*}
    \Pr_{\vth \sim \Nor(\mm^{-1} \vb, \mm^{-1})}\Brack{\vth \not\in \Omega} &\le \Pr_{\vz \sim \Nor(\vzero_k, \id_k)}\Brack{\norm{\mm^{-\half} \vz}_2 \ge \frac{\Delta}{8 \sqrt k}} \\
    &\le \Pr_{\vz \sim \Nor(\vzero_k, \id_k)}\Brack{\norm{\vz}_2 \ge \frac{\Delta \sqrt{\lam}}{8 \sqrt k}} \le \frac \Delta 4,
    \end{align*}
    where we used Fact~\ref{fact:chisquare} with our choice of $\lam$.
\end{proof}

To facilitate our annealing scheme, we use a sampler from \cite{LeeST21} for composite densities.

\begin{proposition}[Corollary 2, \cite{LeeST21}]\label{prop:composite_sample}
Let $f: \R^k \to \R$ be $L$-smooth and $\mu$-strongly convex, let $\kappa \defeq \frac L \mu$, and let $g: \R^k \to \R$ have the form $g(\vx) = \sum_{i \in [k]} g_i(\vx_i)$ for scalar functions $\{g_i: \R \to \R\}_{i \in [k]}$. There is an algorithm which runs in $O(\kappa k \log^3(\frac{\kappa k}{\eps}))$ iterations, each querying $\nabla f$ and performing $O(k)$ additional work, and obtains $\eps$ total variation distance to the density over $\R^d$, $\pi \propto \exp(-f - g)$.
\end{proposition}
\begin{proof}
This is almost the statement of Corollary 2 in \cite{LeeST21}; we briefly justify the differences here. First, if $g$ is coordinatewise-separable, then the RGO access required by \cite{LeeST21} takes $O(k)$ time, under our assumption of $O(1)$-time integration and sampling over $\R$. Next, Corollary 2 in \cite{LeeST21} requires access to the minimizer of $f + g$, but the tolerance of the algorithm to inexactness is discussed in Appendix A of that paper, and it is justified why the cost of computing an approximate minimizer for composite functions does not dominate the sampler's gradient complexity. Finally, the runtime in \cite{LeeST21} is in expectation, but can be bounded with probability $1 - O(\delta)$ by using standard Chernoff bounds over the complexity of rejection sampling steps. The failure probability of the runtime being bounded can be charged to the total variation distance, via a union bound.
\end{proof}

We put together the pieces to obtain our desired estimator for normalizing constants.

\begin{corollary}\label{cor:normalization}
Suppose that $\ma \in \PSD^{k \times k}$ satisfies $\mu \id_k \preceq \ma \preceq L \id_k$, and that $\vb \in \R^k$ satisfies $\norm{\vb}_2 \le R$. Fix an approximation tolerance $\Delta \in (0, 1)$ and a failure probability $\delta \in (0, 1)$. There is an algorithm for computing $\tZ$ satisfying \eqref{eq:approx_z}, with probability $\ge 1 - \delta$, using time
\[O\Par{\frac{Lk^4}{\mu\Delta^2}\log^4\Par{\frac{LRk}{\mu\Delta}}\log\Par{\frac 1 \delta}}.\]
\end{corollary}
\begin{proof}
We instantiate Proposition~\ref{prop:est_Z} with $\sig_1^{-2} \gets \lam$ satisfying \eqref{eq:lam_lower_bound}, and $\alg \gets $ the sampler from Proposition~\ref{prop:composite_sample}. In particular, for any $i \in [M]$, the density $\propto \exp(-f_i)$ has the structure required by Proposition~\ref{prop:composite_sample}, where in the statement of Proposition~\ref{prop:composite_sample}, we set
\[f(\vth) \gets \half \vth^\top \mm \vth - \vb^\top \vth + \frac{1}{2\sig_i^2}\norm{\vth}_2^2,\; g(\vth) \gets \norm{\vth}_1.\]
Observe that for any $i \in [M]$, the parameter $\kappa$ in Proposition~\ref{prop:composite_sample} satisfies
\[\kappa \le \frac{L + \frac 1 {\sig_i^2}}{\mu + \frac 1 {\sig_i^2}} \le \frac L \mu.\]
Thus, the time complexity of each call to $\alg$ is at most
\[O\Par{\frac{Lk^3}{\mu} \log^3\Par{\frac{NLk}{\mu}}}, \text{ for } N = \Theta\Par{\frac{k}{\Delta^2}\log\Par{\frac{k}{\mu\sig_1^2}}} = \Theta\Par{\frac{k}{\Delta^2}\log\Par{\frac{kR}{\mu\Delta}}},\]
since the cost of computing $\nabla f(\vth)$ is $O(k^2)$ time. The conclusion follows from Proposition~\ref{prop:est_Z}.
\end{proof}

\subsection{Approximate centered rejection sampling}\label{ssec:reject_laplace}

In this section, we give the analog of Algorithm~\ref{alg:sample_tpi} in the Laplace prior setting.

\begin{algorithm}[H]
    \caption{$\productsl(\mx, \vy, \vhth, \sigma, \vq, k^\star)$}
    \label{alg:sample_tpi_laplace}
    \begin{algorithmic}[1]
    \State \textbf{Input:} $\mx \in \R^{n \times d}$, $\vy \in \R^n$, $\sigma > 0$, $\vq \in [0, 1]^d$ produced by Model~\ref{model:sas_basic}, $\vhth \in \R^d$, $k^\star \in \N$
    \State \textbf{Output:} Sample $S$ from a conditional Poisson distribution over
    \begin{equation}\label{eq:Omega_vth_k_laplace} S \in \Omega_{\vhth, k^\star} \defeq \Brace{S \subseteq [d] \mid S \supseteq \supp(\vhth), |S| \le k^\star}\end{equation}
    \State $\vu \gets \frac{1}{\sigma^2}\mx^\top(\vy - \mx \vhth)$
        \State $T \gets \supp(\vhth)$\;
        \For {$i \in T^c$}
            \State $\vv_i^- \gets \frac 1 {\sqrt {2\pi}}\int_{\bbR}\exp\paren{-\frac{1+\epsilon}{2\sigma^2} x^2+\vu_i x-|x|}\diff x$
            \State $\vr_i \gets \vq_i\cdot \vvm_i$
            \State $\vp_i \gets \frac{\vr_i}{1 - \vq_i + \vr_i}$
        \EndFor
        \State \Return $T \cup \cps(\vp, k^\star - |T|)$
    \end{algorithmic}
\end{algorithm}

\begin{lemma}
\label{lem:rejection_rate_bound_laplace}
In the setting of Model~\ref{model:sas_basic}, take $\mu = \Lap(0, 1)$. 
Let $\vhth\in \bbR^d$, and let $\tilde{\pi}_{\supp}(\cdot) := \textup{Law}(S)$ where $S \in \Omega_{\vhth, k^\star}$ (defined in \eqref{eq:Omega_vth_k_laplace}) is a sample from Algorithm~\ref{alg:sample_tpi_laplace}.
Then, 
\begin{equation}\label{eq:def_Q_S_laplace}\tpisupp(S) \propto \exp\left(-\norms{\vhth}_1\right)\prod_{i\in S} \paren{\frac{\vq_i \vvm_i}{1 - \vq_i}} =: Q(S),\end{equation}
and further, for $\vr \defeq \vy - \mx \vhth$, the conditional distribution of $\pisupp(\cdot \mid \mx, \vy)$ over $\Omega_{\vhth, k^\star}$ satisfies
\begin{equation}\label{eq:def_P_S_laplace}
\begin{aligned}
\pisupp\Par{S \mid \mx, \vy, S \in \Omega_{\vhth, k^\star}} &\propto \paren{\frac{1}{2\pi}}^{\frac{|S|}{2}}\paren{\prod_{i\in S} \frac{\vq_i}{1 - \vq_i}}\\
&\cdot \int_{\bbR^{S}} \exp\paren{-\frac{1}{2\sigma^2}\twonorm{\mx_{:S}\vDelta_S - \vr}^2 - \norm{\vDelta_S + \vhth_S}_1} \diff \vDelta_S \\
&=: P(S).
\end{aligned}
\end{equation}
Furthermore, suppose that $\mx$ is $(\eps, k^\star)$-RIP for $\eps \in (0, \half)$. Then if $\sig \le \frac 1 4$, for any $S \in \Omega_{\vhth, k^\star}$, 
\begin{equation}\label{eq:PQ_bound_laplace}
   1\leq \frac{P(S)}{Q(S)}\leq \exp\paren{\frac{2\sigma^2\sqrt{k^\star\|\vu_S\|^2_2} + 2\sigma^2\epsilon(\|\vu_S\|^2_2 + k^\star)}{1 - \epsilon^2}+\frac{6k^\star\sigma}{\sqrt{1-\epsilon}} + \frac{k^\star \epsilon}{1 - \epsilon}}
.\end{equation}
\bk
\end{lemma}

\begin{proof}
The proof that \eqref{eq:def_Q_S_laplace} holds proceeds analogously to the proof that \eqref{eq:def_Q_S} holds in Lemma~\ref{lem:unnorm_post}. We next prove \eqref{eq:def_P_S_laplace} holds. Following an analogous derivation to \eqref{eq:pisupp_post} but using $\mu = \Lap(0, 1)$,
\begin{equation}\label{eq:P_derive_laplace}
\begin{aligned}
\pisupp\Par{S \mid \mx, \vy, S \in \Omega_{\vhth, k^\star}} &\propto \Par{\frac 1 {2\pi}}^{\frac{|S|}{2}}\Par{\prod_{i \in S}\frac{\vq_i}{1 - \vq_i}} \\
&\cdot \int_{\R^S}\exp\Par{-\frac 1 {2\sig^2}\norm{\vy - \mx_{:S} \vth_S}_2^2 - \norm{\vth_S}_1} \dd \vth_S \\
&= \Par{\frac 1 {2\pi}}^{\frac{|S|}{2}}\Par{\prod_{i \in S}\frac{\vq_i}{1 - \vq_i}} \\
&\cdot \int_{\R^S}\exp\Par{-\frac 1 {2\sig^2}\norm{\mx_{:S}\vDelta_S - \vr}_2^2 - \norm{\vDelta_S + \vhth_S}_1} \dd \vDelta_S,
\end{aligned}
\end{equation}
where, performing the change of variables $\vDelta = \vth - \vhth$, we have 
\begin{align*}
    \mx_{:S}\vth_S - \vy &= \mx_{:S}(\vDelta_S + \vhth) - \vy = \mx_{:S}\vDelta_S - (\vy - \mx_{:S}\vhth_{S})= \mx_{:S} \vDelta_S - \vr,
\end{align*}
and the last equality uses $\supp(\vhth) \subseteq S$.
It remains to prove \eqref{eq:PQ_bound_laplace}. First, define for all $i \in [d]$,
\begin{align*}\vvm_i &\defeq \frac 1 {\sqrt{2\pi}}\int_{\R}\exp\Par{-\frac{1+\eps}{2\sig^2}x^2 + \vu_i x - |x|} \dd x,\\ \vvp_i &\defeq \frac 1 {\sqrt{2\pi}} \int_{\R}\exp\Par{-\frac{1-\eps}{2\sig^2}x^2 + \vu_i x + |x|}\dd x.
\end{align*}
Since $[\mx^\top\mx]_{S \times S} \preceq (1 + \eps) \id_S$, we have the lower bound
\begin{align*}
    P(S) &= \paren{\frac{1}{2\pi}}^{\frac{|S|}{2}}\paren{\prod_{i\in S} \frac{\vq_i}{1 - \vq_i}} \int_{\R^S} \exp\paren{-\frac{1}{2\sigma^2}\twonorm{\mx_{:S}\vDelta_S - \vr}^2 - \norm{\vDelta_S + \vhth_S}_1} \diff \vDelta_S\\
    &= \paren{\frac{1}{2\pi}}^{\frac{|S|}{2}}\paren{\prod_{i\in S} \frac{\vq_i}{1 - \vq_i}} \int_{\bbR^S} \exp\left(-\frac{1}{2\sigma^2}\norm{\vDelta_S}^2_{[\mxtx]_{S\times S}} + \vu_S^\top \vDelta_S  - \onenorm{\vhth_S+\vDelta_S}\right)\diff \vDelta_S\\
    &\stackrel{(a)}{\geq} \exp\left(-\norms{\vhth}_1\right) \paren{\frac{1}{2\pi}}^{\frac{|S|}{2}}\paren{\prod_{i\in S} \frac{\vq_i}{1 - \vq_i}} \int_{\bbR^S} \exp\paren{-\frac{1+\epsilon}{2\sigma^2}\twonorm{\vDelta_S}^2 + \vu_S^\top\vDelta_S-\|\vDelta_S\|_1}\diff \vDelta_S\\
    &= \exp\left(-\norms{\vhth}_1\right)\prod_{i\in S} \paren{\frac{\vq_i \vvm_i}{1 - \vq_i}},
\end{align*}
where $(a)$ is by the triangle inequality on $\onenorm{\cdot}$ and the fact that $\norms{\vhth}_1 = \norms{\vhth_S}_1$  as $\supp(\vhth)\subseteq S$. 

Similarly, since $[\mx^\top \mx]_{S \times S} \succeq (1 - \eps)\id_S$, we have the upper bound
\begin{align*}
    P(S) &= \paren{\frac{1}{2\pi}}^{\frac{|S|}{2}}\paren{\prod_{i\in S} \frac{\vq_i}{1 - \vq_i}}\int_{\bbR^{S}} \exp\paren{-\frac{1}{2\sigma^2}\twonorm{\mx_{:S}\Delta_S - \vr}^2 - \norm{\Delta_S + \vhth_S}_1} \diff \vDelta_S\\
    &\le \exp\left(-\norms{\vhth}_1\right)\paren{\frac{1}{2\pi}}^{\frac{|S|}{2}}\paren{\prod_{i\in S} \frac{\vq_i}{1 - \vq_i}} \int_{\R^S}\exp\Par{-\frac{1-\eps}{2\sig^2}\norm{\vDelta_S}_2^2 + \vu_S^\top \vDelta_S + \norm{\vDelta_S}_1} \dd \vDelta_S \\
    &= \exp\left(-\norms{\vhth}_1\right)\prod_{i\in S} \paren{\frac{\vq_i \vvp_i}{1 - \vq_i}}.
\end{align*}

Now, by the definition of $Q$ in \eqref{eq:def_Q_S_laplace}, we have
\begin{align*}
    1\leq \frac{P(S)}{Q(S)}\leq \prod_{i\in S}\frac{\vv_i^+}{\vv_i^-}.
\end{align*}
Next, we derive explicit formulas for $\vv_i^+$ and $\vv_i^-$. Let 
\[\sigm \defeq \frac{\sigma}{\sqrt{1 - \epsilon}},\; \sigp \defeq \frac{\sigma}{\sqrt{1 + \epsilon}},\; \vup_i \defeq \vu_i + 1,\; \vum_i \defeq \vu_i - 1.\]
We have 
\begin{align*}
    \vv_i^- &= \frac 1 {\sqrt{2\pi}}\Par{\int_{x \geq 0}\exp\paren{-\frac{1}{2\sigma_+^2}x^2 + \vu_i^-x}\diff x + \int_{x \leq 0}\exp\paren{-\frac{1}{2\sigma^2_+}x^2 + \vu_i^+x}\diff x}\\
    &= \frac 1 {\sqrt{2\pi}}\exp\paren{\frac{\sigma^2_+(\vu_i^-)^2}{2}}\int_{x \geq 0}\exp\paren{-\frac{(x - \sigma_+^2\vu_i^-)^2}{2\sigma_+^2}}\diff x \\
    &+\frac 1 {\sqrt{2\pi}}\exp\paren{\frac{\sigma^2_+(\vu_i^+)^2}{2}}\int_{x \leq 0}\exp\paren{-\frac{(x - \sigma_+^2\vu_i^+)^2}{2\sigma_+^2}}\diff x \\
    &= \sigma_+ \paren{\exp\paren{\frac{\sigma^2_+(\vu_i^-)^2}{2}} \bbP\paren{\calN(\sigma_+^2 \vu_i^-, \sigma_+^2) \geq 0} + \exp\paren{\frac{\sigma^2_+(\vu_i^+)^2}{2}} \bbP\paren{\calN(\sigma_+^2 \vu_i^+, \sigma_+^2) \leq 0}}.
\end{align*}
Similarly, we have 
\begin{equation*}
    \vv_i^+ = \sigma_- \paren{\exp\paren{\frac{\sigma^2_-(\vu_i^+)^2}{2}} \bbP\paren{\calN(\sigma_-^2 \vu_i^+, \sigma_-^2) \geq 0} + \exp\paren{\frac{\sigma^2_-(\vu_i^-)^2}{2}} \bbP\paren{\calN(\sigma_-^2 \vu_i^-, \sigma_-^2) \leq 0}}.
\end{equation*}

Using the fact that $\bbP[\calN(\mu,\sigma^2)\leq 0] =\bbP [Z\leq -\frac \mu \sig]$, where $Z\sim\Nor(0, 1)$, we have:
\begin{align*}
 \vv_i^+ &= \sigma_- \paren{\exp\paren{\frac{\sigma^2_-(\vu_i^+)^2}{2}} \bbP\Brack{Z \geq -\sigma_-\vu_i^+} + \exp\paren{\frac{\sigma^2_-(\vu_i^-)^2}{2}} \bbP\Brack{Z \leq -\sigma_-\vu_i^-}},\\
    \vv_i^- &= \sigma_+ \paren{\exp\paren{\frac{\sigma^2_+(\vu_i^-)^2}{2}} \bbP\Brack{Z \geq -\sigma_+\vu_i^-} + \exp\paren{\frac{\sigma^2_+(\vu_i^+)^2}{2}} \bbP\Brack{Z  \leq -\sigma_+\vu_i^+}}.
\end{align*}
Thus, we have 
\begin{align*}
    \frac{\vv_i^+}{\vv_i^-} \leq \sqrt{\frac{1 + \epsilon}{1 - \epsilon}} \cdot \frac{\max\Par{\exp\paren{\frac{\sigma^2_-(\vu_i^+)^2}{2}}, \exp\paren{\frac{\sigma^2_-(\vu_i^-)^2}{2}}} \paren{\bbP\Brack{Z \geq -\sigma_-\vu_i^+} + \bbP\Brack{Z \leq -\sigma_-\vu_i^-}}}{\min\Par{\exp\paren{\frac{\sigma^2_+(\vu_i^-)^2}{2}}, \exp\paren{\frac{\sigma^2_+(\vu_i^+)^2}{2}}} \paren{\bbP\Brack{Z \geq -\sigma_+\vu_i^-} + \bbP\Brack{Z  \leq -\sigma_+\vu_i^+}}}.
\end{align*}
We begin by bounding the exponential terms in the above display:
\begin{align*}
    \frac{\exp\paren{\frac{\sigma^2_-(\vu_i^+)^2}{2}}}{\exp\paren{\frac{\sigma^2_+(\vu_i^-)^2}{2}}} &= \exp \paren{\frac{\sigma^2}{2}\paren{\frac{(\vu_i + 1)^2}{1 - \epsilon} - \frac{(\vu_i - 1)^2}{1 + \epsilon}}}\\
    &= \exp\paren{\frac{\sigma^2}{1 - \epsilon^2} \paren{2\vu_i - \epsilon (\vu_i^2 + 1)}} \leq \exp\paren{\frac{2\sigma^2|\vu_i|}{1 - \epsilon^2}},\\
    \frac{\exp\paren{\frac{\sigma^2_-(\vu_i^-)^2}{2}}}{\exp\paren{\frac{\sigma^2_+(\vu_i^+)^2}{2}}} &= \exp \paren{\frac{\sigma^2}{2}\paren{\frac{(\vu_i - 1)^2}{1 - \epsilon} - \frac{(\vu_i + 1)^2}{1 + \epsilon}}}\\
    &= \exp\paren{\frac{\sigma^2}{(1 - \epsilon^2)} \paren{-2\vu_i + \epsilon (\vu_i^2 + 1)}} \leq \exp\paren{\frac{2\sigma^2|\vu_i| + \sigma^2\epsilon(\vu_i^2 + 1)}{1 - \epsilon^2}},\\
    \frac{\exp\paren{\frac{\sigma^2_-(\vu_i^+)^2}{2}}}{\exp\paren{\frac{\sigma^2_+(\vu_i^+)^2}{2}}} &= \exp \paren{\frac{\sigma^2(\vu_i + 1)^2}{2}\paren{\frac{1}{1 - \epsilon} - \frac{1}{1 + \epsilon}}}\\
    &\leq \exp\paren{\frac{\sigma^2\epsilon(\vu_i + 1)^2}{1 - \epsilon^2}} \leq \exp\paren{\frac{2\sigma^2\epsilon(\vu_i^2 + 1)}{1 - \epsilon^2}},\\
    \frac{\exp\paren{\frac{\sigma^2_-(\vu_i^-)^2}{2}}}{\exp\paren{\frac{\sigma^2_+(\vu_i^-)^2}{2}}} &= \exp \paren{\frac{\sigma^2(\vu_i - 1)^2}{2}\paren{\frac{1}{1 - \epsilon} - \frac{1}{1 + \epsilon}}}\\
    &\leq \exp\paren{\frac{\sigma^2\epsilon (\vu_i - 1)^2}{1 - \epsilon^2}} \leq \exp\paren{\frac{2\sigma^2\epsilon (\vu_i^2 + 1)}{1 - \epsilon^2}}.
\end{align*}
Therefore, combining these cases, we have 
\begin{equation*}
    \frac{\max\Par{\exp\paren{\frac{\sigma^2_-(\vu_i^+)^2}{2}}, \exp\paren{\frac{\sigma^2_-(\vu_i^-)^2}{2}}}}{\min\Par{\exp\paren{\frac{\sigma^2_+(\vu_i^-)^2}{2}}, \exp\paren{\frac{\sigma^2_+(\vu_i^+)^2}{2}}}} \leq \exp\paren{\frac{2\sigma^2|\vu_i| + 2\sigma^2\epsilon(\vu_i^2 + 1)}{1 - \epsilon^2}}.
\end{equation*}

Next we consider the ratio of probabilities:
\begin{align*}
    \bbP\Brack{Z \geq -\sigma_-\vu_i^+} + \bbP\Brack{Z \leq -\sigma_-\vu_i^-} &= 1+\bbP\Brack{Z\in \bra{-\sigma_-\vu_i^+, -\sigma_-\vu_i^-}} \leq 1 + 2\sigma_-,\\
    \bbP\Brack{Z \geq -\sigma_+\vu_i^-} + \bbP\Brack{Z \leq -\sigma_+\vu_i^+} &= 1-\bbP\Brack{Z\in \bra{-\sigma_+\vu_i^-, -\sigma_+\vu_i^+}}\geq 1 - 2\sigma_+,
\end{align*}

since the PDF of $\Nor(0, 1)$ is pointwise bounded by $1$. Combining, we have shown
\begin{align*}
    \frac{\vv_i^+}{\vv_i^-}&=\sqrt{\frac{1 +\epsilon}{1 - \epsilon}}
        \exp\paren{\frac{2\sigma^2|\vu_i| + 2\sigma^2\epsilon(\vu_i^2 + 1)}{1 - \epsilon^2}}
        \frac{1+\frac{2\sigma}{\sqrt{1-\epsilon}}}{1-\frac{2\sigma}{\sqrt{1+\epsilon}}}.
\end{align*}
Taking a product over the $|S| \le k^\star$ coordinates, and using $\frac{1}{1-x}\le 1 + \frac x 2$ for $x \in (0, \half)$, as well as $\sqrt{\frac{1+\epsilon}{1-\epsilon}}=\sqrt{1+\frac{2\epsilon}{1-\epsilon}}\leq \exp(\frac{\epsilon}{1-\epsilon})$, we have the desired result:
\bas{
\prod_{i\in S}\frac{\vv_i^+}{\vv_i^-}&\leq \exp\paren{\sum_{i\in S}\frac{2\sigma^2|\vu_i| + 2\sigma^2\epsilon(\vu_i^2 + 1)}{1 - \epsilon^2}}
        \exp\left(\frac{2k^\star\sigma}{\sqrt{1-\epsilon}}+\frac{4k^\star \sigma}{\sqrt{1+\epsilon}}\right)\exp\left(\frac{k^*\epsilon}{1-\epsilon}\right)\\
        &\leq \exp\paren{ \frac{2\sigma^2\sqrt{k^\star \|\vu_S\|^2_2} + 2\sigma^2\epsilon(\|\vu_S\|^2_2 + k^\star)}{1 - \epsilon^2}+\frac{6k^\star \sigma}{\sqrt{1-\epsilon}}+\frac{k^\star \epsilon}{1-\epsilon}}.
}

\end{proof}
Our posterior sampler uses rejection sampling with the unnormalized densities described in \eqref{eq:def_P_S_laplace}, \eqref{eq:def_Q_S_laplace}. However, one difficulty is that unlike in the Gaussian case, there is no explicit formula (e.g., Fact~\ref{fact:gauss_norm}) for $P(S)$. In Section~\ref{ssec:anneal_laplace}, we gave an algorithm for estimating $P(S)$ to multiplicative error, based on sampling access to the induced distribution over $\vth_S \in \R^S$. To facilitate using these approximate evaluations of $P$, we provide the following helper tool.

\begin{lemma}\label{lem:tv_mult_approx}
Let $\pi, \tpi$ be distributions over the some domain $\Omega$, and suppose that $\pi \propto P$ and $\tpi \propto \tP$ for unnormalized densities $P, \tP$. Moreover, suppose that for all $\omega \in \Omega$,
\[\frac {P(\omega)} {\tP(\omega)} \in \Brack{1 - \Delta, 1 + \Delta}\]
for some $\Delta \in (0, \half)$. Then, $\tv{\pi, \tpi} \le \frac{3\Delta}{2}$.
\end{lemma}
\begin{proof}
Throughout the proof, denote the relevant normalization constants by
\[Z \defeq \int_\Omega P(\omega) \dd \omega,\; \tZ \defeq \int_\Omega \tP(\omega) \dd \omega,\]
so the assumption yields $Z \in [(1 - \Delta) \tZ, (1 + \Delta) \tZ]$, and thus $\frac 1 {Z} \in [\frac 1 {(1 + \Delta)\tZ}, \frac 1 {(1 - \Delta)\tZ}] \subseteq [\frac{1 - 2\Delta}{\tZ}, \frac{1 + 2\Delta}{\tZ}]$.
Then, we have the claim by applying the triangle inequality:
\begin{align*}
\tv{\pi, \tpi} &= \half \int_\Omega \Abs{\frac{P(\omega)}{Z} - \frac{\tP(\omega)}{\tZ}} \dd \omega \\
&\le \half \int_\Omega \Abs{\frac{P(\omega)}{Z} - \frac{\tP(\omega)}{Z}} \dd \omega + \half \int_\Omega \Abs{\frac{\tP(\omega)}{Z} - \frac{\tP(\omega)}{\tZ}} \dd \omega \\
&\le \frac \Delta 2 \int_\Omega \frac{P(\omega)}{Z} \dd \omega + \frac{2\Delta}{2} \int_\Omega \frac{\tP(\omega)}{\tZ}\dd \omega = \frac{3\Delta}{2}.
\end{align*}
\end{proof}
Finally, by combining these developments, we obtain the analog of Proposition~\ref{prop:post_supp_sample}.

\begin{algorithm}
    \caption{$\postsl(\mx, \vy, \vhth, \sigma, \vq, k^\star, \delta)$}
    \label{alg:post_sample_laplace}
    \begin{algorithmic}[1]
        \State \textbf{Input:} $\mx \in \R^{n \times d}$, $\vy \in \R^n$, $\sig > 0$, $\vq \in [0, 1]^d$ produced by Model~\ref{model:sas_basic} with $\mu = \Lap(0, 1)$, $\vhth \in \R^d$, $k^\star \in \N$, $\delta \in (0, 1)$
        \State \textbf{Output:} Sample $S$ approximately distributed as $ \pisupp(\cdot \mid \mx, \vy)$ (cf.\ Proposition~\ref{prop:post_supp_sample_laplace})
        \State $\mu \gets \productsl(\mx, \vy, \vhth, \sigma, \vq, \frac 3 4 k^\star)$\;
        \State $P, Q \gets $ unnormalized distributions in \eqref{eq:def_P_S_laplace}, \eqref{eq:def_Q_S_laplace}\;
        \State $\Omega \gets \Omega_{\vhth, \frac 3 4 k^\star}$ defined in \eqref{eq:Omega_vth_k_laplace}
        \State $\tP \gets $ unnormalized distribution satisfying $P \in [(1 - \frac \delta {12})\tP, (1 + \frac \delta {12}) \tP]$ over $\Omega$, using Corollary~\ref{cor:normalization}\label{line:tp_def}
        \State \Return $\rejectionsample(\mu, \tP, Q, \Omega, 4, \frac \delta 8)$
    \end{algorithmic}
\end{algorithm}

\begin{proposition}\label{prop:post_supp_sample_laplace}
In the setting of Model~\ref{model:sas_basic} where $\mu = \Lap(0, 1)$, let $\vhth$ be produced as in Corollary~\ref{cor:vhth_gaussian} with $\delta \gets \frac \delta 4$, let $k^\star$, $\eps$ satisfy \eqref{eq:keps_bound}, and let
$\sig \le \frac 1 {6k^\star}$. 
Then if $\mx$ is $(\eps, C_{\alg} k^\star)$-RIP for the universal constant $C_{\alg}$ from Assumption~\ref{assume:sr_linf}, with probabilty $\ge 1 - \delta$ over the randomness of Model~\ref{model:sas_basic}, Algorithm~\ref{alg:post_sample_laplace} returns $S \sim \nu$ satisfying $\tv{\nu, \pisupp(\cdot \mid \mx, \vy)} \le \delta$. If $\vhth$ is given, Algorithm~\ref{alg:post_sample_laplace} runs in time 
\[O\Par{nd + n\Par{k^\star}^{\omega - 1}\log\Par{\frac 1 \delta}+ \frac{(k^\star)^4}{\delta^2}\log^4\Par{\frac{1}{\sig}\log\Par{\frac d \delta}}\log^2\Par{\frac 1 \delta}}.\]
\end{proposition}
\begin{proof}
As in the proof of Proposition~\ref{prop:post_supp_sample}, with probability $\ge 1 - \delta$ over the randomness of Model~\ref{model:sas_basic}, all of the bounds in \eqref{eq:goodmodel1} and \eqref{eq:goodmodel2} hold, except for the second bound in \eqref{eq:goodmodel1} which is not used in this proof. In particular, this shows that with probability $\ge 1 - \delta$ over Model~\ref{model:sas_basic},
\[\norm{\vu_S}_2^2 \le \frac{16k^\star}{\sig^2}\Par{R_{\alg}\Par{k^\star, \eps, \frac{\delta^2}{320}} + \log\Par{\frac{40d}{\delta}}},\]
for all $S \in \Omega_{\vhth, \frac 3 4 k^\star}$. Thus, by Lemma~\ref{lem:rejection_rate_bound_laplace},
\[\frac{P(S)}{Q(S)} \le \exp\paren{\frac{2\sigma^2\sqrt{k^\star\|\vu_S\|^2_2} + 2\sigma^2\epsilon(\|\vu_S\|^2_2 + k^\star)}{1 - \epsilon^2}+\frac{6k^\star\sigma}{\sqrt{1-\epsilon}} + \frac{k^\star \epsilon}{1 - \epsilon}} \le 3,\]
for our choice of parameters in \eqref{eq:keps_bound} and $\sig \le \frac 1 {6k^\star}$. Thus, we can check that any $\tP$ satisfying the criterion on Line~\ref{line:tp_def} over $\Omega_{\vhth, \frac 3 4 k^\star}$ has
\[\frac 1 4 \le \frac{\tP(S)}{Q(S)} \le 4, \text{ for all } S \in \Omega_{\vhth, \frac 3 4 k^\star}.\]
Therefore, Lemmas~\ref{lem:reject} and~\ref{lem:tv_mult_approx} show that Algorithm~\ref{alg:post_sample_laplace} samples within total variation $\frac \delta 4$ from $\pisupp(S \mid \mx, \vy, S \in \Omega_{\vhth, \frac 3 4 k^\star})$, which is in turn within total variation $\frac \delta 2$ from $\pisupp(S \mid \mx, \vy)$. Our argument allows for a failure probability of $O(\frac{\delta}{\log(1/\delta)})$ in each computation of $\tP$, so that all calls to $\tP$ used by Lemma~\ref{lem:reject} succeed with probability $\ge 1 - \frac \delta 4$.\footnote{To make sure our algorithm is consistent in its choice of $\tP$, we store all $S$ sampled by Algorithm~\ref{alg:sample_tpi_laplace}, along with the estimates $\tP(S)$ computed. This does not dominate the asymptotic space or time complexity.} This concludes the correctness proof. 

For the runtime, we proceed similarly to Proposition~\ref{prop:post_supp_sample}, except we substitute the cost of computing $[\mxtx]_{S \times S}$ and $\tP$ to the required accuracy by using Corollary~\ref{cor:normalization} in each iteration, with $\delta \gets O(\frac{\delta}{\log(1/\delta)})$, $\frac L \mu = O(1)$ (due to RIP), and for $\vb \defeq \frac 1 {\sig^2} \mx_{S:}^\top \vy$,
\begin{align*}
\norm{\vb}_2 \le \frac 1 {\sig^2} \norm{\mx_{S:}^\top \vy} = O\Par{\textup{poly}\Par{k^\star, \log\Par{\frac d \delta}, \frac 1 \sig}}.
\end{align*}
The above holds by using our bound on $\norm{\mx^\top \vxi}_\infty$ in the proof of Lemma~\ref{lem:post_concentration}, and because 
\begin{equation}\label{eq:laplace_bound}
\norm{\vths}_2^2 = O\Par{k^\star \log^2\Par{\frac d \delta}}
\end{equation}
within the failure probability on Model~\ref{model:sas_basic}, by standard tail bounds on Laplace random variables.
\end{proof}

\subsection{Proof of Theorem~\ref{thm:sas_laplace}}\label{ssec:proof_laplace}

We now combine Propositions~\ref{prop:sparse_recovery} and~\ref{prop:post_supp_sample_laplace} to provide our main sampling result under a Laplace prior.

\begin{theorem}\label{thm:sas_laplace}
Let $\delta \in (0, 1)$. In the setting of Model~\ref{model:sas_basic}, suppose $\mu = \Lap(0, 1)$, $\sig = O(\frac 1 {k^\star})$, and that $\mx$ is $(\eps, k^\star)$-RIP, where
\[k^\star = \Omega\Par{k + \log\Par{\frac 1 \delta}},\; \eps = O\Par{\frac 1 {\Par{k + \log\Par{\frac 1 \delta}}\log\Par{\frac d \delta}}},\]
for appropriate constants. There is an algorithm (Algorithm~\ref{alg:post_sample_laplace} using $\vhth$ produced by $\alg_\infty$ in Proposition~\ref{prop:sparse_recovery}) that returns $\vth \sim \pi'$ for a distribution $\pi'$ satisfying $\tv{\pi', \pi(\cdot \mid \mx, \vy)} \le \delta$, with probability $\ge 1 - \delta$ over the randomness of Model~\ref{model:sas_basic}. The algorithm runs in time
\[O\Par{d^{1.5}n^2\log\Par{\frac{d\kappa\log\Par{\frac 1 \delta}}{\sigma}} + \frac{(k^\star)^4}{\delta^2}\log^4\Par{\frac{\log\Par{\frac d \delta}}{\sig}}\log^2\Par{\frac 1 \delta}}.\]
Alternatively, suppose $\mu = \Lap(0, 1)$, $\sig = O(\frac 1 {k^\star})$, $\mx$ is $(\eps, k^\star)$-RIP, and for some $m \in [n]$, $\sqrt{m/n} \cdot \mx_{[m]:}$ is $(\frac 1 {10}, k^\star)$-RIP where
\[k^\star = \Omega\Par{k + \log\Par{\frac 1 \delta}},\; \eps = O\Par{\frac{1}{\Par{k + \log\Par{\frac 1 \delta}}\Par{m + \log\Par{\frac 1 \delta}}}},\]
for appropriate constants. There is an algorithm (Algorithm~\ref{alg:post_sample_laplace} using $\vhth$ produced by $\alg_2$ in Proposition~\ref{prop:sparse_recovery}) that returns $\vth \sim \pi'$ for a distribution $\pi'$ satisfying $\tv{\pi', \pi(\cdot \mid \mx, \vy)} \le \delta$, with probability $\ge 1 - \delta$ over the randomness of Model~\ref{model:sas_basic}. The algorithm runs in time
\[O\Par{nd\log\Par{\frac{\log\Par{\frac d \delta}}{\sig}} + n\Par{k^\star}^{\omega - 1}\log\Par{\frac 1 \delta}+ \frac{(k^\star)^4}{\delta^2}\log^4\Par{\frac{\log\Par{\frac d \delta}}{\sig}}\log^2\Par{\frac 1 \delta}}.\]
\end{theorem}
\begin{proof}
The proof is almost identical to the proof of Theorem~\ref{thm:sas_basic}, where we substitute Proposition~\ref{prop:post_supp_sample_laplace} in place of Proposition~\ref{prop:post_supp_sample}, which also gives the upper bound on $\sig$. We summarize the remaining differences here. First, when using $\alg_\infty$, we use the same choice of $r_\infty$ as in Theorem~\ref{thm:sas_basic}, and
\[R_\infty = \Theta\Par{r_\infty + k^\star \log^2\Par{\frac d \delta}}\]
by substituting the bound \eqref{eq:laplace_bound}. Second, when using $\alg_2$, we similarly use the bound $R_2^2 = O(k^\star\log^2(\frac d \delta))$ as before. Finally, in the Laplace prior setting, we no longer have an exact characterization for sampling $\vth \sim \pi(\mx, \vy, S)$, as in Lemma~\ref{lem:theta_sample}. However, using Proposition~\ref{prop:composite_sample} to perform this sampling to total variation error $\frac \delta 2$ does not dominate the runtime, and it suffices to adjust the failure probability of other steps of the algorithm by a constant factor.
\end{proof}

For convenience, we provide the following specialization of Theorem~\ref{thm:sas_laplace} in the specific case of Gaussian measurements $\mx \in \R^{n \times d}$ with i.i.d.\ entries $\sim \Nor(0, \frac 1 n)$, analogously to Corollary~\ref{cor:gaussian_X}.

\begin{corollary}\label{cor:gaussian_X_laplace}
Let $\delta \in (0, 1)$, and suppose $\mx \in \R^{n \times d}$ has i.i.d.\ entries $\sim \Nor(0, \frac 1 n)$. If
\[n = \Omega\Par{\Par{k + \log\Par{\frac 1 \delta}}^3 \log^3\Par{\frac d \delta}},\]
for an appropriate constant, in the setting of Model~\ref{model:sas_basic} where $\mu = \Lap(0, 1)$ and $\sig = O(\frac 1 {k + \log(1/\delta)})$, there is an algorithm that returns $\vth \sim \pi'$ for a distribution $\pi'$ satisfying $\tv{\pi', \pi(\cdot \mid \mx, \vy)} \le \delta$, with probability $\ge 1 - \delta$ over the randomness of Model~\ref{model:sas_basic} and $\mx$. The algorithm runs in time
\[O\Par{n^2 d^{1.5}\log\Par{\frac{d\log\Par{\frac 1 \delta}}{\sig}} + n^2\sqrt d \log\Par{\frac{d}{\sig\delta}} + \frac{\Par{k + \log(\frac 1 \delta)}^4}{\delta^2}\log^4\Par{\frac{\log\Par{\frac d \delta}}{\sig}}\log^2\Par{\frac 1 \delta}}.\]
Alternatively, if
\[n = \Omega\Par{\Par{k + \log\Par{\frac 1 \delta}}^5 \log^2\Par{\frac d \delta}\log(d)}\]
for an appropriate constant, in the setting of Model~\ref{model:sas_basic} where $\mu = \Lap(0, 1)$ and $\sig = O(\frac 1 {k + \log(1/\delta)})$, there is an algorithm that returns $\vth \sim \pi'$ for a distribution $\pi'$ satisfying $\tv{\pi', \pi(\cdot \mid \mx, \vy)} \le \delta$, with probability $\ge 1 - \delta$ over the randomness of Model~\ref{model:sas_basic} and $\mx$. The algorithm runs in time
\[O\Par{nd\log\Par{\frac{\log\Par{\frac d \delta}}{\sig}} + \frac{\Par{k + \log(\frac 1 \delta)}^4}{\delta^2}\log^4\Par{\frac{\log\Par{\frac d \delta}}{\sig}}\log^2\Par{\frac 1 \delta}}.\]
\end{corollary}

\section*{Acknowledgements}

We would like to thank Eric Price for the pointer to Lemma~\ref{lem:L_infty_bound_RIP}, Arun Jambulapati for suggesting the approach in Appendix~\ref{ssec:lasso_solve}, and Trung Dang for the pointer to Lemma~\ref{lemma:conditional_sampling}. KT would like to thank Sameer Deshpande for helpful conversations on the spike-and-slab sampling literature.

\newpage

\bibliography{ref}
\bibliographystyle{alpha}

\newpage
\appendix

\section{Helper results on sparse recovery}
\label{app:sparse}

\subsection{Sparse recovery preliminaries}\label{app:isometric_cond}

\paragraph{Concentration.} Here we state standard bounds on Gaussian random variables.

\begin{fact}[Mill's inequality]\label{fact:gauss_tail_bound}\label{fact:mills}
If $Z \sim \Nor(0, 1)$ is a standard Gaussian random variable and $t > 0$,
\[\Pr\Brack{|Z| > t} \le \sqrt{\frac 2 \pi} \cdot \frac 1 t \exp\Par{-\frac{t^2}{2}}.\]
\end{fact}

\begin{fact}[$\chi^2$ tail bounds, Lemma 1, \cite{LaurentM00}]\label{fact:chisquare}
Let $\{Z_i\}_{i \in [n]} \sim_{\textup{i.i.d.}} \Nor(0, 1)$ and $\va \in \R^n_{\ge 0}$. Then,
\begin{align*}
\Pr\Brack{\sum_{i \in [n]} \va_i Z_i^2 - \norm{\va}_2^2 \ge 2\norm{\va}_2 \sqrt t + 2\norm{\va}_\infty t} \le \exp(-t),\\ 
\Pr\Brack{\sum_{i \in [n]} \va_i Z_i^2 - \norm{\va}_2^2 \le -2\norm{\va}_2\sqrt t} \le \exp(-t).
\end{align*}
\end{fact}

As a consequence of Facts~\ref{fact:mills} and~\ref{fact:chisquare} we have the following.

\restatemaxrowip*
\begin{proof}
We will show that for all $j \in [d]$, $|\mx_{:j}^\top \vxi| \le $ with probability $\ge 1 - \frac \delta d$. Since $\mx$ satisfies $(\epsilon, 1)$-RIP, for any $j \in [d]$, we have $\twonorm{\mx\ve_j}^2 = \twonorm{\mx_{:j}} \leq 1 + \epsilon$. Moreover, $\mx_{:j}^\top \vxi \sim \Nor(0, \sigma^2\norm{\mx_{:j}}_2^2)$, so
\[\Pr\Brack{|\mx_{:j}^\top \vxi| > \sigma (1 + \epsilon)\sqrt{2\log\Par{\frac d \delta}}} \le \frac \delta d,\]
by Fact~\ref{fact:gauss_norm}. The conclusion follows by taking a union bound over all columns $j \in [d]$.
\end{proof}

\paragraph{Isometric conditions.} We require another notion of isometry from the sparse recovery literature, used in proving Proposition~\ref{prop:sparse_recovery}. 
\begin{definition}[Mutual incoherence]\label{def:MI}
We say $\mx \in \bbR^{n\times d}$ satisfies the mutual incoherence (MI) condition over $S \subseteq [d]$ with parameter $\alpha \in [0, 1)$, or $\mx$ is $\alpha$-MI over $S$, if 
\begin{equation*}
    \max_{j \in S^c} \onenorm{[\mx^\top\mx]_{S\times S}^{-1} \mx_{S:}^\top \mx_{:j}} \leq \alpha.
\end{equation*}
\end{definition}

We next show how Definitions~\ref{def:RIP} and~\ref{def:MI} relate.

\begin{lemma}[RIP implies MI]
\label{lem: RIP2MI}
    For any $\mx \in \bbR^{n\times d}$, if $\mx$ is $(\epsilon, s+1)$-RIP, then $\mx$ is $\alpha$-$\MI$ over $S$ for any $S \subseteq [d]$ with $|S| \leq s$, and $\alpha = \sqrt{2s\eps/(1 - \eps)}$.
\end{lemma}
\begin{proof}
    Fix $S \subseteq [d]$ such that $|S| \leq k$, and fix $j \in S^c$. We define a projection operator: 
    \begin{equation*} 
        \matp_S := \mx_{:S} [\mx^\top \mx]^{-1}_{S\times S} \mx_{S:}^{\top}.    
    \end{equation*}
    It is well known that $\matp_S$ is the projection matrix of the linear subspace: $\mathrm{span}\{\mx_{:j} \in \bbR^n \mid j\in S\}$.
    Let $T \defeq S \cup {j}$. Since $\mx$ satisfies $(\epsilon, s+1)$-RIP, for any $\vth \in \R^d$, we have
    \begin{equation}\label{eq:t_rip}
        \twonorm{\mx_{:T} \vth_T}^2 = \twonorm{\mx_{:S}\vth_S + \mx_{:j}\vth_j}^2 \geq (1 - \epsilon)\twonorm{\vth_T}^2 = (1 - \epsilon)(\twonorm{\vth_S}^2 + \vth_j^2).
    \end{equation}
    By the properties of projection operators, we have 
    \begin{align}
        \label{eq:proj_norm}
        \twonorm{\mx_{:j}}^2 &= \twonorm{\matp_S \mx_{:j}}^2 + \twonorm{(\id - \matp_S)\mx_{:j}}^2,\\
        \label{eq:proj_min}
        \twonorm{(\id - \matp_S) \mx_{:j}}^2 &= \min_{\vth_S \in \bbR^S}\twonorm{\mx_{:S} \vth_S + \mx_{:j}}^2 \geq 1 - \epsilon, 
    \end{align}
    where the inequality in \eqref{eq:proj_min} follows from \eqref{eq:t_rip} with $\vth_j = 1$ and $\norm{\vth_S}_2 \ge 0$. 
    Inserting \eqref{eq:proj_min} into \eqref{eq:proj_norm}, 
    \begin{equation*}
        \twonorm{\matp_S \mx_{:j}}^2 \leq \twonorm{\mx_{:j}}^2 - (1 - \epsilon).
    \end{equation*}
    Next, let $\vx_j \defeq [\mx^\top\mx]_{S\times S}^{-1} \mx_{S:}^{\top}\mx_{:j} \in \R^S$. Then applying RIP again, we have 
    \begin{equation*}
        \twonorm{\matp_S \mx_{:j}}^2 = \twonorm{\mx_{:S} \vx_j}^2 \geq (1 -\epsilon) \twonorm{\vx_j}^2.
    \end{equation*}
    Thus, we have by combining the above two displays that
    \begin{equation*}
        \twonorm{\vx_j}^2 \leq \frac{\twonorm{\mx_{:j}}^2}{1 - \epsilon} - 1.
    \end{equation*}
    Moreover, notice that $\vx_j \in \bbR^{S}$, so the above display implies
    \[\norm{\vx_j}_1 \le \sqrt{s}\norm{\vx_j}_2 \le \sqrt{s} \cdot \sqrt{\frac{\norm{\mx_{:j}}_2^2}{1 - \eps} - 1}.\]
    Finally, by RIP applied with the $1$-sparse test vector $\ve_j$,
    \[\norm{\mx \ve_j}_2^2 = \norm{\mx_{:j}}_2^2 \in [1 - \eps, 1 + \eps] \implies \norm{\vx_j}_1 \le \sqrt{\frac{2s\eps}{1-\eps}}.\]
    Because our choices of $S$ and $j \in S^c$ were arbitrary, we have the desired MI parameter bound.
\end{proof}
\subsection{Solving the Lasso to high precision}\label{ssec:lasso_solve}

Here, we consider the (Lagrangian) Lasso problem, a.k.a.\ $\ell_1$-penalized linear regression, defined as: 
\ba{
    \hat{\vth} := \argmin_{\vth \in \R^{d}}\frac{1}{2}\norm{\mx \vth - \vy}_{2}^{2} + \lambda\norm{\vth}_{1},\label{eq:lagrangian_lasso}
}
where $\mx \in \R^{n \times d}$, $\vth \in \R^{d}$, $\vy \in \R^{n}$ and $\lambda > 0$.
We provide a polynomial-time algorithm achieving $\ell_{\infty}$-norm convergence guarantees to the optimal solution. The reason this is nontrivial using off-the-shelf convex optimization tools, e.g., cutting-plane methods, is because \eqref{eq:lagrangian_lasso} is not strongly convex, so function approximation guarantees do not transfer to distance bounds.

Throughout the section, we make the following assumption.

\begin{assumption}\label{assume:unique_opt}
The columns of $\mx$ are in general position.
\end{assumption}

We now restate a result from \cite{tibshirani2013lasso} that shows Assumption~\ref{assume:unique_opt} implies uniqueness of the Lasso solution \eqref{eq:lagrangian_lasso}, and characterizes the solution. The former fact is attributed to \cite{OsbornePT00}.

\begin{lemma}[Lemma 4, \cite{tibshirani2013lasso}]\label{lemma:lasso_unique_solution} Under Assumption~\ref{assume:unique_opt}, $\eqref{eq:lagrangian_lasso}$ has a unique solution, $\hat{\vth}$, satisfying
\bas{
    \hat{\vth}_{S} = [\mx^{\top}\mx]_{S \times S}^{-1}(\mx_{S:}^{\top}\vy - \lambda \vs), \;\;\; \hat{\vth}_{S^{C}} = \mathbf{0}, \;\;\; \vs := \sign(\mx_{S:}^{\top}(\vy - \mx\hat{\vth})),
}
where $S \defeq \{i \in [d] \mid |\mx_{i:}^\top (\vy - \mx \vhth)| = \lam \}$.
\end{lemma}

We next derive a constrained dual problem for \eqref{eq:lagrangian_lasso}.

\begin{lemma}[Lasso dual]\label{lemma:lagrangian_lasso_dual}
    The dual problem for \eqref{eq:lagrangian_lasso} is given as
    \ba{
        \hat{\vz} =  \argmin_{\|\vz\|_\infty \,\le\,1}\;\bb{\vz - \va}^{\top}(\mx^{\top}\mx)^{\dagger}\bb{\vz - \va}, \;\; \va := \frac{1}{\lambda}\mx^{\top}\vy. \label{eq:lagrange_lasso_dual}
    }
    Furthermore, if Assumption~\ref{assume:unique_opt} holds, then \eqref{eq:lagrange_lasso_dual} has a unique minimizer satisfying
    \bas{
        \hat{\vth}_{S} = [\mx^{\top}\mx]_{S \times S}^{-1}(\mx_{S:}^{\top}\vy - \lambda  \hat{\vz}_{S}), \;\;\; \hat{\vth}_{S^c} = \mathbf{0}, \;\;\;  \hat{\vz} = \va - \frac{1}{\lambda}\mx^{\top}\mx\hat{\vth}
    }
    where $\hat{\vth}$ is the unique minimizer of \eqref{eq:lagrangian_lasso} and $S = \left\{i \in [d] \mid |\hat{\vz}_{i}| = 1\right\}$.
\end{lemma}
\begin{proof}
First, by the standard identity $\|\vth\|_1=
\max_{\|\vz\|_\infty \,\le\,1}
\vz^\top \vth$, \eqref{eq:lagrangian_lasso} is equivalently
\ba{
    \min_{\vth \in \R^{d}}\max_{\|\vz\|_\infty \,\le\,1}\frac{1}{2}\norm{\mx \vth - \vy}_{2}^{2} + \lambda\vz^\top \vth. \label{eq:lasso_primal_formulation}
}
Note that the set $\|\vz\|_\infty \,\le\,1$ is a compact set, and $f(\vth, z) := \frac{1}{2}\norm{\mx \vth - \vy}_{2}^{2} + \lambda\vz^\top \vth$ is convex-concave. Therefore, using Sion's minimax theorem, \cite{sion1958general}, strong duality holds so \eqref{eq:lasso_primal_formulation} equals
\ba{
    \max_{\|\vz\|_\infty \,\le\,1}\min_{\vth \in \R^{d}}\frac{1}{2}\norm{\mx \vth - \vy}_{2}^{2} + \lambda\vz^\top \vth .\label{eq:lasso_dual_formulation}
}
Solving the inner minimization shows that for fixed $\vz$, we should choose 
\bas{
    \vth = (\mx^{\top}\mx)^{\dagger}(\mx^{\top}\vy - \lambda \vz).
}
Substituting this back into \eqref{eq:lasso_dual_formulation} we have for the optimal solution, $\hat{\vz}$, of the dual \eqref{eq:lasso_dual_formulation}, that
\bas{
    \hat{\vz} &= \argmax_{\|\vz\|_\infty \,\le\,1}\;-\frac{\lambda^{2}}{2}\bb{\vz - \va}^{\top}(\mx^{\top}\mx)^{\dagger}\bb{\vz - \va} + \frac{\lambda^{2}}{2}\va^{\top}(\mx^{\top}\mx)^{\dagger}\va + \frac{1}{2}\norm{\bb{\id - \mx(\mx^{\top}\mx)^{\dagger}\mx^{\top}}\vy}_{2}^{2} \\
    &= \argmax_{\|\vz\|_\infty \,\le\,1}\;-\frac{\lambda^{2}}{2}\bb{\vz - \va}^{\top}(\mx^{\top}\mx)^{\dagger}\bb{\vz - \va} = \argmin_{\|\vz\|_\infty \,\le\,1} \;\bb{\vz - \va}^{\top}(\mx^{\top}\mx)^{\dagger}\bb{\vz - \va}.
}
This proves the first conclusion. For the second conclusion, since strong duality holds, by using KKT conditions, we have for primal and dual optimal solutions $\hat{\vth}, \hat{\vz}$ respectively,
\bas{
    \mx^{\top}\mx\hat{\vth} = \mx^{\top}\vy - \lambda \hat{\vz} , \;\; \norm{\hat{\vz}}_{\infty} \leq 1.
}
Note that $\hat{\vz}$ is fixed given $\hat{\vth}$. Under Assumption~\ref{assume:unique_opt}, we conclude uniqueness of $\hat{\vz}$. Next, we have 
\bas{
   \hat{\vz}_{i} = \frac{1}{\lambda}\mx_{i}^{\top}( \vy - \mx\hat{\vth}) \implies \max_{i \in [d]}|\mx_{i}^{\top}( \vy - \mx\hat{\vth})| \leq \lambda.
}
Let $S \defeq \left\{i \in [d] : |\hat{\vz}_{i}| = 1\right\}$, which by the above is consistent with the definition in Lemma~\ref{lemma:lasso_unique_solution}. We have the remaining conclusions due to Lemma~\ref{lemma:lasso_unique_solution}.
\end{proof}

Our strategy is now to show that restricted to $ \range(\mx^{\top})$, the dual problem \eqref{eq:lagrange_lasso_dual} is strongly convex, and consequently, we can suboptimality guarantees to distance bounds on the parameter $\vz$.

\begin{lemma}\label{lemma:lasso_dual_strong_convexity}Let $\mx$ satisfy Assumption~\ref{assume:unique_opt} with $n \le d$, with singular value decomposition
\bas{
    \mx = \muu\msig\mv^{\top}, \muu \in \R^{n \times n}, \msig = \diag{\vsig} \text{ for } \vsig \in \R^n_{> 0},\; \mv \in \R^{d \times n}.
}
Consider the following optimization problem:
\ba{
    \hat{\vw} =  \argmin_{\norm{ \mv\vw}_{\infty} \leq 1}\;(\vw-\vb)^{\top}\msig^{-2}(\vw-\vb), \;\; \mv\vb := \frac{1}{\lambda}\mx^{\top}\vy.  \label{eq:strongly_convex_dual}
}
Then, $\hat{\vz} = \mv\hat{\vw} $, where $\hat{\vz}$ is the unique optimal dual solution as defined in Lemma~\ref{lemma:lagrangian_lasso_dual}.
\end{lemma}
\begin{proof}
    By Lemma~\ref{lemma:lagrangian_lasso_dual}, the optimal solution to \eqref{eq:lagrange_lasso_dual} is of the form 
    \bas{
        \hat{\vz} = \va - \frac{1}{\lambda}\mx^{\top}\mx\hat{\vth} = \frac{1}{\lambda}\mx^{\top}(\vy - \mx\hat{\vth})
    }
    where $\va :=  \frac{1}{\lambda}\mx^{\top}\vy$.
    Therefore, $\hat{\vz} \in \range\bb{\mx^{\top}}$ and $\exists \vw \in \R^n$ such that $\hat{\vz} = \mv\hat{\vw}$. Similarly, $\va \in \range\bb{\mx^{\top}}$. Therefore, $\exists \vb \in \R^{n}$ such that $\mv\vb = \frac{1}{\lambda}\mx^{\top}\vy$.
    
    Thus, if we were to restrict the dual problem to solutions of the form 
    \ba{
        \vz = \mv\vw  \label{eq:z_to_w_substitution}
    }
    then we would restrict our constraint set to a subset of the original set in \eqref{eq:lagrange_lasso_dual},  $\left\{\vz \mid \norm{\vz}_{\infty} \leq 1\right\}$. However, since  $\hat{\vz} = \mv\hat{\vw}$ holds, this restriction does not affect the problem. The uniqueness of $\hat{\vw}$ implies the uniqueness of $\hat{\vz}$. The conclusion of the lemma follows by using \eqref{eq:z_to_w_substitution}.  
\end{proof}

Lemma~\ref{lemma:lasso_dual_strong_convexity} shows that \eqref{eq:strongly_convex_dual}, our reformulation of \eqref{eq:lagrange_lasso_dual}, is strongly convex, with a convex constraint set. Consequently, a standard convex optimization algorithm can now be used to achieve function approximation guarantees, which we formalize, using a result from \cite{NesterovN94}, restated below.

\begin{lemma}[Eq.\ (8.1.5), \cite{NesterovN94}]\label{lemma:qplc} Consider the following optimization problem: 
\bas{
    \min \psi(\vx) := \frac{1}{2}\vx^{\top}\ma \vx - \vc^{\top}\vx,\;
    \text{subject to } \vx \in \R^{n}, -\vc_{i}^{\top}\vx + \vr_i \geq 0 \text{ for all } i \in [d],
}
where $\ma \in \PSD^{n \times n}$ and $\left\{\vc_{i}\right\}_{i \in [d]} \cup \{\vc\} \subset \R^{n}$, and $\vr \in \R^d$. Define the sets 
\[\calG := \left\{\vx \mid  \vc_{i}^{\top}\vx + \vr_i \geq 0 \text{ for all } i \in [d]\right\},\; \calG' \defeq \left\{\vx \mid  \vc_{i}^{\top}\vx + \vr_i > 0 \text{ for all } i \in [d]\right\}.\] 
If $\calG'$ is non-empty, then for any $\epsilon > 0$, there is an algorithm which returns $\vx_{\eps} \in \calG'$ satisfying 
\bas{
    \psi(\vx_{\alpha}) - \min_{\vx \in \calG}\psi(\vx) \leq \epsilon\Par{\max_{\vx \in \calG}\psi(\vx) - \min_{\vx \in \calG}\psi(\vx)},
}
initialized at $\vx_0 \in \calG$, and runs in time 
\bas{
    O\bb{d^{1.5}n^{2}\log\bb{\frac{d}{\alpha\bb{G, \vx_{0}}\eps}}}
}
where $\alpha\bb{\calG, \vx_{0}} := \max\left\{\alpha \mid \vx_{0} + \alpha(\vx_{0} - \calG) \subset \calG\right\}$.
\end{lemma}

We next bound the asymmetry coefficient $\alpha(\calG, \vx_0)$ for our quadratic program \eqref{eq:strongly_convex_dual}.

\begin{lemma}\label{lemma:asymmetry_coefficient_bound}
Let $\calG := \left\{\vw \in \R^n \mid \norm{\mv\vw}_{\infty} \leq 1\right\}$ where $\mv \in \R^{d \times n}$ satisfies $\mv^{\top}\mv = \id_{n}$. Then 
    \bas{
        \alpha(\calG, \vzero_n) \defeq \max\Brace{\alpha \mid \vzero_n + \alpha(\vzero_n - \calG) \subset \calG} \ge \half.
    }
\end{lemma}
\begin{proof}
Since $\calG$ is symmetric about the origin and nonempty, $-\half \calG = \half \calG \subset \calG$ as claimed.
\end{proof}

Finally, we combine these results to provide a fast algorithm for approximating \eqref{eq:lagrangian_lasso}.

\begin{proposition}\label{prop:recover_lasso}
Let Assumption~\ref{assume:unique_opt} hold. Following the notation in \eqref{eq:lagrangian_lasso}, there is an algorithm which returns $\vth$ satisfying $\norms{\vth - \vhth}_2 \le \eps$,
in time
\[O\Par{d^{1.5} n^2\log\Par{\frac{(\lam + \norm{\mx^\top \vy}_2) d}{\eps} \cdot \frac{\vsig_1(\mx)}{\vsig_n(\mx)} }}.\]
\end{proposition}
\begin{proof}
We first claim it suffices to solve \eqref{eq:strongly_convex_dual} to function error
\[\Delta \defeq \frac{\eps^2}{\lam^2} \cdot \frac{\vsig_n(\mx)^4}{\vsig_1(\mx)^2}.\]
To see this, let $\vw$ satisfy $f(\vw) - f(\hvw) \le \Delta$, where $\hvw$ minimizes \eqref{eq:strongly_convex_dual}. By strong convexity of \eqref{eq:strongly_convex_dual},
\[\norm{\vw - \hvw}_2^2 \le \vsig_1(\msig)^2 \Delta = \vsig_1(\mm)^2 \Delta. \]
Moreover, letting $\vz \defeq \mv \vw$ and $\hvz \defeq \mv \hvw$, so that $\hvz$ minimizes \eqref{eq:lagrange_lasso_dual} by Lemma~\ref{lemma:lasso_dual_strong_convexity}, we have $\norm{\vw - \hvw}_2 = \norm{\vz - \hvz}_2$ since $\mv^\top \mv = \id_n$. Finally, by Lemma~\ref{lemma:lagrangian_lasso_dual}, letting $\vhth \defeq (\mx^\top \mx)^\dagger (\mx^\top \vy - \lam \hvz)$ optimize \eqref{eq:lagrangian_lasso} and $\vth \defeq (\mx^\top \mx)^\dagger (\mx^\top \vy - \lam \vz)$, we have
\[\norm{\vth - \vhth}_2 = \lam\norm{(\mx^\top \mx)^\dagger (\vz - \hvz)}_2 \le \frac{\lam}{\vsig_n(\mx)^2} \norm{\vz - \hvz}_2 = \frac{\lam}{\vsig_n(\mx)^2}\norm{\vw - \hvw}_2.\]
It remains to bound the complexity of achieving $\Delta$ function error. Letting $\ma \defeq 2\msig^{-2}$ and $\vc \defeq 2\msig^{-2} \vb$ in the setting of Lemma~\ref{lemma:qplc},
\begin{align*}
\max_{\vw \in \calG} \psi(\vw) - \min_{\vw \in \calG} \psi(\vw) &\le \max_{\vw \in \calG} (\vw - \vb)^\top \msig^{-2} (\vw - \vb) \\
&\le \frac 1 {\vsig_n(\mx)^2} \max_{\norm{\mv \vw}_\infty \le 1} \norm{\vw - \vb}_2^2 \\
&\le \frac 1 {\vsig_n(\mx)^2} \cdot \Par{\max_{\norm{\mv \vw}_2 \le \sqrt{d}} 2\norm{\vw}_2^2 + 2\norm{\vb}_2^2} \le \frac {2(d + \norm{\vb}_2^2)} {\vsig_n(\mx)^2}.
\end{align*}
The conclusion follows from Lemma~\ref{lemma:qplc} using $\norm{\vb}_2 \le \frac 1 \lam \norm{\mx^\top \vy}_2$ by our choice of $\vb = \frac 1 \lam \mv^\top \mx^\top \vy$.
\end{proof}

\subsection{Proof of Proposition~\ref{prop:sparse_recovery}: \texorpdfstring{$\ell_\infty$}\ \ recovery}\label{app:L_infty_estimator}

In this section, we prove the first statement in Proposition~\ref{prop:sparse_recovery}. Our main tool is the following known result on the performance of the Lasso for sparse recovery under MI (Definition~\ref{def:MI}).

\begin{lemma}[Theorem 7.21, \cite{wainwright2019high}]
\label{lem:L_infty_bound_RIP}
   In the setting of Model~\ref{model:sparse_recovery}, let $\supp(\vths) = S$ and $|S| = k$. If $\vlam_{\min}([\mxtx]_{S\times S}) > 0$ and $\mx$ is $\alpha$-MI over $S$, then for any regularization parameter $\lam$ such that 
   \begin{equation}\label{eq:lam_lower}
        \lambda \geq \frac{2}{1 - \alpha} \infnorm{\mx_{S^c:}^\top \mproj_{S^{\perp}}(\mx) \vxi},
    \end{equation}
    where $\mproj_{S^{\perp}}(\mx) \defeq \id_n - \mx_{:S}[\mxtx]_{S \times S}^{-1}\mx^\top_{S:}$,
    the Lasso solution 
    \begin{equation}
        \label{eq:Lag_Lasso}
        \vhth \defeq \argmin_{\vth \in \bbR^d}\cbra{\twonorm{\mx\vth - \vy}^2 + \lambda \onenorm{\vth}}
    \end{equation}
    satisfies \begin{equation}\label{eq:linf_lasso}
            \infnorm{\vhth - \vth^\star} \leq \infnorm{\Brack{\mx^\top \mx}^{-1}_{S\times S}\mx_{S:}^\top\vxi} + \lambda\max_{i \in S}\sum_{j\in S}\Abs{\Brack{\mx^\top \mx}^{-1}_{ij}}.
        \end{equation}
\end{lemma}

We next give a bound on the parameters in \eqref{eq:lam_lower}, \eqref{eq:linf_lasso} under RIP.

\begin{lemma}
\label{lem:L_infty_bound}
   In the setting of Model~\ref{model:sparse_recovery}, let $\supp(\vths) = S$ and $|S| = k$. If  $\mx$ satisfies $(\epsilon, k+1)$-RIP, then for any regularization parameter $\lam$ such that 
    \begin{equation}\label{eq:lam_lower_rip}
        \lambda \geq \frac{2(1 + \alpha)}{1 - \alpha} \infnorm{\mx^\top \vxi},
    \end{equation}
    the Lasso solution \eqref{eq:Lag_Lasso} satisfies
    \begin{equation}\label{eq:linf_lasso_rip}
        \infnorm{\hat{\vth} - \vth^\star} \leq \paren{\frac{1}{1 - \epsilon} + \frac{2\sqrt{k\epsilon}}{1 - \epsilon^2}}\Par{\infnorm{\mx^\top \vxi} + \lam}.
    \end{equation}
\end{lemma}
\begin{proof}
    Since $\mx$ is $(\epsilon, k+1)$ RIP, we have by Lemma~\ref{lem: RIP2MI} that $\mx$ is $\alpha$-MI over any $S\subseteq [d]$ with $|S| \le k$, for  $\alpha = \sqrt{2k\eps/(1-\eps)}$. We begin by bounding the right-hand side of \eqref{eq:lam_lower}. By the definition of $\mproj_{S^\perp}(\mx)$ and the triangle inequality, we have that
    \begin{equation}
        \label{eq:proj_decomp}
        \infnorm{\mx^\top_{S^c:}\mproj_{S^\perp}(\mx)\vxi} \leq \infnorm{\mx^\top_{S^c:}\vxi} + \infnorm{\paren{\Brack{\mx^\top\mx}_{S\times S}^{-1}\mx_{S:}^\top \mx_{: S^c}}^\top \mx_{S:}^\top \vxi}.
    \end{equation}
    Also by the definition of $\infnorm{\cdot}$, we have 
    \begin{equation}\label{eq:mi_apply}\begin{aligned}
        \infnorm{\paren{\Brack{\mx^\top\mx}_{S\times S}^{-1}\mx_{S:}^\top \mx_{: S^c}}^\top \mx_{S:}^\top \vxi} \leq \max_{j\in S^c}\onenorm{\Brack{\mx^\top\mx}_{S\times S}^{-1}\mx_{S:}^\top \mx_{:j}}\cdot \infnorm{\mx_{S:}^\top \vxi}    \le \alpha \norm{\mx_{S:}^\top \vxi}_\infty,
        \end{aligned}
    \end{equation}
    where the last inequality used MI. Combining \eqref{eq:proj_decomp} and \eqref{eq:mi_apply}, we have shown 
    \begin{equation*}
        \infnorm{\mx^\top_{S^c\times S^c}\mproj_{S^\perp}(\mx) \vxi} \leq (1 + \alpha)\infnorm{\mx^\top\vxi},
    \end{equation*}
    proving that \eqref{eq:lam_lower_rip} is a sufficient condition for \eqref{eq:lam_lower} to apply. Next we bound the right-hand side of \eqref{eq:linf_lasso}.
    Let $\mm \defeq [\mx^\top\mx]_{S\times S}^{-1}$. Because $\norms{\mm}_{\infty \to \infty} = \max_{i \in S} \norms{\mm_{i:}}_1$, we have from \eqref{eq:linf_lasso} that
    \begin{equation}\label{eq:bound_inf_op}\norm{\vhth - \vths}_\infty \le \norm{\mm}_{\infty \to \infty} \Par{\norm{\mx^\top \vxi}_\infty + \lam},\end{equation}
    so it is enough to provide a bound on $\norm{\mm}_{\infty \to \infty}$. Because $\mx$ satisfies $(\epsilon, k+1)$-RIP, we have 
    \begin{equation}
        \label{eq:M_inv_eigen}
        \vlam(\mm^{-1}) \in [1-\epsilon, 1+\epsilon]^k \implies \vlam(\mm) \in \bra{\frac{1}{1+\epsilon}, \frac{1}{1 - \epsilon}}^k.
    \end{equation}
    Thus, because $\norm{\ve_i}_2 = 1$,
    \begin{equation}\label{eq:M_diag}
        \mm_{ii} = \ve^\top_i \mm \ve_i \in \Brack{\vlam_{\min}(\mm), \vlam_{\max}(\mm)}  = \Brack{\frac 1 {1 + \eps}, \frac 1 {1 - \eps}}.
    \end{equation}
    Moreover, we have that
    \begin{align*}
    \sum_{\substack{j \in S \\ j \neq i}} \mm_{ij}^2 = \norm{\mm \ve_i}_2^2 - \mm_{ii}^2 \le \frac 1 {(1 - \eps)^2} - \frac 1 {(1 + \eps)^2} = \frac{4\eps}{(1 - \eps^2)^2},
    \end{align*}
    where the inequality used the upper bound in \eqref{eq:M_inv_eigen} and the lower bound in \eqref{eq:M_diag}. Finally, since $\vv \in \R^S$  has $\norm{\vv}_1 \le \sqrt{k}\norm{\vv}_2$, by combining the above display with the upper bound in \eqref{eq:M_diag},
    \begin{align*}
    \norm{\mm_{i:}}_1 = \mm_{ii} + \sum_{\substack{j \in S \\ j \neq i}} \Abs{\mm_{ij}} \le \frac 1 {1 - \eps} + \frac{2\sqrt{k\eps}}{1 - \eps^2}.
    \end{align*}
    
    Finally, plugging the above into \eqref{eq:bound_inf_op} yields the desired claim \eqref{eq:linf_lasso_rip}.
\end{proof}

We are now ready to prove the first part of Proposition~\ref{prop:sparse_recovery}.

\begin{proof}[Proof of Proposition~\ref{prop:sparse_recovery}: $\ell_\infty$ recovery]
Under Assumption~\ref{assume:linf}, we have by Lemma~\ref{lem: RIP2MI} that for
\[\alpha = \sqrt{\frac{2k\eps}{1-\eps}} \le \sqrt{\frac 2 3},\; \lam = \frac{2(1 + \alpha)}{1 - \alpha} \norm{\mx^\top \vxi}_\infty = \Theta\Par{\norm{\mx^\top \vxi}_\infty},\]
that $\mx$ is $\alpha$-MI, and hence Lemma~\ref{lem:L_infty_bound} shows that the solution to \eqref{eq:linf_lasso} satisfies
\[\norm{\vhth - \vths}_\infty = O\Par{\norm{\mx^\top \vxi}_\infty}.\]
Now it suffices to return an estimate of $\vhth$ within $\ell_\infty$ distance $O(\norm{\mx^\top \vxi}_\infty)$. To do so, we use Proposition~\ref{prop:recover_lasso}, giving the conclusion.
\end{proof}

\subsection{Proof of Proposition~\ref{prop:sparse_recovery}: \texorpdfstring{$\ell_2$}\ \ recovery}\label{app:L_2_estimator}

In this section, we prove the second statement in Proposition~\ref{prop:sparse_recovery}. Our proof is a simple modification of a fairly standard approach based on projected gradient descent over $\ell_1$ balls. Our result is implicit in \cite{KelnerLLST23}, but we give a brief explanation of how to derive it.

\begin{proof}[Proof of Proposition~\ref{prop:sparse_recovery}: $\ell_2$ recovery]
This result is implicit in the proof of Theorem 4, \cite{KelnerLLST23}, which handles a much more general setting of a \emph{semi-random} RIP observation matrix. The exposition in \cite{KelnerLLST23} loses several additional logarithmic factors in the runtime due to the need to construct a ``noisy step oracle,'' but in our simpler standard RIP setting, we can use uniform weights over the rows of $\mx_{[m]:}$ to skip this step. There is a presentation of this simpler setting in \cite{Tian24}, Theorem 3, in the noiseless case, and it extends to the noisy case in the same way as done in \cite{KelnerLLST23}. In particular, in the noisy case it suffices to apply Theorem 3 in \cite{Tian24} with $R \gets R_2$ and $r \gets r_2$, i.e., the noise level at which point we can no longer guarantee progress.
\end{proof}
\section{Additional deferred proofs from Section~\ref{sec:prelims}}
\label{append:preliminary}

\subsection{Deferred proofs from Section~\ref{ssec:posterior_estimate}}\label{app:posterior_estimate}

\restateesttosamp*
\begin{proof}
Let $\event_{\vth, \vhth}$ denote the event that $\met(\vth, \vhth) \le \eps$, and similarly define $\event_{\vths, \vhth}$. We have 
\begin{align*}
\Pr_{\vths, \vhi, \vth}\Brack{\met(\vth, \vths) \le 2\eps} &\ge \Pr_{\vths, \vhi, \vhth, \vth}\Brack{\met(\vth, \vhth) \le \eps \wedge \met(\vths, \vhth) \le \eps} \\
&= \Pr_{\vths, \vhi, \vhth, \vth}\Brack{\event_{\vth, \vhth} \wedge \event_{\vths, \vhth}} = \E_{\vths, \vhi, \vhth, \vth}\Brack{\ind_{\event_{\vth, \vhth}} \cdot \ind_{\event_{\vths, \vhth}}} \\
&= \E_{\vhi, \vhth}\Brack{\E_{\vths, \vth}\Brack{\ind_{\event_{\vth, \vhth}} \cdot \ind_{\event_{\vths, \vhth}} \mid \vhi, \vhth}} \\
&= \E_{\vhi, \vhth}\Brack{\E_{\vth}\Brack{\ind_{\event_{\vth, \vhth}}  \mid \vhi, \vhth} \cdot \E_{\vths}\Brack{\ind_{\event_{\vths, \vhth}} \mid \vhi, \vhth}} \\
&= \E_{\vhi, \vhth}\Brack{\Pr_{\vth}\Brack{\event_{\vth, \vhth}  \mid \vhi, \vhth} \cdot \Pr_{\vths}\Brack{\event_{\vths, \vhth} \mid \vhi, \vhth}} \\
&= \E_{\vhi, \vhth}\Brack{\Pr_{\vths}\Brack{\event_{\vths, \vhth} \mid \vhi, \vhth}^2}.
\end{align*}
The first line used that $\met$ is a metric and applied the triangle inequality, the second used the definitions of $\event_{\vth, \vhth}, \event_{\vths, \vhth}$, the third iterated expectations, the fourth used that $\vth$ and $\vths$ are independent conditioned on $\vhi, \vhth$, and the last used that $(\vths, \vhi, \vhth)$ and $(\vth, \vhi, \vhth)$ have the same joint distribution. Finally, the conclusion follows by Jensen's inequality and \eqref{eq:assume_estimator}:
\begin{align*}
\E_{\vhi, \vhth}\Brack{\Pr_{\vths}\Brack{\event_{\vths, \vhth} \mid \vhi, \vhth}^2} &\ge \E_{\vhi, \vhth}\Brack{\Pr_{\vths}\Brack{\event_{\vths, \vhth} \mid \vhi, \vhth}}^2 \\
&= \Pr_{\vths, \vhi, \vhth}\Brack{\event_{\vths, \vhth}} = (1 - \delta)^2 \ge 1 - 2\delta.
\end{align*}
\end{proof}

\restategoodmodels*

\begin{proof}
Let $p \defeq \Pr_\alpha[g(\alpha) \le 1 - \delta_1]$. By expanding, and using $g(\alpha) \in [0, 1]$ for all $\alpha$,
\begin{align*} 1 - \delta_1 \delta_2 &\le \Pr_{(\alpha, \beta)}\Brack{\event_{\alpha, \beta}} = \E_{\alpha}\Brack{g(\alpha)} \\
&= p\E_\alpha\Brack{g(\alpha) \mid g(\alpha) \le 1 - \delta_1} + (1 - p)\E_{\alpha}\Brack{g(\alpha) \mid g(\alpha) > 1 - \delta_1}  \\
&\le p(1 - \delta_1) + 1 - p = 1 - p\delta_1.
\end{align*}
Rearranging shows that $p \le \delta_2$ as claimed.
\end{proof}

\subsection{Deferred proofs from Section~\ref{ssec:rejection_sampling}}\label{app:reject}

We first restate (a special case of) Lemma 12 from \cite{LeeST21}, which we use to prove Lemma~\ref{lem:reject}.

\begin{lemma}[Lemma 12, \cite{LeeST21}]
\label{lemma:reject_lee_stl21}
Let $\pi$, $\mu$ be distributions over the same domain, and suppose that $\pi \propto P$ and $\mu \propto Q$ for unnormalized densities $P, Q$. Moreover, suppose for $\eps \in (0, 1)$ and $C \ge 1$, there is a set $\Omega$ such that $\Pr_{\omega \sim \pi}[\omega \in \Omega] \ge 1 - \eps$, and 
\begin{align*}
\frac{P(\omega)}{Q(\omega)} \le C \text{ for all } \omega \in \Omega,\; \frac{\int Q(\omega) \dd \omega}{\int P(\omega) \dd \omega} \le 1.
\end{align*}
There is an algorithm that, for any $\delta \in (0, 1)$ outputs a sample within total variation distance $\eps + \delta$ from $\pi$. The algorithm uses $O(\frac {C}{1 - \eps}\log(\frac 1 \delta))$ samples from $\mu$, and evaluates $\frac{P(\omega)}{Q(\omega)}$ $O(\frac {C}{1 - \eps}\log(\frac 1 \delta))$ times.
\end{lemma}

Now we present the proof of Lemma~\ref{lem:reject}.

\rejectionsampling*
\begin{proof}
We apply Lemma~\ref{lemma:reject_lee_stl21} with $\eps = 0$, i.e., with $\Omega$ as the whole domain. 
For all $\omega$, let $Q'(\omega) := \frac{Q(\omega)}{C}$. Then, we have the result by applying Lemma~\ref{lemma:reject_lee_stl21}  with $P \leftarrow P$ and $Q \leftarrow Q'$, since
\bas{
    \frac{P(\omega)}{Q'(\omega)} = \frac{CP(\omega)}{Q(\omega)} \leq C^2,\; \int Q'(\omega) \dd \omega = \frac 1 C \int Q(\omega) \dd \omega \le \int P(\omega) \dd \omega.
}
\end{proof}

\subsection{Deferred proofs from Section~\ref{ssec:conditional_poisson_sampling}}\label{app:cps_helper}

\begin{algorithm}[H]
\caption{$\cps(\vp, k)$}
\label{alg:cond_sampling}
\begin{algorithmic}[1]
\State \textbf{Input:} $\vp \in [0, 1]^d$ inducing $\pi \defeq \bigotimes_{i \in [d]} \Bern(\vp_i)$, maximum allowed successes $k \in [d]$
\State \textbf{Output:} $S \in \Omega_k \defeq \{S \subseteq [d] \mid |S| \le k\}$ distributed as $\pi(S \mid S \in \Omega_k)$
\For{\(j \in [0,k]\)}\Comment{\textit{Precompute dynamic programming table.}}\label{line:dp_start}
    \State \(F(d+1,j) \gets \ind_{j = 0}\)
\EndFor
\For{\(i = d\) \textbf{to} \(1\)}
    \For{\(j \in [0,k]\)}
        \If{\(j = 0\)}
            \State \(F(i,j) \gets 1\)
        \Else
            \State \(F(i,j) \gets F(i+1,j) + \frac{\vp_i}{1 - \vp_i}\,F(i+1,j-1)\)
        \EndIf
    \EndFor
    \State $i \gets i - 1$
\EndFor\label{line:dp_end}

\State Sample \(\ell \in [0,k]\) proportionally to \(F(1,\ell)\) \Comment{\textit{Sample target number of successes.}}
\State \(S \gets \emptyset\), $r \gets \ell$ \Comment{\(r\) \textit{is the number of successes remaining to be allocated.}}
\For{\(i = 1\) \textbf{to} \(d\)}\Comment{\textit{Sequentially select indices.}}
    \If{\(r=0\)}
        \State \textbf{break} \Comment{\textit{No more successes to allocate.}}
    \EndIf
    \State Let $\alpha \sim \Bern(\frac{\vp_i\,F(i+1,r-1)}{(1 - \vp_i)F(i,r)})$
    \If{$\alpha = 1$}
        \State $S \gets S \cup \{i\}$
        \State $r \gets r-1$
    \EndIf
\EndFor
\State \Return \(S\)
\end{algorithmic}
\end{algorithm}

    
    
    
    
    
        
            

\restatecps*
\begin{proof}
The algorithm is displayed in Algorithm~\ref{alg:cond_sampling}, and its runtime is clearly $O(dk)$. We now prove correctness. First, observe that for all $S \subseteq [d]$, 
\[\pi(S) \propto \prod_{i\in S}\frac{\vp_i}{1 - \vp_i} =: P(S).\]
We claim that after Lines~\ref{line:dp_start} to~\ref{line:dp_end} have finished running, we have for all $i \in [d]$, $j \in [k]$, that
    \begin{equation}
    \label{eq:def_F_ij}
        F(i,j) = \sum_{\substack{S \subseteq \{i, i+1, \ldots, d\} \\ |S| = j}} P(S).
    \end{equation}
    In \eqref{eq:def_F_ij}, we define empty sums to be $0$ and $P(\emptyset) \defeq 1$. Note that $F(d + 1, j)$ for $j \in [0, k]$ satisfies \eqref{eq:def_F_ij}. Now, supposing all entries $F(i + 1, j)$ for $j \in [0, k]$ satisfy \eqref{eq:def_F_ij}, we have for any $j \in [0, k]$ that
    \begin{align*}
        \sum_{\substack{S \subseteq \{i, i+1, \ldots, d\} \\ |S| = j}} P(S) 
        &= \sum_{\substack{S \subseteq \{ i+1, i+2, \ldots, d\} \\ |S| = j}} P(S) + \frac{\vp_i}{1 - \vp_i}\sum_{\substack{S \subseteq \{i + 1, i+2, \ldots, d\} \\ |S| = j - 1}}P(S)\\
        &= F(i+1, j) + \frac{\vp_i}{1 - \vp_i}F(i+1, j-1) = F(i, j),
    \end{align*}
    as computed by Lines~\ref{line:dp_start} to~\ref{line:dp_end}.
    Thus, for any $S \in \Omega_k$, we have 
    \begin{equation*}
        \pi(S \mid |S| \leq k) = \frac{P(S)}{\sum_{S \mid |S| \leq k}P(S)} = \frac{P(S)}{\sum_{j\in [0, k]}F(1, j)}.
    \end{equation*}
    In Algorithm~\ref{alg:cond_sampling}, we first sample $\ell$ with probability $\frac{F(1, \ell)}{\sum_{j \in [k]} F(1, \ell)}$. Let $S = \{i_1, i_2, \ldots, i_\ell\}$ have $|S| = \ell$. Algorithm~\ref{alg:cond_sampling} outputs $S$ with probability:
    \begin{align*}
       \frac{F(2, \ell)}{F(1, \ell)}\frac{F(3, \ell)}{F(2, \ell)} \cdots\frac{\vp_{i_1}}{1 - \vp_{i_1}}\frac{F(i_1 + 1, \ell - 1)}{F(i_1, \ell)}\cdots\frac{\vp_{i_l}}{1 - \vp_{i_l}}\frac{F(i_l + 1, 0)}{F(i_l, 1)}\frac{F(d+1, 0)}{F(d, 0)}
        &= \frac{\prod_{i\in S}\frac{\vp_i}{1 - \vp_i}}{F(1, \ell)} \\
        &= \frac{P(S)}{F(1, \ell)}.
    \end{align*}
    Thus, we have 
    \begin{equation*}
        \bbP(\text{output }S) = \frac{P(S)}{F(1, \ell)}\cdot \frac{F(1, \ell)}{\sum_{j \in [0, k]}F(1, j)} = \pi(S \mid |S| \leq k).
    \end{equation*}
\end{proof}

\end{document}